\documentclass[10pt]{article} 
\usepackage[accepted]{tmlr}

\usepackage{amsmath,amsfonts,bm}

\def\eqref#1{equation~\ref{#1}}

\def\1{\bm{1}}

\DeclareMathAlphabet{\mathsfit}{\encodingdefault}{\sfdefault}{m}{sl}
\SetMathAlphabet{\mathsfit}{bold}{\encodingdefault}{\sfdefault}{bx}{n}

\DeclareMathOperator*{\argmin}{arg\,min}

\usepackage{hyperref}
\usepackage{url}
\usepackage{amssymb}
\usepackage{amsthm}
\usepackage{enumitem}
\usepackage{algorithm}
\usepackage{algpseudocode}
\usepackage{booktabs}
\usepackage{multirow}
\usepackage{xcolor}
\usepackage{graphicx}
\usepackage{wrapfig}
\providecommand{\Description}[1]{}

\newtheorem{theorem}{Theorem}
\newtheorem{lemma}[theorem]{Lemma}
\newtheorem{proposition}[theorem]{Proposition}
\newtheorem{corollary}[theorem]{Corollary}
\newtheorem{assumption}{Assumption}
\theoremstyle{remark}
\newtheorem{remark}{Remark}

\title{Implicit Dynamical Flow Fusion (IDFF) for Generative Modeling}

\author{\name Mohammad R. Rezaei \\
      \addr University of Toronto \\
      \addr Vector Institute
      \AND
      \name Milos R. Popovic \\
      \addr University of Toronto
      \AND
      \name Milad Lankarany \\
      \addr University of Toronto
      \AND
      \name Rahul G. Krishnan \\
      \addr University of Toronto \\
      \addr Vector Institute}

\def\month{11}  
\def\year{2026} 
\def\openreview{\url{https://openreview.net/forum?id=RWnohJJKbq}} 

\begin{document}

\maketitle

\begin{abstract}
Conditional Flow Matching (CFM) generates high-quality samples by learning a deterministic transport from noise to data, but typically requires more than a hundred network function evaluations (NFEs) per sample. We propose \emph{Implicit Dynamical Flow Fusion} (IDFF), which augments the CFM vector field with $K$th-order score-derivative terms and attains comparable sample quality while requiring significantly fewer NFEs. At first order ($K=1$), the method reduces to a Langevin-augmented flow, and we prove that the marginal $p_t$ is preserved exactly in continuous time; the empirical NFE reduction is therefore a statement about \emph{discretization robustness}, not about continuous-time acceleration, and we make that reading explicit. For $K=2$, we further provide an error analysis that justifies the optional second-order curvature correction we introduce, showing that its deviation is bounded by the weighted score-matching loss already minimized by our objective; the bound is formulated for general $K$ and guarantees that this correction cannot worsen the order of the overall scheme. The practical enabler behind both regimes is a re-parameterization identity: every higher-order marginal derivative can be obtained from the learned first-order score by automatic differentiation, so no additional networks are trained. Empirically, IDFF reduces NFEs by an order of magnitude with no loss in sample quality. On CIFAR-10 the $K=1$ configuration achieves an FID of 2.78 at 10 NFEs, the best result at that budget among the CFM variants we retrain at a matched backbone and training budget, and on par with methods that need over a hundred evaluations. For time-series modelling, IDFF performs strongly on molecular-dynamics simulation and sea-surface-temperature forecasting at a fraction of the compute. Overall, score-Langevin enhancement of CFM offers a principled and efficient route to few-step generative modelling, with a closed-form error budget for curvature-augmented extensions. Code is available at \url{https://github.com/MrRezaeiUofT/Implicit_Dynamical_Flow_Fusion_for_Genereative_Modeling}.
\end{abstract}

\section{Introduction}
\begin{figure}
\centering
\includegraphics[width=.9\textwidth]{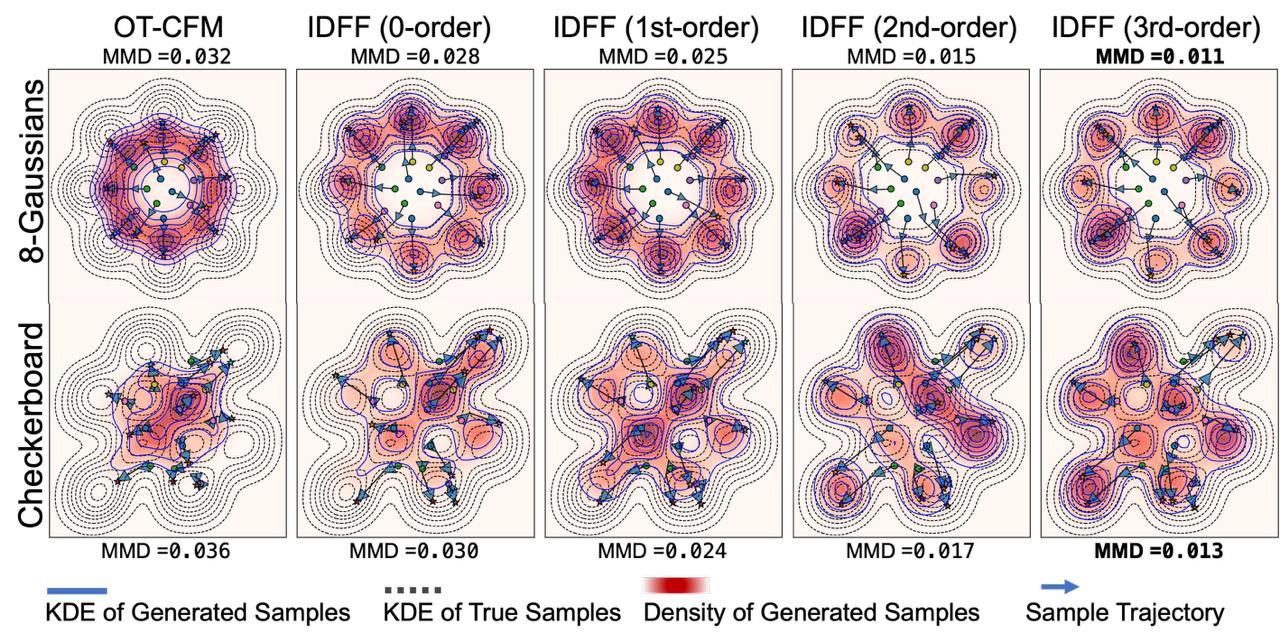}
\caption{Comparison of trajectory sampling between OT-CFM and IDFF at different orders on a 2-D toy problem with NFE${=}2$. The figure displays 4096 final samples generated by each model, with KDE contours shown in blue and ground-truth samples represented by black contours. Twelve individual trajectories are overlaid to illustrate the sampling paths. On this low-dimensional problem the third-order variant attains the smallest Maximum Mean Discrepancy (MMD) to the true distribution. We stress that this is a $d=2$ observation only: in the image and time-series settings of Section~\ref{sec:experiments} the third-order Jacobians are numerically unstable (Appendix~\ref{sec:high_dim_robustness}), and every experiment we report uses $K\leq 2$. The figure is included to illustrate the geometry of the augmented drift, not as evidence that $K=3$ is a usable configuration.}
\label{fig:teaser}
\end{figure}
Diffusion models have emerged as powerful generative tools, iteratively transforming noise into structured data with state-of-the-art results in image generation \citep{song2020score}, text generation \citep{li2022diffusionlm}, video generation \citep{liu2024sora}, and other domains \citep{cachay2023dyffusion, jing2022torsional}. However, a significant practical cost is the number of function evaluations (NFEs) per sample \citep{ho2020denoising}, an issue that persists despite recent advances such as DPM-solvers \citep{lu2022dpm, lu2022dpm2} and Denoising Diffusion Implicit Models \citep{song2020denoising}. Conditional Flow Matching (CFM) \citep{liu2022flow, albergo2022building, albergo2025stochastic} offers an alternative to diffusion models by learning deterministic vector fields that transport probability distributions from noise to data. To minimize trajectory lengths, OT-CFMs \citep{tong2023improving} incorporate optimal transport couplings. However, sampling from CFMs still requires over 100 NFEs for high-quality generation \citep{dao2023flow}, which is particularly problematic for time-series applications where costs scale with sequence length.

Methods that use higher spatial derivatives of the log-density incorporate geometric information about the probability landscape. \citet{lu2022maximum} and \citet{meng2021estimating} estimate such higher-order score derivatives with auxiliary networks, and \citet{dockhorn2022genie} use them inside a higher-order denoising solver. A key challenge in all such methods is the gap between \emph{conditional} quantities available during training and \emph{marginal} quantities needed for correct sampling dynamics.

We introduce \emph{Implicit Dynamical Flow Fusion} (IDFF), which augments CFM with score-derivative terms $\nabla^k\log p_t$ together with a matched diffusion compensation (Figure~\ref{fig:teaser} illustrates the effect on a 2D-toy distribution). The order $K$ counts \emph{spatial derivatives of the score}, not network evaluations per step and not auxiliary state variables (Section~\ref{sec:related} separates the three senses of ``higher order''); our theory cleanly separates two regimes.

\paragraph{First-order regime ($K=1$): exact preservation.}
Our default schedule simply \emph{adds} a Langevin term $\gamma_t^1\,\nabla\log p_t$ to the CFM drift, together with the diffusion compensation $g_t^2 = 2\gamma_t^1$ (no slowdown of $\mathbf v_t$). With this schedule, the augmented SDE preserves the CFM marginal \emph{exactly}; the proof is one line of Fokker--Planck cancellation (Theorem~\ref{thm:first_order_preservation}). At the SDE level this is the same Langevin-augmented probability-flow construction that has appeared in the score-Langevin SDEs of \citet{song2020score} and the stochastic-interpolant SDEs of \citet{albergo2025stochastic}; the role of the $K=1$ result here is therefore not the SDE itself, but to serve as the rigorous baseline on top of which the closed-form error budget for the $K\geq 2$ regime and for learned scores is built (Theorem~\ref{thm:higher_order_deviation}). Because the continuous-time law is unchanged, the empirical NFE reduction we report is a discretization effect; Section~\ref{sec:why_nfe} states the mechanism we believe is responsible and the evidence for it.

\paragraph{Higher-order regime ($K\geq 2$): controlled deviation.}
At higher orders we prove a quantitative endpoint bound in which the leading term is $\sqrt{\tfrac12\mathcal{E}_{\mathrm{score}}}$, and $\mathcal{E}_{\mathrm{score}}$ is \emph{exactly} the weighted $L^2$ score-matching loss our training objective already minimizes. Under the recommended schedule $\gamma_t^k = O(\Delta t^{k-1})$, this gives $\mathrm{TV}(\tilde p_1,p_1)\to 0$ as $\mathrm{NFE}\to\infty$ for any converged score network (Theorem~\ref{thm:higher_order_deviation}, Corollary~\ref{cor:nfe_convergence}). The bound certifies that the $K\geq 2$ correction does not degrade the order of the scheme.

\paragraph{The practical enabler.}
Training only has access to \emph{conditional} log-density derivatives, but the dynamics need \emph{marginal} ones; at $k=1$, Fisher's identity bridges the gap, but no such identity holds for $k\geq 2$. We show (Theorem~\ref{thm:kth_order_marginal}) that the optimal first-order score $\hat{s}_t^*$ already encodes all higher orders: $\nabla^k \log p_t = \nabla_{\mathbf{x}}^{k-1} \hat{s}_t^*$. So a single trained score network suffices, and the higher-order score derivatives are computed at sampling time by differentiating it automatically.

The contributions of our work are as follows:
\begin{enumerate}[align=left]
    \item We show that once Fisher's identity \citep{cappe2005inference} grants $\hat s_t^*=\nabla\log p_t$ at the optimum of standard score matching, every higher-order marginal derivative is obtained by the chain-rule re-parameterization $\nabla^k\log p_t = \nabla_{\mathbf{x}}^{k-1}\hat s_t^*$. A single trained score head therefore supplies all higher-order score derivatives at sampling time via autodiff, with no auxiliary higher-order networks required. We additionally quantify the propagation of score-estimation error into higher orders.

    \item We restate that the $K=1$ Langevin-enhanced SDE serving as our exact-preserving baseline is itself a known marginal-preserving construction (\citealp{song2020score, albergo2025stochastic}; Corollary~\ref{cor:idff_langevin_albergo}). On top of it we build a unified Girsanov$\to$Pinsker certificate: a single closed-form path-KL bound that covers (i) the exact $K=1$ default ($\mathrm{TV}=0$ in continuous time, Theorem~\ref{thm:first_order_preservation}), (ii) the convex-constrained $K=1$ variant ($\mathrm{TV}=O(\sqrt{\bar\gamma})$, Proposition~\ref{prop:K1_convex_deviation}), and (iii) the second-order correction we deploy, $K=2$ with a learned score ($\mathrm{TV}\leq\sqrt{\tfrac12\mathcal E_{\mathrm{score}}}+O(\Delta t)$, Theorem~\ref{thm:higher_order_deviation}, stated there for general $K$), with the same bound holding uniformly in $t$ (Corollary~\ref{cor:intermediate_TV}). We emphasise that this bound certifies that the $K\geq 2$ correction does not degrade the scheme; we do \emph{not} claim exact endpoint preservation at $K\geq 2$ for finite NFE.

    \item Our empirical results show that with $K=1$, IDFF generates high-quality images at 5--10 NFEs on CIFAR-10, CelebA, ImageNet-64, CelebA-HQ, LSUN-Church, and LSUN-Bedroom, where standard CFM variants need over 100 NFEs. For time-series modelling (3D attractors, molecular dynamics, sea-surface-temperature forecasting), IDFF reduces sampling cost by an order of magnitude while improving quality. To pin down what is actually driving these gains, we provide a dedicated ablation (Table~\ref{tab:ablation-langevin-decomp}) that isolates the score-Langevin drift from generic SDE stochasticity and from the extra capacity of the joint score head.
\end{enumerate}

\section{Background and Preliminaries}

Generative modeling aims to learn the underlying structure of complex high-dimensional data distributions from finite samples. Given empirical samples from a dataset $\mathcal{D} = \{\mathbf{x}^{(i)}\}_{i=1}^M$ where each sample $\mathbf{x}^{(i)} \in \mathbb{R}^d$ is drawn from an unknown data distribution $p_{\text{data}}(\mathbf{x})$, the goal is to learn a model that can generate new samples that match $p_{\text{data}}$. For time-series data, $\mathcal{D} = \{\mathbf{x}^{(i)}_{1:N}\}_{i=1}^M$ where $\mathbf{x}^{(i)}_n \in \mathbb{R}^d$ for $n \in \{1, \ldots, N\}$, and the goal is to model the joint law, which we factorise as
\[
p_{\text{data}}(\mathbf{x}_{1:N}) \;=\; p_{\text{data}}(\mathbf{x}_1)\prod_{n=2}^{N} p_{\text{data}}(\mathbf{x}_n \,|\, \mathbf{x}_{1:n-1}),
\]
so that the first factor $p_{\text{data}}(\mathbf{x}_1)$ is modelled unconditionally and the conditional factors are modelled for $n\geq 2$. Appendix~\ref{app:timeseries} uses the Markov specialisation $p(\mathbf{x}_{1:N}) = p(\mathbf{x}_1)\prod_{n=2}^{N} p(\mathbf{x}_n\,|\,\mathbf{x}_{n-1})$.

Throughout, $\sigma_t = \sigma_0\sqrt{t(1-t)}$ denotes the \emph{standard deviation of the conditional probability path} used at training time, while $g_t$ denotes the \emph{diffusion coefficient of the sampling SDE}. These are distinct objects and we keep them notationally separate.

\subsection{Diffusion Models}

Score-based generative models learn to reverse a forward SDE $d\mathbf{x}_t=\mathbf{f}(\mathbf{x}_t,t)dt+g(t)d\mathbf{w}_t$ that carries $p_0=p_{\text{data}}$ to a Gaussian prior; the reverse-time SDE replaces the drift by $\mathbf{f}-g^2\nabla_{\mathbf{x}}\log p_t(\mathbf{x}_t)$, so a network need only learn the score $\nabla_{\mathbf{x}}\log p_t$, and the density evolves under the Fokker--Planck equation $\partial_t p_t=-\nabla\cdot(\mathbf{f}p_t)+\tfrac{g(t)^2}{2}\Delta p_t$ \citep{song2020score}. Appendix~\ref{app:higher_order_background} states the forward and reverse SDEs in full.

\citet{lu2022maximum} extend score matching beyond first-order gradients of the log density, and \citet{meng2021estimating} study the estimation of these higher-order gradients by denoising directly. Both introduce the higher-order score variables $\hat{\boldsymbol{\epsilon}}^{(k)}_t(\mathbf{x}) = \nabla_{\mathbf{x}}^k \log p_t(\mathbf{x})$, the $k$-th order derivatives of the log-density, which we use as notation throughout. Under Gaussian transitions these have closed conditional forms, and prior work trains a separate network $\hat{\boldsymbol{\epsilon}}^{(k)}(\mathbf{x}_t,t;\theta_k)$ per order to match them, then adds the learned terms to the reverse-time drift with time-dependent coefficients; Appendix~\ref{app:higher_order_background} states these targets and the resulting SDE. Section~\ref{sec:method} avoids the auxiliary networks entirely.

\subsection{Conditional Flow Matching (CFM)}
Conditional Flow Matching learns transport maps between probability distributions. Given a time-dependent vector field $\mathbf{v}_t: \mathbb{R}^d \times [0,1] \to \mathbb{R}^d$ we can transport samples from a base distribution $p_0 = \mathcal{N}(\mathbf{0}, \mathbf{I})$ to a target distribution $p_1 = p_{\text{data}}$ via $\frac{d\mathbf{x}_t}{dt} = \mathbf{v}_t(\mathbf{x}_t)$ with $\mathbf{x}_0 \sim p_0$; this induces a probability path $p_t(\mathbf{x})$. The resulting continuity equation
\begin{equation}
\frac{\partial p_t(\mathbf{x})}{\partial t} + \nabla \cdot (p_t(\mathbf{x}) \mathbf{v}_t(\mathbf{x})) = 0,
\end{equation}
ensures that probability mass is conserved along the trajectories.

\paragraph{Learning objective and sampling.}
Instead of modeling the marginal flow directly, CFM decomposes the complex marginal flow into simpler conditional flows. Given pairs $\mathbf{z} = (\mathbf{x}_0, \mathbf{x}_1)$ where $\mathbf{x}_0$ is typically a sample from a normal distribution and $\mathbf{x}_1$ are samples from $\mathcal{D}$, we can define Gaussian conditional paths $p_t(\mathbf{x}_t|\mathbf{x}_0, \mathbf{x}_1) = \mathcal{N}(\mathbf{x}_t; \mu_t, \sigma_t^2\mathbf{I})$ where $\mu_t = t\mathbf{x}_1 + (1-t)\mathbf{x}_0$ and $\sigma_t = \sigma_0\sqrt{t(1-t)}$. Then, the conditional vector field becomes
\begin{equation}
\mathbf{v}_t(\mathbf{x}_t|\mathbf{x}_0, \mathbf{x}_1) = \mathbf{x}_1 - \mathbf{x}_0 + \frac{\dot{\sigma}_t}{\sigma_t}(\mathbf{x}_t - \mu_t),
\label{eq:cfm_cond_field}
\end{equation}
simplifying to $\mathbf{v}_t = \mathbf{x}_1 - \mathbf{x}_0$ when $\sigma_0 = 0$. CFMs train a neural network $\hat{\mathbf{v}}_t(\mathbf{x}; \theta)$ with parameters $\theta$ to approximate the conditional field via
\begin{equation}
\mathcal{L}_{\text{CFM}}(\theta) = \mathbb{E}_{t \sim \mathcal{U}[0,1], (\mathbf{x}_0, \mathbf{x}_1) \sim q, \mathbf{x}_t} \left\|\hat{\mathbf{v}}_t(\mathbf{x}_t; \theta) - \mathbf{v}_t(\mathbf{x}_t|\mathbf{x}_0, \mathbf{x}_1)\right\|^2,
\end{equation}
where $\hat{\mathbf{v}}_t$ takes as input the current position $\mathbf{x}_t\sim p_t(\cdot|\mathbf{x}_0,\mathbf{x}_1)$ and time $t$ and outputs a velocity vector in $\mathbb{R}^d$.
The coupling $q(\mathbf{x}_0, \mathbf{x}_1)$ is a joint distribution over pairs of noise and data samples, which can be independent, $q(\mathbf{x}_0, \mathbf{x}_1) = p_0(\mathbf{x}_0) p_{\text{data}}(\mathbf{x}_1)$, or use optimal transport couplings that minimize the expected transport cost $\mathbb{E}_{q}[\|\mathbf{x}_1 - \mathbf{x}_0\|^2]$.
After training, samples are generated by integrating $\mathbf{x}_{t+\Delta t} = \mathbf{x}_t + \hat{\mathbf{v}}_t(\mathbf{x}_t; \theta) \Delta t$ from $\mathbf{x}_0 \sim p_0$ with step size $\Delta t = 1/\text{NFE}$. Section~\ref{sec:method} uses the equivalent field obtained by conditioning on $\mathbf{x}_1$ alone rather than on the pair $(\mathbf{x}_0,\mathbf{x}_1)$, namely $\mathbf{v}_t(\mathbf{x}_t|\mathbf{x}_1)=(\mathbf{x}_1-\mathbf{x}_t)/(1-t)$; by Theorem~\ref{thm:cfm_validity} both conditionings induce the same marginal field, since each is a conditional expectation of \eqref{eq:cfm_cond_field} under a coarser conditioning.

The validity of the conditional vector field rests on the following fundamental result:
\begin{theorem}[CFM Marginal Consistency \citep{lipman2022flow}]
\label{thm:cfm_validity}
If conditional paths $p_t(\mathbf{x}|\mathbf{z})$ with fields $\mathbf{v}_t(\mathbf{x}|\mathbf{z})$ satisfy the continuity equation, then the marginal $p_t(\mathbf{x}) = \int p_t(\mathbf{x}|\mathbf{z})q(\mathbf{z})d\mathbf{z}$ with $\mathbf{v}_t(\mathbf{x}) = \mathbb{E}_{q(\mathbf{z}|\mathbf{x})}[\mathbf{v}_t(\mathbf{x}|\mathbf{z})]$ also satisfies the continuity equation.
\end{theorem}
By Theorem~\ref{thm:cfm_validity}, learning to match $\mathbf{v}_t(\mathbf{x}_t|\mathbf{x}_0, \mathbf{x}_1)$ is sufficient to recover the marginal flow and results in being able to sample from $p_{\text{data}}$.

\subsection{Challenges in Incorporating Higher-Order Terms into CFM}
\label{sec:challenges}

Diffusion and flow matching are not separate paradigms: the score, noise, denoiser and velocity parameterizations are related by standard identities, and probability-flow ODEs and stochastic interpolants provide a common framework for both \citep{song2020score, albergo2025stochastic, gao2025diffusion}. Our starting point is therefore not a cross-paradigm transfer but a concrete algebraic obstruction: augmenting the CFM velocity with any score-derivative term leaves a residual $\sum_k \gamma_t^k\,\nabla\!\cdot\!(p_t\nabla^k\log p_t)$ in the continuity equation~\eqref{eq:naive_augmentation_residual}, which is already nonzero at $k=1$. The questions we address are what compensation makes this residual vanish, and how large the residual is when it cannot be made to vanish exactly.

Concretely, the direct addition of score-derivative terms to the vector field for CFMs runs into two technical hurdles that motivate every design choice in IDFF.

\paragraph{Challenge 1: continuity-equation violation.}
Consider a naive augmentation $\tilde{\mathbf{v}}_t = \gamma_t^0\,\mathbf{v}_t + \sum_{k=1}^K \gamma_t^k \hat{\boldsymbol{\epsilon}}^{(k)}$ where $\hat{\boldsymbol{\epsilon}}^{(k)} = \nabla_{\mathbf{x}}^k \log p_t$ and $\gamma_t^0>0$ is the coefficient on the base velocity (the general form used from Eq.~\eqref{eq:kth_order_idff} onward; the default sets $\gamma_t^0\equiv 1$ and the convex-constrained variant sets $\gamma_t^0 = 1-\sum_{k\geq 1}\gamma_t^k$). Substituting into the continuity equation yields
\begin{equation}
\frac{\partial p_t}{\partial t} + \nabla \cdot (p_t \tilde{\mathbf{v}}_t) \;=\; \frac{\partial p_t}{\partial t} + \gamma_t^0\,\nabla \cdot (p_t \mathbf{v}_t) + \sum_{k=1}^K \gamma_t^k \nabla \cdot (p_t \nabla_{\mathbf{x}}^k \log p_t).
\label{eq:naive_augmentation_residual}
\end{equation}
Since the original CFM satisfies $\partial_t p_t + \nabla\cdot(p_t\mathbf{v}_t) = 0$, the right-hand side equals $(1-\gamma_t^0)\,\partial_t p_t + \sum_{k=1}^K \gamma_t^k \nabla\cdot(p_t \nabla_{\mathbf{x}}^k \log p_t)$, so the augmented field is consistent with the same marginal flow only if \emph{both} contributions vanish. At the default $\gamma_t^0\equiv 1$ the first term disappears and the condition reduces to $\sum_{k=1}^K \gamma_t^k \nabla\cdot(p_t \nabla_{\mathbf{x}}^k \log p_t)\equiv 0$; already at first order, $\nabla\cdot(p_t \nabla_{\mathbf{x}}\log p_t) = \Delta p_t\neq 0$ in general. Under the convex constraint $\gamma_t^0 = 1-\sum_{k\geq 1}\gamma_t^k<1$ the additional slowdown term $(1-\gamma_t^0)\partial_t p_t$ survives as well, which is exactly the source of the residual quantified in Proposition~\ref{prop:K1_convex_deviation}. Either way the naive augmentation violates the hypotheses of Theorem~\ref{thm:cfm_validity}, and the augmented flow no longer transports $p_0$ to $p_1 = p_{\text{data}}$ exactly.

\paragraph{Challenge 2: conditional-marginal gap.}
We can only train on conditional targets $\nabla^k \log p_t(\mathbf{x}_t|\mathbf{x}_1)$, but sampling requires marginal quantities $\nabla^k \log p_t(\mathbf{x}_t)$. For $k=1$, Fisher's identity ensures these coincide at the optimum. For $k \geq 2$, prior work treats the conditional--marginal gap as an approximation and closes it by training auxiliary networks. The route we take instead is a re-parameterization: at the optimum the marginal $k$-th derivative is already the $(k-1)$-th derivative of the score, $\nabla^k \log p_t = \nabla_{\mathbf{x}}^{k-1} \hat{s}_t^*$, so no gap beyond the $k=1$ closure of Fisher's identity has to be closed at all. For a learned $\hat{s}_t \approx \hat{s}_t^*$, this holds approximately, enabling higher-order dynamics without separate training while inheriting approximation quality from score estimation.

Section~\ref{sec:method} answers both challenges: a compensating diffusion $g_t^2=2\gamma_t^1$ exactly cancels the first-order term of residual~\eqref{eq:naive_augmentation_residual}, a coefficient-scaling rule $\gamma_t^k=O(\Delta t^{k-1})$ bounds what remains at $K\geq 2$, and Theorem~\ref{thm:kth_order_marginal} supplies the higher-order marginals by autodiff of $\hat{s}_t$ rather than by training auxiliary networks.

\section{Implicit Dynamical Flow Fusion Model (IDFF)}
\label{sec:method}

IDFF augments the CFM vector field with a score-derivative drift that accelerates convergence of the \emph{discretized} sampler without breaking marginal consistency. The first-order augmented flow is
\begin{equation}
\tilde{\mathbf{v}}_t(\mathbf{x}_t) \;=\; \mathbf{v}_t(\mathbf{x}_t) \;+\; \gamma_t^1\,\nabla_{\mathbf x} \log p_t(\mathbf{x}_t),
\label{eq:augmented_K1}
\end{equation}
where $\gamma_t^1\geq 0$ is a Langevin-amplitude schedule that vanishes at the endpoints ($\gamma_0^1=\gamma_1^1=0$). The score-Langevin drift $\boldsymbol{\xi}_t = \gamma_t^1\,\nabla_{\mathbf x}\log p_t(\mathbf{x}_t)$ pushes particles along the log-density gradient, back towards high-probability regions of $p_t$. Paired with the diffusion compensation $g_t^2=2\gamma_t^1$, this is exactly the Langevin enhancement of CFM that preserves the marginal $p_t$ exactly (Theorem~\ref{thm:first_order_preservation}).

For full generality, we write the same drift in the form $\tilde{\mathbf{v}}_t = \gamma_t^0\,\mathbf{v}_t+\gamma_t^1\,\nabla\log p_t$. Our default sets $\gamma_t^0\equiv 1$ and recovers the Langevin-enhanced flow above. The alternative convex-constrained schedule $\gamma_t^0+\gamma_t^1=1$ is then a single notational change: it also slows $\mathbf{v}_t$, no longer preserves $p_t$ exactly, and incurs the quantified TV cost of Proposition~\ref{prop:K1_convex_deviation}.

We can generalize this drift to incorporate higher-order score derivatives. For $k \geq 2$, the quantity $\nabla_x^k \log p_t$ is a $k$-th order tensor; to obtain a vector-valued contribution, we define the \emph{contracted score-derivative terms}
\begin{equation}
\mathbf{m}_t^{(1)} = \nabla_x \log p_t(\mathbf{x}_t),\qquad
\mathbf{m}_t^{(k)} = \bigl(\nabla_x^k \log p_t(\mathbf{x}_t)\bigr)[\mathbf{u}_t]^{k-1} \in \mathbb{R}^d \quad (k \geq 2),
\label{eq:momentum_contraction}
\end{equation}
where $\mathbf{u}_t = \mathbf{v}_t/\|\mathbf{v}_t\|$ is the normalized velocity direction and $[\mathbf{u}_t]^{k-1}$ denotes $(k-1)$-fold contraction along $\mathbf{u}_t$. Explicitly, $\mathbf{m}_t^{(2)}$ is the Hessian-vector product capturing directional curvature along the flow. The generalized score-derivative drift is then
\begin{equation}
\boldsymbol{\xi}_t(\mathbf{x}_t) = \sum_{k=1}^K \gamma_t^k \mathbf{m}_t^{(k)}(\mathbf{x}_t).
\end{equation}
The second-order term carries local curvature information; higher orders would in principle carry finer geometry of the log-density, but we do not deploy them (Appendix~\ref{sec:high_dim_robustness}).

At $K=1$ with the default schedule ($\gamma_t^0\equiv 1$, $g_t^2=2\gamma_t^1$) the augmented density equals the CFM marginal \emph{exactly} in continuous time, $\tilde p_t = p_t$ for all $t\in[0,1]$ (Theorem~\ref{thm:first_order_preservation}); the convex variant $\gamma_t^0+\gamma_t^1=1$ also slows $\mathbf{v}_t$ and pays a closed-form $O(\sqrt{\bar\gamma})$ TV cost (Proposition~\ref{prop:K1_convex_deviation}, Corollary~\ref{cor:K1_schedule_constants}). At $K\geq 2$ preservation is no longer exact: the deviation obeys $\mathrm{TV}(\tilde p_1,p_1)\leq\sqrt{\tfrac{1}{2}\mathcal{E}_{\mathrm{score}}}+O(\Delta t)$, and the same bound holds uniformly in $t$ (Theorem~\ref{thm:higher_order_deviation}, Corollary~\ref{cor:intermediate_TV}), where $\mathcal{E}_{\mathrm{score}}$ is precisely the weighted $L^2$ score-matching loss our training objective already minimizes. Endpoint preservation is asymptotic, not exact, at finite NFE. The statements are proved for general $K$, but our experiments on real data use only a second-order correction or less. For $K=2$ we claim only that the budget guarantees the correction does not degrade the order of the scheme; we do not claim it accelerates sampling.

The $K=1$ exact-preserving SDE is a member of the Langevin-augmented probability-flow family of~\citet{song2020score} and~\citet{albergo2025stochastic}, recovered as a special case in Corollary~\ref{cor:idff_langevin_albergo}. We use it as the base on which the error budget is built; what is new is the unified Girsanov$\to$Pinsker certificate that covers the convex-constrained $K=1$ variant, the learned score, and the $K\geq 2$ correction in a single closed-form bound. Both bounds follow one argument on path space: because the diffusion is non-degenerate on the interior ($g_t^2=2\gamma_t^1>0$), Girsanov applies to the path measure directly and leaves the weighted $L^2$ drift mismatch (exactly our training loss) on the right-hand side. This avoids a transport--Gr\"onwall step that would require regularity on $\nabla\log p_t\cdot\mathbf{m}_t^{(k)}$ we do not assume, and follows recent flow-matching convergence analyses \citep{chen2023sampling, benton2024error}.

\subsection{Why the Score-Langevin Drift Reduces NFE}
\label{sec:why_nfe}

Theorem~\ref{thm:first_order_preservation} says that at $K=1$ the augmented SDE has the \emph{same} time-marginals as the CFM probability flow. It therefore cannot, and does not, accelerate anything in continuous time. The empirical NFE reduction must consequently be a property of the discretization, and we state here the mechanism we believe is responsible, together with the evidence for it. We present this as an explanation of the observed behaviour, not as a theorem.

\paragraph{Error forgetting versus error accumulation.}
Write $\rho_i$ for the law of the numerical iterate after $i$ Euler(--Maruyama) steps and $p_{t_i}$ for the exact CFM marginal. For the deterministic CFM/ODE sampler, the update map is a diffeomorphism: the local truncation error committed at step $i$ is transported forward by the remaining flow, and the discrepancy $\rho_i \not= p_{t_i}$ has no mechanism to shrink. Errors therefore accumulate additively across the trajectory, which is why the ODE needs many small steps.

Adding the pair $(\gamma_t^1\nabla\log p_t,\;g_t^2=2\gamma_t^1)$ changes this. At a frozen time $t$, that pair is exactly the overdamped Langevin generator whose invariant law is $p_t$; the transport term $\mathbf{v}_t$ moves the target, and the Langevin part relaxes the current law \emph{towards} the moving target. If $p_t$ satisfies a log-Sobolev inequality with constant $C_{\mathrm{LSI}}(t)$ on the region visited by the sampler, the Langevin part contracts relative entropy at rate $2\gamma_t^1/C_{\mathrm{LSI}}(t)$, so heuristically
\begin{equation}
\mathrm{KL}(\rho_{i+1}\,\|\,p_{t_{i+1}})
\;\lesssim\;
e^{-2\gamma_{t_i}^1\Delta t / C_{\mathrm{LSI}}(t_i)}\;\mathrm{KL}(\rho_{i}\,\|\,p_{t_i})
\;+\;C\,\Delta t^{2},
\label{eq:heuristic_contraction}
\end{equation}
i.e.\ the error injected at each step is subsequently damped rather than merely transported. The sampler is in this sense self-correcting: it does not need each individual step to be accurate, only for the accumulated deviation to stay inside the basin in which the score can pull it back. This is the standard reason stochastic samplers tolerate coarse discretizations better than their probability-flow ODE counterparts \citep{song2020score, karras2022elucidating}, and it is what the schedule shape exploits: $\gamma_t^1\propto t(1-t)$ places the strongest correction in the middle of the flow, where the marginal is most complex and the truncation error largest, while switching it off at the endpoints where exactness is required.

Equation~\eqref{eq:heuristic_contraction} predicts that neither ingredient works alone, and this is what we observe (Table~\ref{tab:ablation-langevin-decomp}). Injecting noise with the IDFF diffusion schedule but \emph{no} score drift is not a $p_t$-preserving Langevin pair; it blurs the iterate and leaves about 40\% of the gap to IDFF open ($11.87\to 6.41$, against $2.78$). Adding the score drift with \emph{no} diffusion is likewise not a Langevin pair; it biases the flow towards the modes and \emph{hurts} ($9.74$). Only the matched combination (the one the Fokker--Planck cancellation singles out) reaches $2.78$, and mismatching the constant in $g_t^2=c\,\gamma_t^1$ degrades it again (Table~\ref{tab:ablation-diffusion}). The gain is therefore attributable to the specific score-Langevin pairing rather than to stochasticity, to extra capacity, or to a modified continuous-time target.

This account also explains why we do \emph{not} claim NFE reduction from $K\geq 2$. The second-order term changes the drift without a compensating cancellation, so it perturbs the continuous-time law; Theorem~\ref{thm:higher_order_deviation} bounds that perturbation but supplies no contraction of the kind in \eqref{eq:heuristic_contraction}. Every NFE-reduction claim in this paper is a $K=1$ claim.

\subsection{Higher-Order Marginals from the Learned Score}

A central challenge is that we train on \emph{conditional} targets while the dynamics require \emph{marginal} quantities $\nabla^k \log p_t(\mathbf{x}_t)$. For $k=1$, Fisher's identity ensures these coincide at the optimum. For $k \geq 2$, no such conditional-marginal closure exists. Theorem~\ref{thm:kth_order_marginal} below resolves this by a \emph{re-parameterization identity}: once $\hat s_t^* = \nabla\log p_t$ is granted by Fisher's identity, every higher-order marginal derivative is obtained by differentiating $\hat s_t^*$ a further $(k-1)$ times,
\begin{equation}
\nabla^k \log p_t(\mathbf{x}_t) = \nabla_{\mathbf{x}}^{k-1} \hat{s}_t^*(\mathbf{x}_t).
\label{eq:key_identity}
\end{equation}
This is the chain rule applied to the relation $\hat s_t^*=\nabla\log p_t$, and consequently does \emph{not} close a new conditional-marginal gap beyond the $k=1$ closure that Fisher's identity already provides. It is therefore stated below as a re-parameterization lemma. It shows that a single trained score network suffices to instantiate the higher-order score derivatives at sampling time via automatic differentiation, with no auxiliary networks. For a learned approximation $\hat{s}_t \approx \hat{s}_t^*$, \eqref{eq:key_identity} holds approximately and the error is inherited from score estimation (quantified in Theorem~\ref{thm:higher_order_deviation}, Part~3).

\begin{theorem}[Lemma: Re-parameterization of Higher-Order Marginals]
\label{thm:kth_order_marginal}
Under Assumption~\ref{ass:regularity}(R1), let $\hat{s}_t^*(\mathbf{x}_t) = \nabla \log p_t(\mathbf{x}_t)$ be the optimal first-order score (which equals the marginal score by Fisher's identity, Lemma~\ref{lem:fisher_app}). Then for every $k$ with $1\leq k\leq K$,
\begin{equation}
\nabla^k \log p_t(\mathbf{x}_t) \;=\; \nabla_{\mathbf{x}}^{k-1} \hat{s}_t^*(\mathbf{x}_t).
\end{equation}
In particular, the marginal Hessian equals the Jacobian of the optimal score, $\nabla^2 \log p_t(\mathbf{x}_t) = \nabla_{\mathbf{x}} \hat{s}_t^*(\mathbf{x}_t)$.
\end{theorem}

\begin{proof}
Fisher's identity \citep{cappe2005inference} gives $\hat s_t^*=\nabla\log p_t$ at the optimum of the score term of \eqref{eq:idff_loss} (Lemma~\ref{lem:fisher_app}). Differentiating that \emph{equation} $k-1$ further times in $\mathbf{x}$ gives $\nabla_{\mathbf{x}}^{k-1}\hat s_t^* = \nabla_{\mathbf{x}}^{k-1}\nabla\log p_t = \nabla^k\log p_t$, which is legitimate for $k\leq K$ under the $C^K$ hypothesis of R1.
\end{proof}

\begin{remark}[How to read Theorem~\ref{thm:kth_order_marginal}]
\label{rem:reading_thm2}
The mathematical content of the statement is the chain rule, and we do not claim otherwise; what it buys is a change in the \emph{training pipeline}. Three points make the reading precise.
(i) \emph{What is being asserted.} The equality is between two functions of $\mathbf{x}$ at fixed $t$: the $k$-th spatial derivative of the log-marginal, and the $(k-1)$-th spatial derivative of the function that the score head converges to. It says nothing about the conditional quantities $\nabla^k\log p_t(\mathbf{x}_t|\mathbf{x}_1)$, which for $k\geq2$ are constant ($-\mathbf{I}/\sigma_t^2$ at $k=2$, zero beyond) and carry no information about the marginal geometry.
(ii) \emph{Why it is not vacuous in practice.} Prior work \citep{lu2022maximum, meng2021estimating, dockhorn2022genie} regresses a separate network onto those conditional targets for each order. Theorem~\ref{thm:kth_order_marginal} says such networks are redundant at the optimum: the information is already in $\hat s_t^*$, and the only thing needed at sampling time is autodiff. The saving is one trained network per order, not a new statistical result.
(iii) \emph{What breaks with a learned score.} The identity holds for $\hat s_t^*$. For a learned $\hat s_t$, the object the sampler actually uses is $\nabla_{\mathbf{x}}^{k-1}\hat s_t$, whose distance from $\nabla^k\log p_t$ is a $C^{k-1}$ distance rather than an $L^2$ distance on the score itself (Proposition~\ref{prop:error_propagation}). Score matching controls the $L^2$ error, not the derivative error, and this gap is the reason we cap $K$ at 2; we measure the resulting amplification empirically in Table~\ref{tab:error-propagation}.
\end{remark}

Because $\hat s_t$ is already trained, all higher-order marginal derivatives are available at sampling time by autodiff, so unlike \citet{lu2022maximum}, \citet{meng2021estimating} and GENIE \citep{dockhorn2022genie} we train no auxiliary higher-order networks. Appendix~\ref{app:corrected_theory} also includes a more detailed Tweedie-style derivation for the case $k=2$ (Remark~\ref{rem:tweedie_route}, Theorems~\ref{thm:tweedie}, \ref{thm:main_app}). This alternative route is logically equivalent to the argument presented here and is not required by the sampler. We keep it because the posterior covariance derived from the score Jacobian, in the sense of \citet{efron2011tweedie} and \citet{boys2024tweedie}, is used in the practical error analysis in Appendix~\ref{app:practical_error}.

\subsection{Central Theorem: K-th Order IDFF Continuity and Sampling}

\begin{theorem}[K-th Order IDFF Continuity and Sampling]
\label{thm:kth_order_idff}
Let $\mathbf{v}_t(\mathbf{x}_t)$ be a vector field that generates marginal density $p_t(\mathbf{x}_t)$ via the continuity equation. Define the $K$-th order IDFF augmented vector field (with $\gamma_t^0\equiv 1$ as the default; the convex-constrained variant uses $\gamma_t^0=1-\sum_{k\geq 1}\gamma_t^k$ and is handled by Proposition~\ref{prop:K1_convex_deviation}) as
\begin{equation}
\tilde{\mathbf{v}}_t(\mathbf{x}_t) \;=\; \gamma_t^0\,\mathbf{v}_t(\mathbf{x}_t) \;+\; \sum_{k=1}^K \gamma_t^k\,\mathbf{m}_t^{(k)}(\mathbf{x}_t),
\label{eq:kth_order_idff}
\end{equation}
where $\mathbf{m}_t^{(1)} = \hat{s}_t$ and $\mathbf{m}_t^{(k)} = (\nabla_{\mathbf{x}}^{k-1}\hat{s}_t)[\mathbf{u}_t]^{k-1}$ for $k \geq 2$. Sampling integrates the SDE
\begin{equation}
d\mathbf{x}_t \;=\; \tilde{\mathbf{v}}_t(\mathbf{x}_t)\,dt \;+\; g_t\,d\mathbf{w}_t,
\label{eq:idff_sde}
\end{equation}
with diffusion coefficient $g_t$, boundary conditions $g_t \to 0$ and $\{\gamma_t^k\}_{k=1}^{K} \to 0$ as $t \to \{0, 1\}$, and Assumption~\ref{ass:regularity}; the following hold:
\begin{enumerate}
    \item \textbf{$K=1$, exact marginal preservation (default schedule).}
    Take $K=1$, $\gamma_t^0\equiv 1$, $g_t^2=2\gamma_t^1$, and the optimal score $\hat{s}_t=\hat{s}_t^*$. Then the augmented density equals the CFM marginal in continuous time:
    \[
        \tilde p_t(\mathbf{x}) \;=\; p_t(\mathbf{x}) \quad \text{for all } t\in[0,1]
    \]
    (Theorem~\ref{thm:first_order_preservation}). Two natural sources of error appear in practice: Euler--Maruyama discretization contributes $O(\Delta t)$, and using a learned $\hat{s}_t\ne\hat{s}_t^*$ adds a Girsanov bias controlled by $\|\hat{s}_t-\hat{s}_t^*\|_{L^2(p_t)}$ (Theorem~\ref{thm:higher_order_deviation}). The convex-constrained variant $\gamma_t^0+\gamma_t^1=1$ trades exactness for the closed-form bound
    \[
        \mathrm{TV}(\tilde p_1,p_1) \;\leq\; \tfrac{1}{2\sqrt{2}}\,\Bigl(\textstyle\int_0^1\gamma_t^1\,\mathbb{E}\|\mathbf{v}_t\|^2\,dt\Bigr)^{1/2}
    \]
    (Proposition~\ref{prop:K1_convex_deviation}; explicit constants in Corollary~\ref{cor:K1_schedule_constants}).

    \item \textbf{$K\geq 2$, bounded deviation.}
    The higher-order terms $\{\mathbf{m}_t^{(k)}\}_{k=2}^K$ add curvature-aligned corrections along the flow direction at the price of a controlled endpoint deviation:
    \begin{equation}
    \mathrm{TV}(\tilde p_1,p_1)\;\leq\; \sqrt{\tfrac{1}{2}\,\mathcal{E}_{\mathrm{score}}} \;+\; O(\Delta t),
    \qquad
    \mathcal{E}_{\mathrm{score}} \;=\; \tfrac{1}{2}\!\int_0^1\!\gamma_t^1\,\mathbb{E}_{\tilde p_t}\|\hat{s}_t-\hat{s}_t^*\|^2 dt.
    \end{equation}
    Here $\mathbb{E}_{\tilde p_t}$ is the expectation under the augmented law produced by Girsanov; at the $K=1$ baseline $\tilde p_t = p_t$ (Theorem~\ref{thm:first_order_preservation}), so $\mathcal{E}_{\mathrm{score}}$ coincides with the weighted score-matching loss of \eqref{eq:idff_loss}, and its size is $\delta$-independent by Assumption~\ref{ass:regularity}(R1$'$). Under the recommended scaling $\gamma_t^k=\alpha_k\,\gamma_t^1\,\Delta t^{k-1}$ for $k\geq 2$, the structural residual is $O(\Delta t^2)$ and the overall rate is $\mathrm{TV}(\tilde p_1,p_1)=O(\sqrt{\mathcal{E}_{\mathrm{score}}})+O(\Delta t)$ (Corollary~\ref{cor:nfe_convergence}). We do \emph{not} claim exact endpoint preservation for $K\geq 2$ at finite NFE.

    \item \textbf{Continuity equation.}
    The probability density evolves under the Fokker--Planck equation associated with \eqref{eq:idff_sde},
    \begin{equation}
    \frac{\partial \tilde{p}_t}{\partial t} \;=\; -\nabla \cdot \big(\tilde{\mathbf{v}}_t\,\tilde{p}_t\big) + \frac{g_t^2}{2}\,\Delta\tilde{p}_t,
    \end{equation}
    valid for all $K\geq 0$ with $\tilde{\mathbf{v}}_t$ given by \eqref{eq:kth_order_idff}. At $K=0$ this reduces to the CFM continuity equation; at $K=1$ with $g_t^2=2\gamma_t^1$ and the optimal score, the score-coupled drift and Laplacian cancel exactly (proof of Theorem~\ref{thm:first_order_preservation}); at $K\geq 2$, the additional drift terms $\sum_{k\geq 2}\gamma_t^k\mathbf{m}_t^{(k)}$ contribute the structural residual quantified in Theorem~\ref{thm:higher_order_deviation}.

    \item \textbf{Endpoint behaviour.}
    For all $K \geq 0$, $\tilde p_0 = p_0$ exactly (matching initial condition). At $t=1$: (i) at $K=0$, $\tilde p_1 = p_1$ exactly since $\tilde{\mathbf{v}}_t=\mathbf{v}_t$ (CFM); (ii) at $K=1$ with the default schedule and optimal score, $\tilde p_1=p_1$ exactly in continuous time (Theorem~\ref{thm:first_order_preservation}); (iii) at $K=1$ with the convex-constrained schedule, $\mathrm{TV}(\tilde p_1,p_1)$ is bounded by the closed-form expression in Proposition~\ref{prop:K1_convex_deviation}; (iv) at $K\geq 2$, $\tilde p_1 \to p_1$ in TV as $\mathcal{E}_{\mathrm{score}}\to 0$ and $\Delta t\to 0$, i.e.\ as $\mathrm{NFE}\to\infty$ for a converged score network (Theorem~\ref{thm:higher_order_deviation}, Corollary~\ref{cor:nfe_convergence}).
\end{enumerate}
\end{theorem}

Although stated for general $K$, we deploy only $K\leq 2$; see Appendix~\ref{sec:high_dim_robustness} for the empirical regularity failures that make $K\geq 3$ impractical.

The boundary conditions ensure that $\tilde{\mathbf{v}}_t(\mathbf{x}_t)$ converges to $\mathbf{v}_t(\mathbf{x}_t)$ at $t = 0, 1$, preserving consistency at the endpoints. A complete proof is provided in Appendix~\ref{app:corrected_theory}.

\begin{corollary}[Reduction to Standard CFM]
When $K=0$ and $\gamma_t^0 = 1$, Theorem~\ref{thm:kth_order_idff} reduces to standard CFM where $\tilde{\mathbf{v}}_t = \mathbf{v}_t$.
\end{corollary}

\begin{corollary}[Connection to Score-Based Models and Stochastic Interpolants]
\label{cor:idff_langevin_albergo}
The default $K=1$ IDFF SDE with $\gamma_t^0\equiv 1$, $g_t^2=2\gamma_t^1$, and the optimal score is a Langevin-enhanced probability-flow SDE: at every fixed $t$, the additional drift $\gamma_t^1\nabla\log p_t$ and the compensating diffusion $g_t^2=2\gamma_t^1$ form the canonical score-Langevin pair that preserves the instantaneous marginal $p_t$ \citep{song2020score, albergo2025stochastic}. The construction reduces exactly to the diffusion-Langevin coupling of \citet{albergo2025stochastic} when $\mathbf v_t$ is the OT-CFM probability-flow field; what is new here is not the SDE but the unified Girsanov$\to$Pinsker error budget for its learned, higher-order extension (Theorem~\ref{thm:higher_order_deviation}).
\end{corollary}

\paragraph{Regularity.}
Every statement above invokes Assumption~\ref{ass:regularity}, stated in Appendix~\ref{app:regularity}, whose hypotheses live on the \emph{truncated} interval $[0,1-\delta]$. Truncation is unavoidable: under $\sigma_t=\sigma_0\sqrt{t(1-t)}$ both $\mathbf{v}_t=(\mathbf{x}_1-\mathbf{x}_t)/(1-t)$ and $\nabla\log p_t$ diverge as $t\to 1$, so no uniform-in-$\mathbf{x}$ bound can hold to the endpoint. It also matches what the sampler already does: we integrate to $t_{\mathrm{stop}}=1-\delta$ with $\delta=\Delta t/2$ and read off $\mathbf{x}_1\approx\hat{\mathbf{x}}_1(\mathbf{x}_{t_{\mathrm{stop}}},t_{\mathrm{stop}})$ through the denoiser head. The bounds depend on $\delta$ only through the schedule integrals $\int_0^{1-\delta}\gamma_t^1\,dt$, which remain finite as $\delta\to 0$, and the size of $\mathcal{E}_{\mathrm{score}}$ is $\delta$-independent (Remark~\ref{rem:R1_R2_dependence}).

\subsection{Sample-Based Vector Field Learning}

A key design choice in IDFF is the reparameterization of vector field learning through direct sample prediction. Rather than directly learning $\mathbf{v}_t(\mathbf{x}_t)$, we predict the target sample $\mathbf{x}_1$ and use this to construct the velocity field.

\paragraph{Defining probability paths.}
Following OT-CFM, we define probability paths between the base distribution $(\mathbf{x}_0 \sim \mathcal{N}(\boldsymbol{0}, \mathbf{I}))$ and data samples $(\mathbf{x}_1 \sim p_{\text{data}})$ via
\begin{equation}
\label{eq:ppath}
p_t(\mathbf{x}_t | \mathbf{x}_1) = \mathcal{N}\left(\mathbf{x}_t \big| \mu_t; \sigma_t^2\mathbf{I}\right), \quad \mu_t = t\mathbf{x}_1 + (1-t)\mathbf{x}_0, \quad \sigma_t = \sigma_0\sqrt{t(1-t)}.
\end{equation}
This path induces the conditional velocity field
\begin{equation}
\label{eq:true-vector-field}
\mathbf{v}_t(\mathbf{x}_t | \mathbf{x}_1) = \frac{\mathbf{x}_1 - \mathbf{x}_t}{1-t}.
\end{equation}

\paragraph{Vector field reparameterization.}
Near $t=1$, the velocity $\mathbf{v}_t = (\mathbf{x}_1 - \mathbf{x}_t)/(1-t)$ exhibits a singularity as $(1-t) \to 0$. To avoid this numerical instability, we learn $\hat{\mathbf{x}}_1(\mathbf{x}_t, t; \theta)$ to predict the clean target sample, then construct
\begin{equation}
\label{eq:vector-idff-paths}
\hat{\mathbf{v}}_t(\mathbf{x}_t;\theta) = \frac{\hat{\mathbf{x}}_1(\mathbf{x}_t,t;\theta)-\mathbf{x}_t}{1-t}.
\end{equation}

\paragraph{Training objective.}
Drawing $\mathbf{x}_1 \sim p_{\mathrm{data}}$, $t \sim \mathcal{U}(0,1)$, $\mathbf{x}_0 \sim \mathcal{N}(\mathbf{0}, \mathbf{I})$, and $\boldsymbol{\epsilon} \sim \mathcal{N}(\mathbf{0}, \mathbf{I})$, we construct $\mathbf{x}_t = \mu_t + \sigma_t\boldsymbol{\epsilon}$ and minimize
\begin{equation}
\label{eq:idff_loss}
\mathcal{L}_{\text{IDFF}}(\theta) = \mathbb{E}_{t,\mathbf{x}_0,\mathbf{x}_1,\boldsymbol{\epsilon}}\left[
  \beta(t)^2 \|\hat{\mathbf{x}}_1(\mathbf{x}_t,t;\theta)-\mathbf{x}_1\|^2
  + \lambda_1(t)^2 \left\|\hat{s}_t(\mathbf{x}_t;\theta) + \frac{\boldsymbol{\epsilon}}{\sigma_t}\right\|^2
\right],
\end{equation}
where $\beta(t) = (1-t+\epsilon)^{-1}$ and $\lambda_1(t) = \sigma_t$. The first term trains the denoiser $\hat{\mathbf{x}}_1$ to recover clean samples; the second term regresses the score head $\hat{s}_t$ onto the conditional score $\nabla \log p_t(\mathbf{x}_t|\mathbf{x}_0,\mathbf{x}_1)=-\boldsymbol{\epsilon}/\sigma_t$. By Fisher's identity, $\hat{s}_t^* = \nabla \log p_t(\mathbf{x}_t)$ at the optimum. Algorithm~\ref{alg:idff_train} in Appendix~\ref{app:static_training} summarizes the training procedure.

Under a one-sided diffusion path $\mathbf{x}_t=\alpha_t\mathbf{x}_1+\sigma_t\boldsymbol{\epsilon}$, the score and the denoiser are related by an exact algebraic identity, and only one head is needed \citep{gao2025diffusion}. The same is true for the two-sided interpolant \eqref{eq:ppath} \emph{provided} the coupling is independent, since then $\mathbf{x}_t\,|\,\mathbf{x}_1\sim\mathcal{N}(t\mathbf{x}_1,\,((1-t)^2+\sigma_t^2)\mathbf{I})$ and $\nabla\log p_t(\mathbf{x}_t) = -\bigl(\mathbf{x}_t - t\,\mathbb{E}[\mathbf{x}_1|\mathbf{x}_t]\bigr)/\bigl((1-t)^2+\sigma_t^2\bigr)$. We nevertheless keep a separate head for three reasons. First, that identity fails for the minibatch-OT coupling used in the OT-IDFF variant of Figure~\ref{fig:ablation}(B), where $\mathbf{x}_0$ and $\mathbf{x}_1$ are dependent and $p_t(\cdot|\mathbf{x}_1)$ is no longer the displayed Gaussian. Second, the conversion factor $((1-t)^2+\sigma_t^2)^{-1}$ diverges as $t\to1$, precisely where the sampler is most sensitive, so deriving the score from the denoiser amplifies denoiser error exactly where the Langevin term is active. Third, the cost is small and demonstrably not doing hidden work: the joint head adds $\sim6\%$ parameters, and under a deterministic sampler that extra capacity is inert (11.87 vs.\ 11.62 in Table~\ref{tab:ablation-langevin-decomp}). We regard the single-head variant as a reasonable alternative under the independent coupling and did not find it necessary.

\subsection{Sampling with IDFF}
\label{sec:sampling}
\begin{wraptable}[21]{r}{0.5\textwidth}
\begin{minipage}{0.5\textwidth}
\vspace{-2em}
\begin{algorithm}[H]
\caption{IDFF Sampling with Higher-Order Score Derivatives}
\label{alg:idff_sample}
\begin{algorithmic}
\State \textbf{Input:} Learned networks $\hat{\mathbf{x}}_1, \hat{s}_t$, order $K$, NFE, schedules $\{\gamma_t^k\}_{k=0}^K$
\State \textbf{Initialize:} $\mathbf{x}_0 \sim \mathcal{N}(\mathbf{0}, \mathbf{I})$
\For{$i = 0$ to $\text{NFE}-1$}
    \State $t \leftarrow i/\text{NFE}$, $\Delta t \leftarrow 1/\text{NFE}$, $g_t \leftarrow \sqrt{2\gamma_t^1}$
    \State $\hat{\mathbf{v}}_t \leftarrow (\hat{\mathbf{x}}_1(\mathbf{x}_t, t) - \mathbf{x}_t)/(1-t)$, \quad $\mathbf{u}_t \leftarrow \hat{\mathbf{v}}_t / \|\hat{\mathbf{v}}_t\|$
    \State $\mathbf{m}^{(1)} \leftarrow \hat{s}_t(\mathbf{x}_t)$ \hfill \textcolor{gray}{// First-order: score}
    \For{$k = 2$ to $K$}
        \State $\mathbf{m}^{(k)} \leftarrow (\nabla_{\mathbf{x}}^{k-1} \hat{s}_t)[\mathbf{u}_t]^{k-1}$ \hfill \textcolor{gray}{// Higher-order via autodiff}
    \EndFor
    \State $\tilde{\mathbf{v}}_t \leftarrow \gamma_t^0 \hat{\mathbf{v}}_t + \gamma_t^1 \mathbf{m}^{(1)} + \sum_{k=2}^K \gamma_t^k \mathbf{m}^{(k)}$ \hfill \textcolor{gray}{// SDE drift}
    \State $\boldsymbol{\eta} \sim \mathcal{N}(\mathbf{0}, \mathbf{I})$
    \State $\mathbf{x}_t \leftarrow \mathbf{x}_t + \tilde{\mathbf{v}}_t\,\Delta t + g_t \sqrt{\Delta t} \cdot \boldsymbol{\eta}$ \hfill \textcolor{gray}{// Euler--Maruyama step}
\EndFor
\State \textbf{Return:} $\hat{\mathbf{x}}_1(\mathbf{x}_{t_{\mathrm{stop}}}, t_{\mathrm{stop}})$
\end{algorithmic}
\end{algorithm}
\end{minipage}
\end{wraptable}
After training, we generate samples by integrating the IDFF SDE \eqref{eq:idff_sde} from Theorem~\ref{thm:kth_order_idff}. For $K$-th order IDFF, the sampling procedure computes higher-order score derivatives via automatic differentiation of the learned score:
\begin{equation}
\mathbf{m}^{(1)} = \hat{s}_t(\mathbf{x}_t), \quad \mathbf{m}^{(k)} = \bigl(\nabla_{\mathbf{x}}^{k-1} \hat{s}_t(\mathbf{x}_t)\bigr)[\mathbf{u}_t]^{k-1} \text{for } k \geq 2,
\end{equation}
where $\mathbf{u}_t = \hat{\mathbf{v}}_t / \|\hat{\mathbf{v}}_t\|$. The SDE drift assembled from these terms is
\begin{equation}
\tilde{\mathbf{v}}_t \;=\; \gamma_t^0\,\hat{\mathbf{v}}_t \;+\; \gamma_t^1\,\mathbf{m}^{(1)} \;+\; \sum_{k=2}^K \gamma_t^k\,\mathbf{m}^{(k)},
\label{eq:sampling_drift}
\end{equation}
where $\hat{\mathbf{v}}_t = (\hat{\mathbf{x}}_1 - \mathbf{x}_t)/(1-t)$ and $g_t^2 = 2\gamma_t^1$ is the diffusion-compensation choice from Theorem~\ref{thm:first_order_preservation}.

We discretize with time step $\Delta t = 1/\mathrm{NFE}$ and evolve samples through the Euler--Maruyama scheme
\begin{equation}
\mathbf{x}_{t+\Delta t} = \mathbf{x}_t + \tilde{\mathbf{v}}_t(\mathbf{x}_t)\,\Delta t + g_t \sqrt{\Delta t}\,\boldsymbol{\eta}, \quad \boldsymbol{\eta} \sim \mathcal{N}(0, \mathbf{I}).
\label{eq:euler_maruyama}
\end{equation}

Starting from $\mathbf{x}_0 \sim \mathcal{N}(0,\mathbf{I})$, we integrate forward to $t_{\mathrm{stop}}=1-\Delta t/2$ and read off the sample through the denoiser head. Under the default schedule $\gamma_t^0\equiv 1$, $g_t^2=2\gamma_t^1$, both the score head $\mathbf{m}^{(1)}=\hat{s}_t$ \emph{and} the Brownian increment $g_t\sqrt{\Delta t}\,\boldsymbol{\eta}$ enter the discrete update: the score-coupled drift contribution $\gamma_t^1\hat{s}_t\,\Delta t$ is precisely the Langevin term that cancels the Laplacian half of the SDE's Fokker--Planck equation in the proof of Theorem~\ref{thm:first_order_preservation}, leaving the CFM continuity equation exactly. The score head is already evaluated to form $\mathbf{m}^{(1)}$, so including it in the drift adds no extra forward pass; only $K\geq 2$ introduces Jacobian-vector products through autodiff of $\hat{s}_t$.

The score-Langevin drift makes the discretized sampler tolerant of coarse steps for the reason discussed in Section~\ref{sec:why_nfe}, allowing accurate sampling with significantly fewer NFEs than standard CFM; as the ablation of Table~\ref{tab:ablation-langevin-decomp} shows, on images this is entirely a $K=1$ effect. The accompanying preservation guarantee is the two-regime statement of Theorem~\ref{thm:kth_order_idff}: at $K=1$ with the default schedule, the marginal $p_t$ is preserved \emph{exactly} for all $t\in[0,1]$ in continuous time; at $K\geq 2$, the endpoint deviation is controlled by the quantitative Girsanov$\to$Pinsker bound, not exactly zero at finite NFE. The contraction with $\mathbf{u}_t$ reduces variance compared to full tensor computation; ensuring stable $\nabla_{\mathbf{x}}^{k-1} \hat{s}_t$ in high dimensions requires smooth activations (e.g., GELU). We deploy $K \leq 2$ throughout: at $K=3$ we observe Jacobian-norm spikes that diverge in roughly 2\% of batches (Appendix~\ref{sec:high_dim_robustness}), and we report this as a limitation rather than a configuration we recommend.

\paragraph{Time-series extension.}
For sequences $\mathbf{x}_{1:N}$, we use dual-time indexing: discrete $n \in \{1,\ldots,N\}$ for sequence position and continuous $t \in [0,1]$ for flow. Networks become $\hat{\mathbf{x}}_1(\mathbf{x}_t, t, n; \theta)$ and $\hat{s}_t(\mathbf{x}_t, t, n; \theta)$. For $n > 1$, we initialize $\mathbf{x}_0^n \sim \mathcal{N}(\mathbf{x}_1^{n-1}, \sigma_0^2\mathbf{I})$, creating Markovian temporal coherence. When $N=1$, this reduces to the standard static formulation. Details are provided in Appendix~\ref{app:algorithms}.

\section{Related Work}
\label{sec:related}

\textbf{Conditional Flow Matching.}
Flow Matching~\citep{lipman2022flow} revived interest in continuous-time generative modeling by removing expensive simulation-based training and introducing optimal-transport-shaped conditional paths; OT-CFM and $[\mathrm{SF}]^2\mathrm{M}$~\citep{tong2023improving} improve scalability through minibatch OT approximations, stochastic interpolants~\citep{albergo2025stochastic} give a general framework covering both deterministic and stochastic transport, and Rectified Flow~\citep{liu2022flow} learns straight trajectories via iterative reflow. All of these methods need $\geq 100$ NFEs for high-quality generation. IDFF departs by augmenting the vector field with score-derivative terms $\boldsymbol{\xi}_t = \sum_{k=1}^K \gamma_t^k \mathbf{m}_t^{(k)}$ and a matched diffusion compensation. Diffusion and flow matching are not distinct paradigms here: the score, noise, denoiser and velocity parameterizations are interchangeable up to standard identities \citep{gao2025diffusion}. The result is $\leq 10$ NFEs with two-regime guarantees: \emph{exact} continuous-time preservation of the CFM marginal $p_t$ at $K=1$, and a single quantitative Girsanov$\to$Pinsker TV bound with asymptotic endpoint convergence at $K\geq 2$.

\textbf{Efficient sampling and recent advances.}
NFE reduction has been pursued through multiple strategies: DDIM~\citep{song2020denoising} converts stochastic diffusion to deterministic flows; high-order solvers (DPM-Solver~\citep{lu2022dpm}, DPM-Solver++~\citep{lu2022dpm2}) leverage analytical score dynamics; consistency models~\citep{song2023consistency} learn direct endpoint mappings; and distillation~\citep{salimans2022progressive} compresses many-step sampling. Recent advances include Consistency-FM~\citep{yang2024consistency}, which extends consistency training to flow matching, InstaFlow~\citep{liu2023instaflow}, which straightens trajectories in a single distillation round, and schedule optimization~\citep{sabour2024align}, which learns optimal time discretizations without retraining. Unlike consistency models, which sacrifice the continuous probability path for direct endpoint mappings, IDFF maintains the full flow structure, enabling density evaluation and interpolation at intermediate times. IDFF is complementary to these approaches: our score-derivative augmentation can combine with optimized schedules or consistency training for further gains (Table~\ref{tab:recent-fm-main}).

\textbf{Stochastic samplers and correctors.}
The $K=1$ IDFF sampler belongs to the family of stochastic samplers that add a $p_t$-preserving Langevin component to a deterministic transport: predictor--corrector sampling~\citep{song2020score}, the churn parameter of EDM~\citep{karras2022elucidating}, and the diffusion-Langevin coupling of stochastic interpolants~\citep{albergo2025stochastic} all share this structure, and Section~\ref{sec:why_nfe} gives the mechanism they have in common. What IDFF contributes relative to these is the specific pairing $(\gamma_t^1, g_t^2=2\gamma_t^1)$ with a schedule tied to the CFM path, a closed-form error budget that extends to learned scores and to curvature corrections, and the ablation of Table~\ref{tab:ablation-langevin-decomp} that separates the Langevin pair from generic stochasticity. We note as a limitation (Section~\ref{sec:limitations}) that we do not benchmark a tuned corrector of the predictor--corrector or EDM-churn type on our own trained score head; the closest control we run is the ``Stochastic OT-CFM'' row, which matches the diffusion but omits the score drift.

The term ``higher order'' is used in three unrelated ways in this literature, and we organise the comparison accordingly.

\emph{(i) Higher spatial score-derivative order.} \citet{lu2022maximum} train separate networks for the second- and third-order score derivatives in order to obtain maximum-likelihood training of probability-flow ODEs, and \citet{meng2021estimating} study the estimation of $\nabla^k\log p_t$ by denoising directly, establishing the closed-form conditional targets that any such method must regress against. GENIE~\citep{dockhorn2022genie} distils a higher-order denoising solver, training an auxiliary head that supplies the curvature term used by its truncated Taylor step. IDFF sits in this category but avoids the auxiliary networks: by Theorem~\ref{thm:kth_order_marginal}, $\nabla^k \log p_t = \nabla_{\mathbf{x}}^{k-1} \hat{s}_t^*$, so all higher-order marginals are already encoded in the first-order score and are extracted at sampling time via autodiff. The contracted form $\mathbf{m}_t^{(k)} = (\nabla_{\mathbf{x}}^{k-1} \hat{s}_t)[\mathbf{u}_t]^{k-1}$ keeps the dynamics in data space. We do not run a matched-budget head-to-head against GENIE: doing so would require training its auxiliary higher-order networks from scratch on our backbone, which is beyond the compute we have. Our NFE-reduction claim is a $K=1$ claim (Section~\ref{sec:img-gen}), and the $K\geq 2$ theory is a deviation bound, so the comparison is not load-bearing for any claim we make; we list it as a limitation rather than treat it as settled.

\emph{(ii) Higher numerical integration order.} Classical Runge--Kutta schemes achieve order-$p$ accuracy with local error $O(\Delta t^{p+1})$ by evaluating the vector field several times per step; DPM-Solver and its successors~\citep{lu2022dpm, lu2022dpm2, zheng2023dpm} apply the same principle to score-based models using the semi-linear structure of the probability-flow ODE. This sense of ``order'' counts \emph{network evaluations}, and is orthogonal to ours: IDFF $K=1$ uses one evaluation per step and $K=2$ adds a single JVP, so IDFF changes the dynamics being integrated rather than integrating a fixed ODE more accurately.
\begin{wrapfigure}[24]{r}{0.5\textwidth}
    \centering
\includegraphics[width=\linewidth]{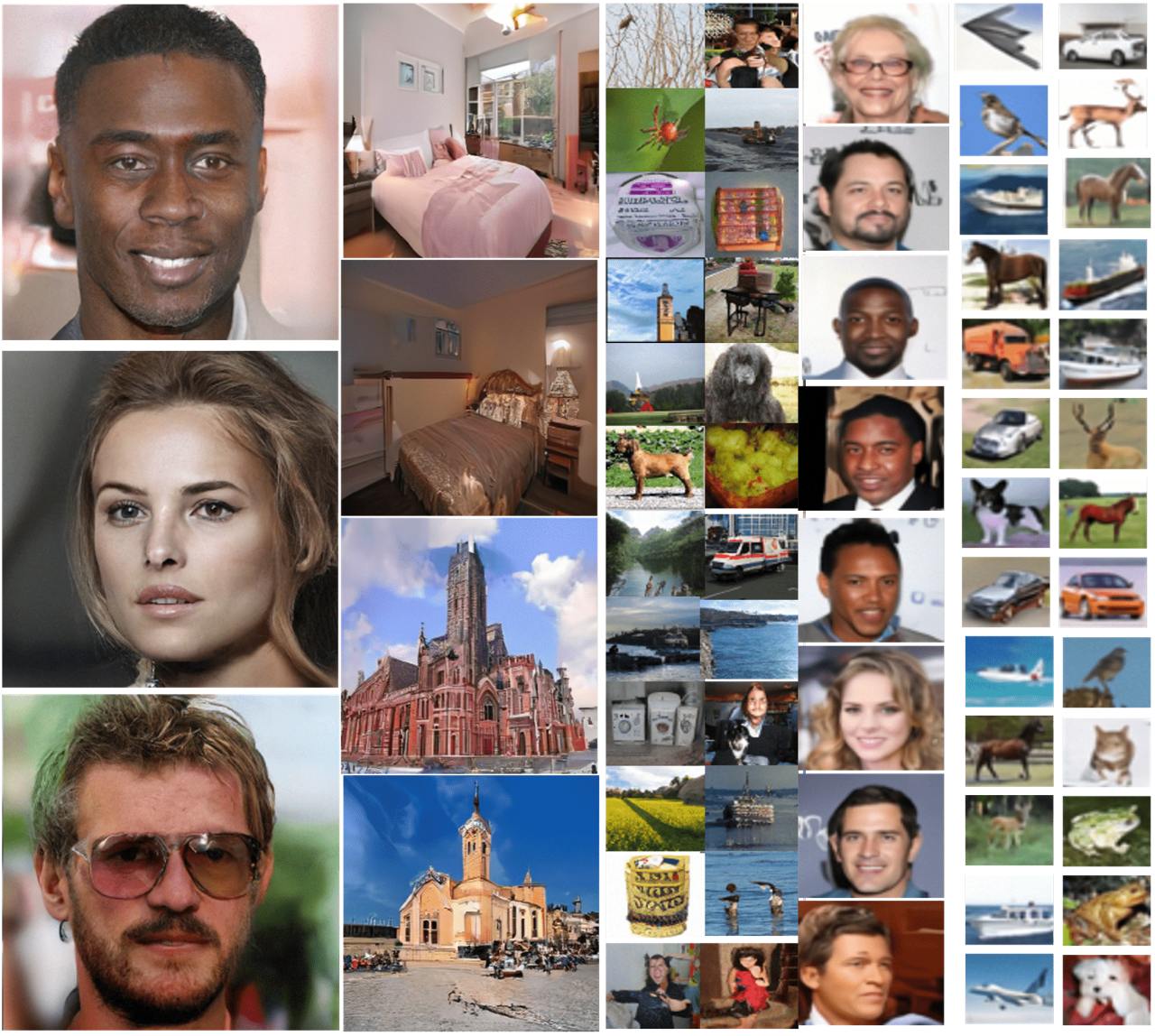}
\caption{Image generation using IDFF across datasets. Additional samples and analysis are provided in Appendix~\ref{app:aditional-result-image}.}
\label{fig:img-gen-iff}
\end{wrapfigure}
\emph{(iii) Higher-order state dynamics.} Critically-damped Langevin diffusion~\citep{dockhorn2022cld}, Hamiltonian flows~\citep{holderrieth2024hamiltonian}, and TADA~\citep{chen2025tada} augment the state with an auxiliary velocity (momentum) variable and run the dynamics in an extended phase space. IDFF introduces no auxiliary momentum state and operates entirely in data space; what we add is a drift term built from spatial derivatives of the score. For this reason we avoid the word ``momentum'' for our $\mathbf{m}_t^{(k)}$ terms and call them contracted score-derivative terms throughout, and the comparison to genuine momentum methods is one of contrast rather than of degree.

IDFF combines three ingredients. (1) Score-Langevin augmentation of the CFM drift (a known marginal-preserving construction) extended to higher orders through contracted tensor operations that stay in data space. (2) Higher-order marginals computed from a single first-order score via autodiff, with no auxiliary networks to train; the identity behind this is elementary, but the consequence for the training pipeline is not shared by \citet{lu2022maximum} or \citet{dockhorn2022genie}. (3) A cleanly separated theoretical picture: \emph{exact} preservation $\tilde p_t = p_t$ for $K=1$ in continuous time under the default Langevin-enhanced schedule (with the convex-constrained alternative bounded in closed form by a Girsanov$\to$Pinsker step), and the quantitative bound $\mathrm{TV}(\tilde p_1,p_1)\leq\sqrt{\tfrac{1}{2}\mathcal{E}_{\mathrm{score}}}+O(\Delta t)$ with asymptotic endpoint convergence for $K\geq 2$ (Theorem~\ref{thm:kth_order_idff}). The result is a method that stands on its own and is also complementary to advances in solvers, distillation, and consistency training.

\section{Experiments}
\label{sec:experiments}

\subsection{Image Generation}
\label{sec:img-gen}

We evaluate on CIFAR-10 (Table~\ref{tab:image-result-cifa10}), with additional results on CelebA, ImageNet-64, CelebA-HQ, and LSUN in Appendix~\ref{app:aditional-result-image}. Our $K=1$ configuration achieves 2.78 FID with only 10 NFEs, matching 1-Rectified Flow (2.58 FID) with $12.7\times$ fewer evaluations and substantially outperforming OT-CFM (11.87 FID at 10 NFE) and DPM-Solver-v3 (3.40 FID) at the same backbone and training budget. We report $K=1$ as the headline throughout because it is the configuration carrying the exact-preservation guarantee and it costs 52\,ms per sample against 68\,ms for $K=2$ (Table~\ref{tab:wallclock-main}); the $K=2$ number (2.65) is listed separately wherever it appears.

\paragraph{Setup parity, stated precisely.}
Three baselines were retrained by us on the ScoreSDE U-Net backbone~\citep{song2020score} under a budget matched to IDFF---700K iterations, horizontal-flip augmentation only, identical evaluation protocol (50k generated samples for FID): OT-CFM, 1-Rectified Flow, and DPM-Solver-v3 (Appendix~\ref{sec:training_fairness}, Table~\ref{tab:training-details}). Every other row of Table~\ref{tab:image-result-cifa10} is quoted from the cited paper and is \emph{not} budget- or backbone-matched to ours; the \textbf{Source} column records this per row, and Table~\ref{tab:training-details} gives the backbone, parameter count, training length and augmentation for every row so that the reader can see exactly which comparisons are matched. In particular the DPM-Solver-v3-EDM row uses the EDM backbone~\citep{karras2022elucidating}, which we did not retrain. For a comparison that is matched end to end, Table~\ref{tab:recent-fm-main} reports IDFF against two recent flow-matching accelerations (schedule-optimized FM and 2-Rectified Flow) at the same backbone and training budget.
\begin{figure}[t]
\centering
\includegraphics[width=\textwidth]{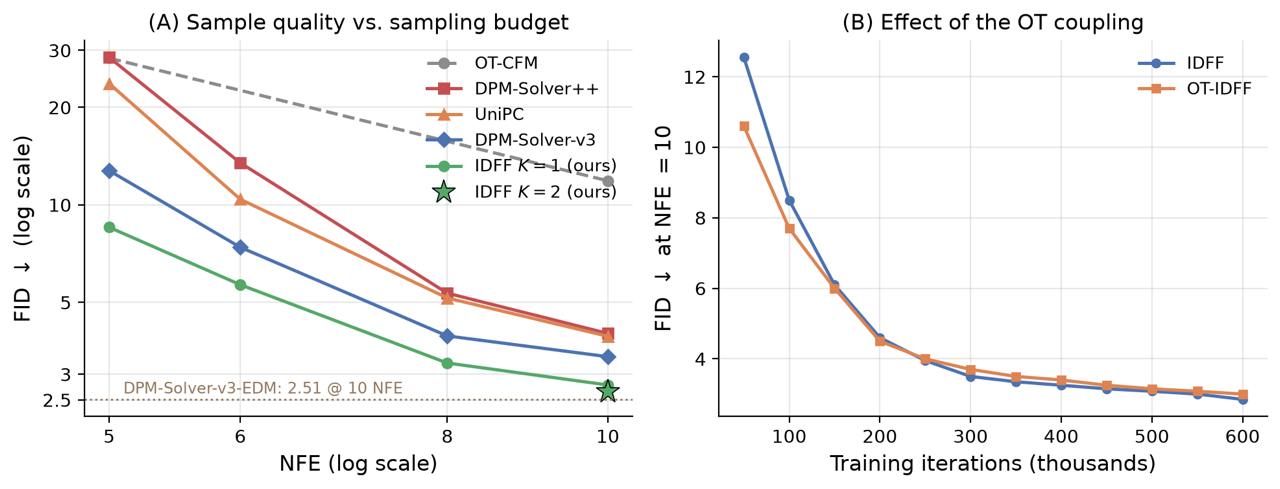}
\caption{\textbf{(A)} FID versus NFE on CIFAR-10 (log--log axes), for the samplers of Table~\ref{tab:ablation-nfe} together with OT-CFM and the two IDFF orders; every point is a value tabulated in this paper, and the star marks the single $K{=}2$ configuration we evaluated (10 NFE, FID 2.65). IDFF $K{=}1$ is below all four of these baselines at every budget in the $5$--$10$ NFE range, and the margin widens as the budget shrinks ($8.53$ vs.\ $12.76$ at 5 NFE). Methods evaluated at a single budget elsewhere in the paper (2-Rectified Flow, the distillation baselines) are not shown here; see Tables~\ref{tab:image-result-cifa10} and~\ref{tab:recent-fm-main}. The dotted line marks DPM-Solver-v3 on the EDM backbone, which we did not retrain and which remains ahead of IDFF at 10 NFE. \textbf{(B)} FID at 10 NFE as a function of the training budget, with (OT-IDFF) and without (IDFF) the minibatch-OT coupling of~\citet{tong2023improving}. The two curves coincide from roughly 200K iterations onward, so IDFF does not depend on the OT coupling for its low-NFE quality.}
\label{fig:ablation}
\end{figure}
\paragraph{Comparison to distillation and consistency methods (different regime).}
Consistency-style methods achieve impressive results at 1--2 NFEs: iCT-deep reaches 2.24 FID at 2 NFE and CTM reaches 1.87 FID at 2 NFE. CD and CTM require a separately pre-trained diffusion teacher and additional distillation training on top of it (800K iterations for CD); iCT-deep needs no teacher but uses a $2\times$-deeper NCSN++ backbone than ours (see Appendix~\ref{sec:training_fairness} for full budgets). Either way they occupy a different operating point from IDFF, which trains from scratch at the ScoreSDE backbone, and in the 1--2 NFE regime they win. At a matched ScoreSDE backbone and matched 700K-iteration training budget, IDFF attains the best FID among the non-distilled methods we evaluate \emph{at 10 NFE} (Tables~\ref{tab:image-result-cifa10}, \ref{tab:ablation-nfe}, \ref{tab:recent-fm-main}). At 5 NFE, 2-Rectified Flow reaches 7.45 against IDFF's 8.53 at the same budget (Table~\ref{tab:recent-fm-main}), and DPM-Solver-v3-EDM attains 2.51 at 10 NFE using the EDM backbone~\citep{karras2022elucidating}, which we did not retrain. We therefore do not claim to be state of the art at this budget in general; we claim the best result among the methods we were able to train under a common protocol. The approaches are complementary rather than competing: IDFF trains a standard diffusion-quality model that any of these pipelines could distil from.

Finally, for applications that need intermediate density evaluation or trajectory interpolation, IDFF with $K=1$ (default schedule $\gamma_t^0\equiv 1$, $g_t^2=2\gamma_t^1$) preserves the CFM marginal $p_t$ \emph{exactly} in continuous time on the original clock (Theorem~\ref{thm:kth_order_idff}, Theorem~\ref{thm:first_order_preservation}); a property that consistency models sacrifice by collapsing to direct endpoint mappings.

Figure~\ref{fig:ablation}(A) plots the same comparison across sampling budgets, and Figure~\ref{fig:ablation}(B) shows that IDFF maintains comparable performance without expensive OT calculations, as the score-Langevin drift alone already guides particles toward high-probability regions. Reducing NFE from 100 to 10 yields a $10\times$ speedup (50k samples in $\sim$30 minutes on an A6000). Importantly, the NFE reduction translates almost one-to-one to wall-clock speedup at $K{=}1$, because the per-step cost of IDFF $K{=}1$ matches OT-CFM (it requires only a forward pass through the joint head, with no autodiff). Table~\ref{tab:wallclock-main} summarizes end-to-end sampling cost on CIFAR-10; see Appendix~\ref{sec:computational_overhead} for the full breakdown including JVP overhead for $K{=}2$.
\begin{table}[tb]
\caption{Comparison of FID and NFE on CIFAR-10. \textbf{Source} states the provenance of each row: \emph{Reproduced} rows were retrained by us on the ScoreSDE backbone~\citep{song2020score} with the same 700K-iteration budget and augmentation as IDFF (Table~\ref{tab:training-details}); \emph{Cited} rows are quoted from the reference given and are not budget- or backbone-matched. $^\ddagger$Requires distillation from a separately pre-trained teacher; iCT-deep carries no $\ddagger$ because improved consistency \emph{training} trains from scratch. The headline IDFF configuration is $K=1$; the $K=2$ row is listed separately and is compared at matched NFE, not at matched wall-clock (Table~\ref{tab:wallclock-main}). No entry is emphasised, because the rows span three different operating points (many-step samplers, 1--2 NFE distillation, and few-step training-from-scratch); Table~\ref{tab:ablation-nfe} gives the like-for-like sweep at a common budget.}
\begin{center}
\begin{tabular}{lccl}
\toprule
Model & FID$\downarrow$ & NFE$\downarrow$ & Source \\
\midrule
\multicolumn{4}{l}{\textit{Diffusion Models + Samplers}} \\
DDIM \citep{song2020denoising} & 13.36 / 6.84 & 10 / 20 & Cited \\
DPM-Solver++ \citep{lu2022dpm2} & 4.01 & 10 & Cited \\
UniPC \citep{zhao2023unipc} & 3.93 & 10 & Cited \\
DPM-Solver-v3 \citep{zheng2023dpm} & 3.40 & 10 & Reproduced (ours) \\
DPM-Solver-v3-EDM \citep{zheng2023dpm} & 2.51 & 10 & Cited \\
\midrule
\multicolumn{4}{l}{\textit{Distillation \& Consistency Models}} \\
$^\ddagger$CD \citep{song2023consistency} & 2.93 & 2 & Cited \\
iCT-deep \citep{song2023improved} & 2.24 & 2 & Cited; trained from scratch \\
$^\ddagger$CTM \citep{kim2023consistency} & 1.98 / 1.87 & 1 / 2 & Cited \\
\midrule
\multicolumn{4}{l}{\textit{Flow Matching Models}} \\
OT-CFM \citep{tong2023improving} & 11.87 / 6.35 & 10 / 142 & Reproduced (ours) \\
1-Rectified Flow \citep{liu2022flow} & 2.58 & 127 & Reproduced (ours) \\
Consistency-FM \citep{yang2024consistency} & 5.34 & 2 & Cited \\
$^\ddagger$FGM \citep{huang2024flow} & 3.08 & 1 & Cited \\
\midrule
IDFF, $K{=}1$ (ours, headline) & 2.78 & 10 & Ours \\
IDFF, $K{=}2$ (ours) & 2.65 & 10 & Ours; $+31\%$ wall-clock \\
\bottomrule
\end{tabular}
\end{center}
\label{tab:image-result-cifa10}
\end{table}

\subsubsection{Ablations}

\begin{wraptable}[6]{r}{0.5\textwidth}
\centering
\vspace{-5em}
\caption{FID$\downarrow$ across NFE budgets on 50k samples. Baseline rows carry the same provenance as in Table~\ref{tab:image-result-cifa10}.}
\label{tab:ablation-nfe}
\begin{tabular}{lcccc}
\toprule
Method & 5 & 6 & 8 & 10 \\
\midrule
DPM-Solver++ & 28.53 & 13.48 & 5.34 & 4.01 \\
UniPC & 23.71 & 10.41 & 5.16 & 3.93 \\
DPM-Solver-v3 & 12.76 & 7.40 & 3.94 & 3.40 \\
IDFF (Ours, $K{=}1$) & \textbf{8.53} & \textbf{5.67} & \textbf{3.25} & \textbf{2.78} \\
\bottomrule
\end{tabular}
\end{wraptable}

Table~\ref{tab:ablation-nfe} shows IDFF outperforms the three diffusion samplers we compare against at every NFE level in this range, with large margins at low NFE (8.53 vs.\ 12.76 at 5 NFE). We use $K=1$ for images (exact continuous-time marginal preservation under the default schedule $\gamma_t^0\equiv 1$, $g_t^2=2\gamma_t^1$, by Theorem~\ref{thm:first_order_preservation}); for molecular dynamics we additionally evaluate $K=2$ to test whether curvature information helps on a rugged energy landscape. The choice of hyperparameters $\alpha_k$ and schedule shapes $\gamma_t^k$ significantly impacts performance. In Appendix~\ref{sec:ablation_coefficients}, we ablate:
\begin{itemize}
    \item \textbf{Second-order scaling $\alpha_2$}: Values around 0.05 give the best FID at $K{=}2$ without excessive path deviation (Table~\ref{tab:ablation-alpha}).
    \item \textbf{Diffusion compensation}: The theoretical choice $g_t^2 = 2\gamma_t^1$ yields best results, confirming the exact Fokker--Planck cancellation of Theorem~\ref{thm:first_order_preservation} (Table~\ref{tab:ablation-diffusion}).
    \item \textbf{Schedule shape}: The Beta(2,2) schedule $\gamma_t^1 \propto t(1-t)$ outperforms triangular and sine alternatives (Table~\ref{tab:ablation-schedule}).
\end{itemize}

\begin{table}[ht]
\centering
\caption{End-to-end sampling cost on CIFAR-10 (NVIDIA A6000, averaged over 1000 samples). IDFF $K{=}1$ has negligible per-step overhead vs.\ OT-CFM, so the $10\times$ NFE reduction translates directly into a $\sim 10\times$ wall-clock speedup. $K{=}2$ adds one JVP per step, costing $31\%$ more wall-clock at the same NFE; on images that buys 2.78 $\to$ 2.65 FID, and since NFE is not a matched budget across $K$ we do not present this as evidence that curvature helps on images. Boldface marks the fastest sampler and the lowest FID among the 10-NFE rows; 1-Rectified Flow's 2.58 is obtained at 127 NFE and $12.7\times$ the wall-clock.}
\label{tab:wallclock-main}
\begin{tabular}{lcccc}
\toprule
Method & NFE & Time/sample (ms) & Peak memory & FID$\downarrow$ \\
\midrule
1-Rectified Flow~\citep{liu2022flow} & 127 & 658 & 2.1\,GB & 2.58 \\
DPM-Solver-v3~\citep{zheng2023dpm}    & 10 &  92 & 2.0\,GB & 3.40 \\
OT-CFM~\citep{tong2023improving}      & 10 &  54 & 2.0\,GB & 11.87 \\
\midrule
IDFF $K{=}1$ (ours, headline)         & 10 &  \textbf{52} & 2.1\,GB & 2.78 \\
IDFF $K{=}2$ (ours)                   & 10 &  68 & 2.4\,GB & \textbf{2.65} \\
\bottomrule
\end{tabular}
\end{table}

\paragraph{Decomposing the IDFF gain:}
The IDFF $K{=}1$ sampler differs from the OT-CFM baseline in three potentially confounded ways: (a) the score-Langevin drift $\gamma_t^1\hat s_t$, (b) injected SDE noise with diffusion $g_t$, and (c) a $\sim 6\%$ larger parameter count from the joint $(\hat{\mathbf x}_1,\hat s_t)$ head. 
\begin{table}[thb]
\centering
\caption{Decomposing the IDFF $K{=}1$ gain on CIFAR-10 at NFE$=10$. Rows isolate (a) the score-Langevin drift $\gamma_t^1\hat s_t$, (b) generic SDE stochasticity with matched $g_t^2=2\gamma_t^1$ but \emph{no} score drift, and (c) extra capacity from the joint $(\hat{\mathbf x}_1,\hat s_t)$ head used at inference under the deterministic OT-CFM sampler. The headline FID 2.78 is recovered only when both the Langevin drift and the matched diffusion are present. Setup parity: same ScoreSDE backbone, 700K iterations, horizontal-flip augmentation.}
\label{tab:ablation-langevin-decomp}
\small
\begin{tabular}{lcccc}
\toprule
Variant & Score drift & Diffusion $g_t$ & Extra capacity & FID$\downarrow$ \\
\midrule
OT-CFM (baseline) & --- & 0 (ODE) & --- & 11.87 \\
OT-CFM + score head (det.\ sampler) & --- & 0 (ODE) & yes ($+\!6\%$ params) & 11.62 \\
\midrule
\multicolumn{5}{l}{\textit{Isolating stochasticity (no Langevin drift):}} \\
Stochastic OT-CFM, $g_t^2 = 2\gamma_t^1$, $\bar\gamma=0.1$ & --- & matched & --- & 7.84 \\
Stochastic OT-CFM, $g_t^2 = 2\gamma_t^1$, $\bar\gamma=0.2$ & --- & matched & --- & 6.41 \\
Stochastic OT-CFM + score head, $g_t^2 = 2\gamma_t^1$ & --- & matched & yes ($+\!6\%$ params) & 6.18 \\
\midrule
\multicolumn{5}{l}{\textit{Adding the score-Langevin drift:}} \\
IDFF $K{=}1$, $g_t = 0$ (ODE, no compensation) & $\gamma_t^1\hat s_t$ & 0 & yes ($+\!6\%$ params) & 9.74 \\
IDFF $K{=}1$, $g_t^2 \neq 2\gamma_t^1$ (mismatched) & $\gamma_t^1\hat s_t$ & mismatched & yes ($+\!6\%$ params) & 3.45 \\
\textbf{IDFF $K{=}1$ (ours), $g_t^2 = 2\gamma_t^1$} & \boldmath$\gamma_t^1\hat s_t$ & matched & yes ($+\!6\%$ params) & \textbf{2.78} \\
\bottomrule
\end{tabular}
\end{table}
Table~\ref{tab:ablation-langevin-decomp} disentangles them on CIFAR-10 (10 NFE, identical 700K-iteration training budget, ScoreSDE backbone). All ``Stochastic OT-CFM'' rows use the same $g_t^2=2\gamma_t^1$ Brownian increment as IDFF $K{=}1$, but with the score-Langevin drift \emph{removed} from the drift, so they test stochasticity in isolation. The ``OT-CFM + score head'' rows train the joint head (same parameter count as IDFF) but use the deterministic OT-CFM sampler at inference time, isolating the effect of extra capacity.
Three things follow. Swapping OT-CFM for an SDE with IDFF's diffusion schedule but no Langevin drift leaves $\sim$40\% of the gap to IDFF open (11.87 $\to$ 6.41, against 2.78), so stochasticity alone is not the explanation. The extra capacity of the joint head is inert under a deterministic sampler (11.87 vs.\ 11.62), so the gain is not a capacity artefact. And the Langevin drift \emph{without} its matched diffusion either degrades FID (ODE limit, 9.74) or sits well above the matched setting (mismatched $g_t$, 3.45). The headline gain is therefore attributable to the score-Langevin drift paired with its matched diffusion compensation---exactly the configuration Theorem~\ref{thm:first_order_preservation} singles out, and exactly the pairing whose contraction property Section~\ref{sec:why_nfe} identifies as the mechanism.

IDFF with $K \geq 2$ requires smooth activations (SiLU/GELU) for well-defined Jacobians; standard ReLU is incompatible. We monitor Jacobian norms during sampling and find they remain bounded throughout (see Appendix~\ref{sec:architecture_stability} for statistics and stability analysis). Score estimation errors propagate with approximately $2\times$ amplification to second-order terms (Table~\ref{tab:error-propagation}), which motivates our conservative choice of $K \leq 2$.

\subsection{Surrogate Models for Molecular Dynamics}
\label{sec:md-sim}

\begin{wraptable}[11]{r}{0.5\textwidth}
\vspace{-4em}
\caption{MAE, RMSE, and CC results for the MD simulation, at a matched NFE budget of 10.}
\label{tab:md-simulation}
\begin{center}
\begin{tabular}{lccc}
\toprule
Method & MAE$\downarrow$ & RMSE$\downarrow$ & CC (\%)$\uparrow$ \\
\midrule
SRNN & 82.6$_{\pm28}$ & 91.9$_{\pm25}$ & 10.2$_{\pm0.27}$ \\
DVAE & 78.1$_{\pm27}$ & 88.1$_{\pm25}$ & 30.4$_{\pm0.35}$ \\
NODE & 25.3$_{\pm6.3}$ & 28.8$_{\pm6.2}$ & 10.5$_{\pm0.41}$ \\
\midrule
OT-CFM & 13.3$_{\pm1.1}$ & 16.3$_{\pm2.4}$ & 86.1$_{\pm0.1}$ \\
TFM & 11.4$_{\pm1.1}$ & 14.2$_{\pm2.7}$ & 89.4$_{\pm0.4}$ \\
IDFF ($K=1$) & 9.2$_{\pm0.9}$ & 12.5$_{\pm2.8}$ & 95.6$_{\pm0.1}$ \\
IDFF ($K=2$) & \textbf{4.9$_{\pm1.1}$} & \textbf{9.7$_{\pm2.8}$} & \textbf{97.8$_{\pm0.1}$} \\
\bottomrule
\end{tabular}
\end{center}
\end{wraptable}
Molecular dynamics (MD) simulations are essential for drug discovery and materials science but remain computationally prohibitive. Simulating even microseconds of protein dynamics can require weeks of computation~\citep{marx2009ab, car1985unified}. Learning surrogate models that can generate realistic trajectories offers a path to accelerate scientific discovery~\citep{shaw2008anton}. We evaluate IDFF's ability to learn complex molecular dynamics from limited simulation data.

We employ traditional physics-based MD simulations using the AMBER force field to generate training data. The system consists of a fully extended polyalanine structure containing 253 atoms and 46 dihedral angles, simulated for 400 picoseconds at 300K in vacuum conditions with dihedral angles recorded at regular intervals.

IDFF learns to model the generative distribution of these MD trajectories, functioning as an efficient surrogate model for the underlying dynamical system rather than performing computationally expensive ab initio calculations.

\begin{wrapfigure}[27]{r}{0.5\textwidth}
\centering
\includegraphics[width=\linewidth]{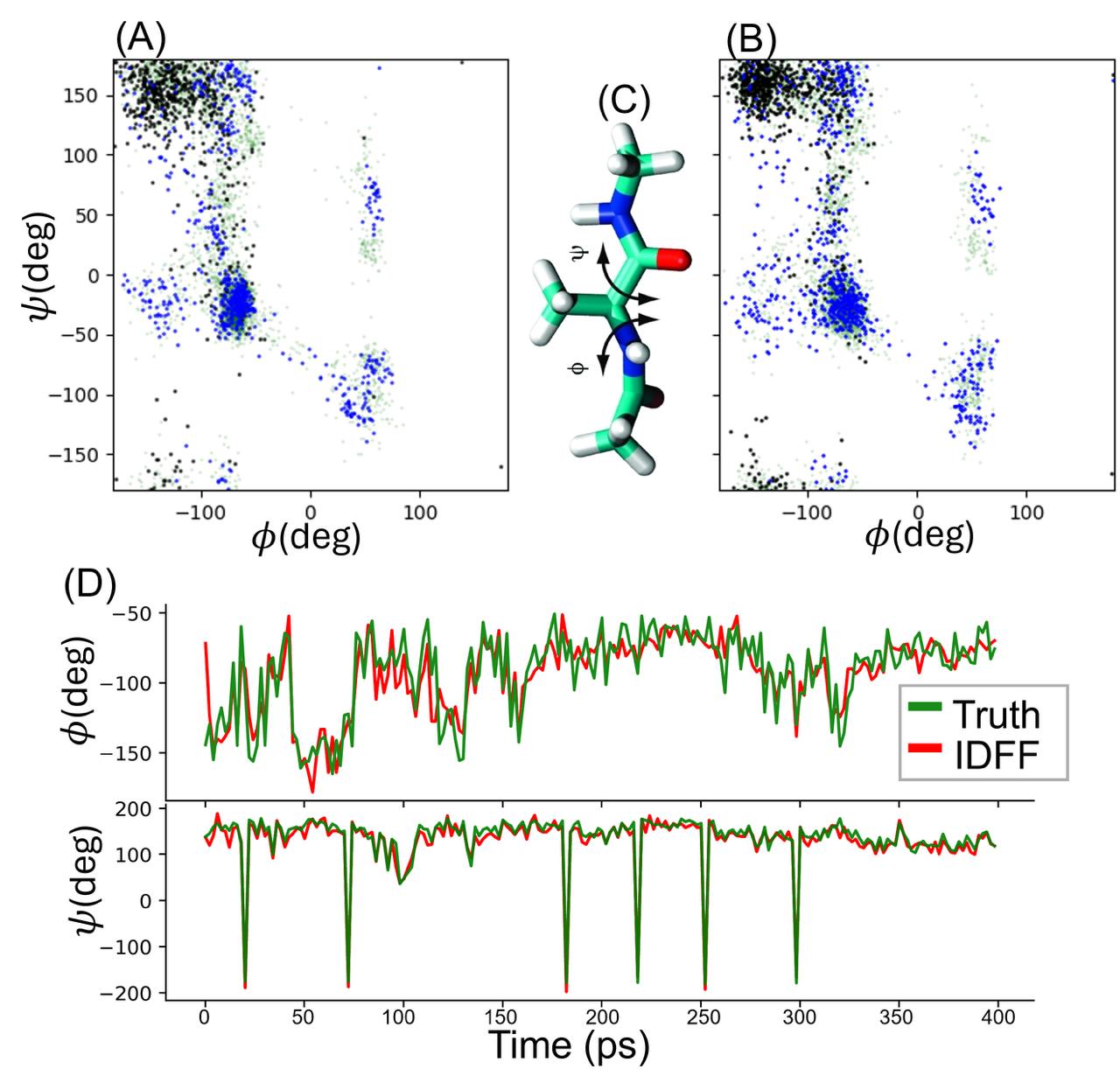}
\caption{(A) True and (B) generated dihedral angle distributions. (C) Dihedral angles for an alanine molecule. (D) True and generated angle trajectories by IDFF ($K=1$).}
\label{fig:md-sim-iff}
\end{wrapfigure}

IDFF demonstrates strong precision in predicting dynamics for complex molecular structures. Figure~\ref{fig:md-sim-iff} illustrates the model's performance, presenting distributions of actual (A) and generated (B) dihedral angles. Panel (C) shows the dihedral angles for a single alanine molecule, while panel (D) provides a trajectory comparison between actual and generated angles over time. We assess performance using three metrics: root mean squared error (RMSE), mean absolute error (MAE), and correlation coefficient (CC) between generated and actual trajectories. When benchmarked against established dynamical models including Sequential Recurrent Neural Networks (SRNN), Variational Autoencoders (DVAE), and Neural Ordinary Differential Equations (NODE), the flow-based methods OT-CFM, TFM, and IDFF perform better across all metrics (Table~\ref{tab:md-simulation}). At a matched NFE budget, IDFF with $K=2$ achieves the best results. The contracted Hessian-vector product $\mathbf{m}^{(2)} = (\nabla_{\mathbf{x}} \hat{s}_t)\mathbf{u}_t$ captures directional curvature along the flow, which appears to be worth its cost on the complex energy landscapes of molecular systems; the wall-clock caveat is discussed below.
At a matched NFE budget, $K=2$ improves on $K=1$ (MAE: 9.2 $\to$ 4.9, CC: 95.6\% $\to$ 97.8\%), the largest $K=2$ gain we observe in any setting. While $K=2$ introduces a bounded path deviation (Theorem~\ref{thm:higher_order_deviation}), the terminal deviation $\mathrm{TV}(\tilde{p}_1, p_1)$ is controlled by the closed-form Girsanov$\to$Pinsker bound, and vanishes as NFE $\to \infty$ \emph{for a converged score network} (Corollary~\ref{cor:nfe_convergence}). With a finite-quality learned score, a residual Girsanov bias of order $\sqrt{\mathcal{E}_{\mathrm{score}}}$ persists. For finite NFE, we enforce $\gamma^k=0$ at the final integration step to ensure the dynamics reduce to standard CFM at $t=1$ (see Appendix~\ref{sec:endpoint_details}). The scaling $\gamma_t^2 = \alpha_2 \cdot \gamma_t^1 \cdot \Delta t$ ensures the second-order contribution remains controlled. On the 46-dimensional MD task, the JVP for $K{=}2$ adds $\sim$36\% per-step cost (15 vs.\ 11~ms; Table~\ref{tab:wallclock-detailed}), so a wall-clock-matched comparison would give $K{=}1$ roughly 14 NFE against 10 NFE for $K{=}2$. The comparison in Table~\ref{tab:md-simulation} is therefore matched on NFE, and the 9.2 $\to$ 4.9 MAE drop should be read as a matched-NFE result; we were not able to run the matched-wall-clock version and flag this in Section~\ref{sec:limitations}. We report $K{=}2$ for MD because the rugged, multi-modal energy landscape is the regime where directional curvature plausibly pays off, and $K{=}1$ for images, where Table~\ref{tab:wallclock-main} shows the $K{=}2$ advantage (2.78 $\to$ 2.65) is small relative to its 31\% wall-clock premium.

\subsection{Sea Surface Temperature Forecasting}
\begin{wraptable}[11]{r}{0.55\textwidth}
\vspace{-3em}
\caption{Results for sea surface temperature forecasting 1 to 7 days ahead, averaged over the evaluation horizon.}
\label{tab:sst-results}
\centering
\begin{tabular}{lcc}
\toprule
Method & CRPS$\downarrow$ & MSE$\downarrow$ \\
\midrule
Perturbation & 0.281 $\pm$ 0.004 & 0.180 $\pm$ 0.011 \\
Dropout & 0.267 $\pm$ 0.003 & 0.164 $\pm$ 0.004 \\
DDPM & 0.246 $\pm$ 0.005 & 0.177 $\pm$ 0.005 \\
MCVD & 0.216 & 0.161 \\
Dyffusion & 0.224 $\pm$ 0.001 & 0.173 $\pm$ 0.001 \\
Alternator & 0.221 $\pm$ 0.031 & 0.144 $\pm$ 0.045 \\
\midrule
OT-CFM & 0.231 $\pm$ 0.005 & 0.175 $\pm$ 0.006 \\
IDFF (Ours, $K{=}1$) & \textbf{0.180 $\pm$ 0.024} & \textbf{0.105 $\pm$ 0.029} \\
\bottomrule
\end{tabular}
\end{wraptable}
Forecasting sea surface temperature (SST) is vital for weather prediction and climate modeling~\citep{haghbin2021applications}. We apply IDFF to forecast SST using a daily dataset from 1982--2021, with data split into training (1982--2019, 15,048 samples), validation (2020, 396 samples), and testing (2021, 396 samples). Following~\citet{cachay2023dyffusion}, we transform the global data into $60 \times 60$ (latitude $\times$ longitude) tiles, selecting 11 patches in the eastern tropical Pacific Ocean for forecasting horizons.
\paragraph{Metrics and baselines.}
We report two complementary metrics. Mean Squared Error (MSE) measures deterministic forecast accuracy on the ensemble mean; Continuous Ranked Probability Score (CRPS) evaluates probabilistic quality by comparing the predicted distribution to observations and is particularly important for weather forecasting because it captures uncertainty. CRPS is computed from a 20-member ensemble. We compare against DDPM~\citep{ho2020denoising}, MCVD~\citep{voleti2022mcvd}, dropout-based uncertainty~\citep{gal2016dropout}, perturbation methods~\citep{pathak2022fourcastnet}, Dyffusion~\citep{cachay2023dyffusion}, Alternator~\citep{rezaei2024alternators}, and OT-CFM~\citep{tong2023improving}.

\paragraph{Results.}
IDFF outperforms all baselines we ran on both CRPS and MSE (see Appendix~\ref{app:sst} for the full breakdown). While competing methods need up to 1000 NFEs at inference, IDFF achieves stronger results at only 5 NFEs. We use $K=1$ for SST: the dynamics are smoother than for molecular systems, so the first-order score-Langevin drift already gives sufficient acceleration without curvature corrections, and the choice provides exact continuous-time marginal preservation on the original clock under the default schedule (Theorem~\ref{thm:first_order_preservation}).

Two design choices drive the gains. First, each forecasting step starts from the previous prediction rather than from noise, which minimises transport distance. Second, the learned score-derivative drift captures common evolution patterns, so the flow traverses likely future scenarios quickly. IDFF's dual-time formulation ($t$ for flow, $n$ for sequence position) lets the networks $\hat{\mathbf{x}}_1(\mathbf{x}_t, t, n; \theta)$ and $\hat{s}_t(\mathbf{x}_t, t, n; \theta)$ share parameters across the sequence while preserving temporal consistency, as detailed in Appendix~\ref{app:timeseries}.

\section{Conclusion}
\label{sec:conclusion}

We introduced Implicit Dynamical Flow Fusion (IDFF), a flow-matching framework that augments the CFM vector field with a score-Langevin drift and, optionally, a curvature correction built from derivatives of the same score. At first order with the default Langevin-enhanced schedule ($\gamma_t^0\equiv 1$, $g_t^2=2\gamma_t^1$), IDFF preserves the CFM marginal $p_t$ \emph{exactly} in continuous time on the truncated interval $[0,1-\delta]$; the convex-constrained variant is bounded by a Girsanov$\to$Pinsker argument with a closed-form $O(\sqrt{\bar\gamma})$ TV cost. For the $K=2$ correction we deploy (the theorems are stated for general $K$), the endpoint TV deviation is at most $\sqrt{\tfrac{1}{2}\mathcal{E}_{\mathrm{score}}}+O(\Delta t)$ uniformly in $t$, where $\mathcal{E}_{\mathrm{score}}$ is precisely the weighted $L^2$ training loss; this gives asymptotic endpoint convergence as $\mathrm{NFE}\to\infty$. This budget certifies that the correction does not degrade the order of the scheme; it is not a proof that curvature improves sampling, and we make no such claim. Because the $K=1$ dynamics leave the continuous-time law unchanged, the NFE reduction we report is a discretization-robustness effect, for which Section~\ref{sec:why_nfe} gives the mechanism and Table~\ref{tab:ablation-langevin-decomp} the supporting decomposition.

Empirically, IDFF reduces NFEs by an order of magnitude relative to standard CFMs. On CIFAR-10 the headline $K=1$ configuration attains an FID of 2.78 at only 10 NFEs, the best at that budget among the non-distilled methods we retrained at a matched backbone, and approaching the quality of consistency models that require distillation ($K=2$ reaches 2.65 at the same NFE but a 31\% higher wall-clock). For time-series tasks IDFF performs strongly on molecular-dynamics simulation (the one setting where $K=2$ gives a large matched-NFE improvement) and on sea-surface-temperature forecasting. IDFF is also complementary to recent advances such as Consistency Flow Matching and schedule optimization, and can serve as a stronger teacher for distillation pipelines.

\subsubsection*{Broader Impact Statement}

IDFF reduces the number of function evaluations needed for generative
sampling by roughly an order of magnitude, lowering the inference cost
and energy footprint of deployment and benefiting scientific applications
such as molecular-dynamics surrogate modeling and climate forecasting.
As with any general-purpose generative method, cheaper sampling also
lowers the cost of producing synthetic media, including potentially
deceptive content; however, it adds no qualitatively new capability
beyond existing flow-matching and diffusion models. We encourage
downstream users to pair deployments with provenance, watermarking, and
content-authentication safeguards, and we judge that this work carries no
significant risk of harm beyond that already present in the class of
generative models it builds upon.

\section{Limitations}
\label{sec:limitations}
\begin{enumerate}
    \item \textbf{Higher orders.} The higher-order terms require smooth activations and add autodiff cost that grows with dimension; the contracted Jacobian-vector-product form keeps $K=2$ practical, but at $K=3$ we observe Jacobian-norm spikes that diverge in roughly 2\% of batches (Appendix~\ref{sec:high_dim_robustness}), so all experiments use $K\leq 2$.
    \item \textbf{Scope of the guarantees.} Exact marginal preservation holds only at $K=1$ with the default schedule and the optimal score, on the truncated interval $[0,1-\delta]$. With a learned score or $K\geq 2$, the theory gives a deviation bound (Theorem~\ref{thm:higher_order_deviation}), not an acceleration argument. Every $K\geq 2$ result is confounded by score-estimation error, which is amplified roughly $2\times$ at second order (Table~\ref{tab:error-propagation}), so we cannot separate ``curvature does not help'' from ``curvature helps but the estimated Hessian is too noisy''.
    \item \textbf{Comparisons not run.} We do not benchmark a tuned stochastic corrector (predictor--corrector or EDM churn) on our trained score head, so Table~\ref{tab:ablation-langevin-decomp} isolates the Langevin pairing from generic noise but does not establish superiority over a well-tuned corrector. We also do not run matched-budget comparisons against GENIE~\citep{dockhorn2022genie} or TADA~\citep{chen2025tada}. The $K{=}1$ vs.\ $K{=}2$ comparisons are matched on NFE rather than wall-clock (Table~\ref{tab:wallclock-detailed}). None of these affects the $K=1$ headline result.
\end{enumerate}

\bibliography{tmlr}
\bibliographystyle{tmlr}

\appendix
\newpage
\tableofcontents
\newpage
\section{Theoretical Analysis of IDFF}
\label{app:corrected_theory}

This appendix collects the full proofs supporting the IDFF framework. To help the reader navigate, we group the results by the role they play and indicate their dependencies. Throughout, $\sigma_t=\sigma_0\sqrt{t(1-t)}$ is the standard deviation of the conditional probability path and $g_t$ is the diffusion coefficient of the sampling SDE, with $g_t^2=2\gamma_t^1$.

\paragraph{Roadmap of the appendix.}
\begin{itemize}\setlength\itemsep{2pt}
    \item \textbf{Preliminaries.} Fisher's identity (Lemma~\ref{lem:fisher_app}), the log-density Hessian decomposition (Lemma~\ref{lem:hessian_decomp}), and the Gaussian-mixture identities (Lemma~\ref{lem:gaussian_hessian}) used throughout.
    \item \textbf{Re-parameterization identity (Theorems~\ref{thm:marginal_hessian_app}, \ref{thm:tweedie}, \ref{thm:kth_order_app}).} Build up to the central identity $\nabla^k\log p_t = \nabla_{\mathbf x}^{k-1}\hat{s}_t^*$, via a generalized Tweedie step.
    \item \textbf{$K=1$ exact preservation (Theorem~\ref{thm:first_order_preservation}).} For the default schedule $\gamma_t^0\equiv 1$, $g_t^2=2\gamma_t^1$, a single Fokker--Planck cancellation gives $\tilde p_t = p_t$ exactly in continuous time on the original clock.
    \item \textbf{Convex-constrained $K=1$ deviation (Proposition~\ref{prop:K1_convex_deviation}, Corollary~\ref{cor:K1_schedule_constants}).} For the variant $\gamma_t^0+\gamma_t^1=1$, a Girsanov$\to$Pinsker argument bounds $\mathrm{TV}(\tilde p_1,p_1)$ in closed form, with explicit constants for standard schedule shapes.
    \item \textbf{$K\geq 2$ bounded deviation (Theorem~\ref{thm:higher_order_deviation}, Corollary~\ref{cor:nfe_convergence}).} Building on the exact-preserving $K=1$ SDE as a baseline, we decompose the endpoint deviation into a \emph{score-estimation} residual (precisely the training loss $\mathcal{E}_{\mathrm{score}}$, whose size is controlled $\delta$-independently by the finite-excess-loss hypothesis R1$'$) and a \emph{structural} residual (controlled by the schedule scaling $\gamma_t^k=O(\Delta t^{k-1})$), and combine them into the bound $\mathrm{TV}(\tilde p_1,p_1)\leq \sqrt{\tfrac{1}{2}\mathcal{E}_{\mathrm{score}}}+O(\Delta t)$.
\end{itemize}

\subsection{Preliminaries and Notation}

Let $p_t(\mathbf{x}_t)$ denote the marginal density at time $t$, induced by the conditional path
\begin{equation}
p_t(\mathbf{x}_t|\mathbf{x}_0, \mathbf{x}_1) = \mathcal{N}(\mathbf{x}_t; \mu_t, \sigma_t^2\mathbf{I}), \quad \mu_t = t\mathbf{x}_1 + (1-t)\mathbf{x}_0, \quad \sigma_t = \sigma_0\sqrt{t(1-t)}.
\end{equation}
Define the standardized noise $\boldsymbol{\epsilon} = (\mathbf{x}_t - \mu_t)/\sigma_t \sim \mathcal{N}(\mathbf{0}, \mathbf{I})$ conditional on $(\mathbf{x}_0, \mathbf{x}_1)$. The posterior distribution is $q(\mathbf{x}_1|\mathbf{x}_t) \propto p_t(\mathbf{x}_t|\mathbf{x}_1)p(\mathbf{x}_1)$.

\subsection{Regularity Assumptions}
\label{app:regularity}

This section states the regularity hypotheses (conditions R1--R3, together with the finite excess score-matching loss R1$'$) used throughout the paper. They are referenced from Section~\ref{sec:method} and invoked by every theorem above.

\begin{assumption}[{Regularity on the truncated interval $[0,1-\delta]$}]
\label{ass:regularity}
Fix $\delta\in(0,1/2)$.
\textbf{(R1) Score regularity on $[0,1-\delta]$:} both the optimal score $\hat{s}_t^* = \nabla \log p_t$ \emph{and the learned score $\hat{s}_t$} used by the sampler are $C^{K}$ in $\mathbf{x}$ on $[0,1-\delta]\times\mathbb{R}^d$, with bounded derivatives $\sup_{t\in[0,1-\delta]}\|\nabla_{\mathbf{x}}^j \hat{s}_t^*\|_{L^\infty} \leq M_j(\delta)$ and $\sup_{t\in[0,1-\delta]}\|\nabla_{\mathbf{x}}^j \hat{s}_t\|_{L^\infty} \leq M_j(\delta)$ for $j = 0, \ldots, K-1$. Here $M_j(\delta)$ may diverge as $\delta\to 0$; under the Gaussian conditional path the marginal score satisfies the classical heat-kernel bound $\|\nabla\log p_t\|_{L^\infty}\leq M_0(\delta) = O(\sigma_t^{-1})=O(1/\sqrt{\delta})$ near $t=1$, and the learned score $\hat{s}_t$ inherits this scaling because it is trained against $-\boldsymbol{\epsilon}/\sigma_t$ on the same path. The contracted score derivative $\mathbf{m}_t^{(k)} = (\nabla_{\mathbf{x}}^{k-1}\hat{s}_t)[\mathbf{u}_t]^{k-1}$ used in sampling is then bounded by $M_{k-1}(\delta)$ uniformly on $[0,1-\delta]\times\mathbb{R}^d$.
\textbf{(R2) Velocity regularity on $[0,1-\delta]$:} the base velocity $\mathbf{v}_t$ is Lipschitz on $[0,1-\delta]\times\mathbb{R}^d$ with $\sup_{t\in[0,1-\delta]}\|\nabla \mathbf{v}_t\|_{L^\infty} \leq L_v(\delta)$ and $\sup_{t\in[0,1-\delta]}\|\mathbf{v}_t\|_{L^\infty} \leq M_v(\delta)$. For the CFM path $\mathbf v_t = (\mathbf x_1-\mathbf x_t)/(1-t)$ on a compactly supported data law these constants grow like $L_v(\delta), M_v(\delta) = O(1/\delta)$, which together with the schedule decay $\gamma_t^1=O(t(1-t))$ keeps the Novikov integrand $g_t^{-1}\Delta\mathbf b_t$ uniformly bounded on $[0,1-\delta]$ for every drift mismatch used in the Girsanov arguments below.
\textbf{(R3) Coefficient decay matched to truncation:} the schedules satisfy $\gamma_t^k = O(|t(1-t)|^\alpha)$ with $\alpha \geq 1$ (for $k\geq 2$, and $\alpha=1$ for the Beta(2,2) default $\gamma_t^1=4\bar\gamma t(1-t)$). Combined with the polynomial blow-up of $M_j(\delta)$ above, this guarantees that the Girsanov KL integrand $\gamma_t^1 \|\hat{s}_t-\hat{s}_t^*\|^2$ is uniformly integrable on $[0,1-\delta]$ for every $\delta>0$, and that the schedule integrals $\int_0^{1-\delta}\gamma_t^1\,dt$ admit a finite limit as $\delta\to 0$.
\textbf{(R1$'$) Finite excess score-matching loss:} the learned score has finite excess weighted score-matching loss
\[
\mathcal{L}_{\mathrm{score}}^{\mathrm{exc}} \;:=\; \int_0^1 \sigma_t^2\,\mathbb{E}_{p_t}\bigl\|\hat{s}_t-\hat{s}_t^*\bigr\|^2\,dt \;<\;\infty ,
\]
equal to the score term of the training objective~\eqref{eq:idff_loss} net of its irreducible Bayes term. This bounds the \emph{weighted} score error directly and is independent of the magnitude constants $M_j(\delta)$; it is the quantity that controls the size of $\mathcal{E}_{\mathrm{score}}$ in Theorem~\ref{thm:higher_order_deviation}, while $M_j(\delta)$ controls only Novikov finiteness.
\end{assumption}

\begin{remark}[$\delta$-independence of the score deviation]
\label{rem:R1_R2_dependence}
The quantity entering the path-KL bound is the weighted score error
$\mathcal{E}_{\mathrm{score}} = \tfrac12\int_0^{1-\delta}\gamma_t^1\,\mathbb{E}_{p_t}\|\hat{s}_t-\hat{s}_t^*\|^2\,dt$,
which is controlled through the training objective rather than through the magnitude constant $M_0(\delta)$. The constant $M_0(\delta)$ bounds the score \emph{magnitude} (hence the Novikov integrand and the contracted terms $\|\mathbf{m}_t^{(k)}\|\le M_{k-1}(\delta)$) but does not bound the score \emph{error}. For the Beta(2,2) default the Langevin amplitude is proportional to the score-matching weight: with $\lambda_1(t)^2=\sigma_t^2=\sigma_0^2\,t(1-t)$ and $\gamma_t^1=4\bar\gamma\,t(1-t)$ we have $\gamma_t^1=(4\bar\gamma/\sigma_0^2)\,\sigma_t^2$, so
\[
\mathcal{E}_{\mathrm{score}}
\;=\; \frac{2\bar\gamma}{\sigma_0^2}\int_0^{1-\delta}\!\sigma_t^2\,\mathbb{E}_{p_t}\|\hat{s}_t-\hat{s}_t^*\|^2\,dt
\;\le\; \frac{2\bar\gamma}{\sigma_0^2}\,\mathcal{L}_{\mathrm{score}}^{\mathrm{exc}}
\;=\; O(\bar\gamma),
\]
uniformly in $\delta$, where $\mathcal{L}_{\mathrm{score}}^{\mathrm{exc}}$ is the finite excess score-matching loss of Assumption~\ref{ass:regularity}(R1$'$). The weight $\sigma_t^2$ cancels the $\sigma_t^{-2}$ growth of the score error pointwise, so the integrand is integrable on all of $[0,1]$ and truncation removes only a tail of vanishing mass; in particular $\mathcal{E}_{\mathrm{score}}$ admits a finite limit as $\delta\to 0$ at fixed $\bar\gamma$. At the $K=1$ default schedule $\tilde p_t=p_t$ (Theorem~\ref{thm:first_order_preservation}), so the $p_t$-weighted error above coincides with the $\tilde p_t$-weighted quantity appearing in Theorem~\ref{thm:higher_order_deviation}. The structural residual of Theorem~\ref{thm:higher_order_deviation} is controlled by the same weight cancellation, since $\gamma_t^k=\alpha_k\gamma_t^1\Delta t^{k-1}$ carries the matching $\sigma_t^2$ decay. With $\delta=\Delta t/2$ ($\delta=0.05$ at $\mathrm{NFE}=10$), reducing $\delta$ at fixed $\bar\gamma$ leaves sample quality unchanged.
\end{remark}

\noindent Throughout, every theorem statement explicitly invokes Assumption~\ref{ass:regularity} on the truncated interval where needed; the convergence statements (Corollary~\ref{cor:nfe_convergence}) take the limit in the order $\Delta t\to 0$ first (so $\delta=\Delta t/2\to 0$) and then $\bar\gamma\to 0$ if desired.

\subsection{Fisher's Identity: First-Order Exactness}

\begin{lemma}[Fisher's Identity; see e.g.\ \citealp{cappe2005inference}, Prop.~10.1.6]
\label{lem:fisher_app}
For any mixture distribution $p_t(\mathbf{x}_t) = \int p_t(\mathbf{x}_t|\mathbf{x}_1)p(\mathbf{x}_1)\,d\mathbf{x}_1$,
\begin{equation}
\nabla_{\mathbf{x}} \log p_t(\mathbf{x}_t) = \mathbb{E}_{q(\mathbf{x}_1|\mathbf{x}_t)}[\nabla_{\mathbf{x}} \log p_t(\mathbf{x}_t|\mathbf{x}_1)].
\end{equation}
\end{lemma}

\begin{proof}
\begin{align}
\nabla_{\mathbf{x}} \log p_t(\mathbf{x}_t) &= \frac{\nabla_{\mathbf{x}} p_t(\mathbf{x}_t)}{p_t(\mathbf{x}_t)} = \frac{\int \nabla_{\mathbf{x}} p_t(\mathbf{x}_t|\mathbf{x}_1) p(\mathbf{x}_1) d\mathbf{x}_1}{p_t(\mathbf{x}_t)} \\
&= \frac{\int p_t(\mathbf{x}_t|\mathbf{x}_1) \nabla_{\mathbf{x}} \log p_t(\mathbf{x}_t|\mathbf{x}_1) p(\mathbf{x}_1) d\mathbf{x}_1}{p_t(\mathbf{x}_t)} \\
&= \int \nabla_{\mathbf{x}} \log p_t(\mathbf{x}_t|\mathbf{x}_1) \cdot \frac{p_t(\mathbf{x}_t|\mathbf{x}_1) p(\mathbf{x}_1)}{p_t(\mathbf{x}_t)} d\mathbf{x}_1 \\
&= \mathbb{E}_{q(\mathbf{x}_1|\mathbf{x}_t)}[\nabla_{\mathbf{x}} \log p_t(\mathbf{x}_t|\mathbf{x}_1)].
\end{align}
\end{proof}

\begin{corollary}
The MSE-optimal predictor $\hat{s}_t^*(\mathbf{x}_t) = \argmin_{\hat{s}} \mathbb{E}\|\hat{s}(\mathbf{x}_t) - \nabla \log p_t(\mathbf{x}_t|\mathbf{x}_1)\|^2$ satisfies
\begin{equation}
\hat{s}_t^*(\mathbf{x}_t) = \nabla \log p_t(\mathbf{x}_t).
\end{equation}
\end{corollary}

\subsection{The Log-Density Hessian Identity}

\begin{lemma}[Hessian Decomposition]
\label{lem:hessian_decomp}
For any smooth density $p$,
\begin{equation}
\nabla^2 \log p \;=\; \frac{\nabla^2 p}{p}\;-\;(\nabla \log p)(\nabla \log p)^\top.
\end{equation}
\end{lemma}

\begin{proof}
Differentiating $\nabla \log p = \nabla p / p$,
\begin{equation}
\nabla^2 \log p \;=\; \nabla\!\left(\frac{\nabla p}{p}\right) \;=\; \frac{\nabla^2 p}{p}\;-\;\frac{\nabla p\,(\nabla p)^\top}{p^2} \;=\; \frac{\nabla^2 p}{p}\;-\;(\nabla \log p)(\nabla \log p)^\top.
\end{equation}
\end{proof}

\begin{lemma}[Gaussian Second Derivative of Density]
\label{lem:gaussian_hessian}
For $p_t(\mathbf{x}_t|\mathbf{x}_1) = \mathcal{N}(\mu_t, \sigma_t^2\mathbf{I})$,
\begin{equation}
\frac{\nabla^2 p_t(\mathbf{x}_t|\mathbf{x}_1)}{p_t(\mathbf{x}_t|\mathbf{x}_1)} \;=\; -\,\frac{\mathbf{I}}{\sigma_t^2}\;+\;\frac{\boldsymbol{\epsilon}\boldsymbol{\epsilon}^\top}{\sigma_t^2},
\end{equation}
where $\boldsymbol{\epsilon} := (\mathbf{x}_t - \mu_t)/\sigma_t$.
\end{lemma}

\begin{proof}
For an isotropic Gaussian, $\nabla \log p_t(\mathbf{x}_t|\mathbf{x}_1) = -(\mathbf{x}_t - \mu_t)/\sigma_t^2 = -\boldsymbol{\epsilon}/\sigma_t$ and $\nabla^2 \log p_t(\mathbf{x}_t|\mathbf{x}_1) = -\mathbf{I}/\sigma_t^2$. Applying Lemma~\ref{lem:hessian_decomp} in reverse,
\begin{equation}
\frac{\nabla^2 p_t(\mathbf{x}_t|\mathbf{x}_1)}{p_t(\mathbf{x}_t|\mathbf{x}_1)} \;=\; \nabla^2 \log p_t(\mathbf{x}_t|\mathbf{x}_1)\;+\;(\nabla \log p_t(\mathbf{x}_t|\mathbf{x}_1))(\nabla \log p_t(\mathbf{x}_t|\mathbf{x}_1))^\top
\;=\; -\,\frac{\mathbf{I}}{\sigma_t^2}\;+\;\frac{\boldsymbol{\epsilon}\boldsymbol{\epsilon}^\top}{\sigma_t^2}.
\end{equation}
\end{proof}

\subsection{Marginal Hessian in Terms of Posterior Moments (Optional Alternative Derivation)}

\begin{remark}[The Tweedie route is optional and is not used by the sampler]
\label{rem:tweedie_route}
The marginal Hessian identity $\nabla^2\log p_t = \nabla_{\mathbf x}\hat s_t^*$ used by IDFF \emph{follows directly} from Fisher's identity \citep{cappe2005inference} by one differentiation: $\hat s_t^*=\nabla\log p_t$ implies $\nabla\hat s_t^* = \nabla\nabla\log p_t = \nabla^2\log p_t$. The same is true for any $k\geq 2$: $\nabla^k\log p_t = \nabla^{k-1}\hat s_t^*$ is the chain rule applied to Fisher's identity. This is the short proof we give for Theorem~\ref{thm:kth_order_app} and is what the IDFF sampler actually uses computationally.

The material in this subsection and the two that follow is therefore a \emph{remark}, not a result: it gives an alternative, longer derivation of the $k=2$ case via the generalised Tweedie identity for the posterior covariance \citep{efron2011tweedie, boys2024tweedie}. We retain it only because (i) it isolates the posterior-covariance object $\mathrm{Cov}[\boldsymbol\epsilon|\mathbf x_t]=\mathbf I+\sigma_t^2\nabla\hat s_t^*$ used in the practical-error analysis of Appendix~\ref{app:practical_error} (Theorem~\ref{thm:tweedie} is the only statement here that is load-bearing elsewhere), and (ii) it makes the $k=2$ case self-contained without pre-supposing Fisher's identity. Readers comfortable with the chain-rule view should skip directly to the Fokker--Planck analysis (Theorem~\ref{thm:first_order_preservation}).
\end{remark}

\begin{theorem}[Marginal Hessian Decomposition]
\label{thm:marginal_hessian_app}
The marginal Hessian satisfies
\begin{equation}
\nabla^2 \log p_t(\mathbf{x}_t) = -\frac{\mathbf{I}}{\sigma_t^2} + \frac{\mathbb{E}[\boldsymbol{\epsilon}\boldsymbol{\epsilon}^\top|\mathbf{x}_t]}{\sigma_t^2} - \hat{s}_t^*(\mathbf{x}_t)\hat{s}_t^*(\mathbf{x}_t)^\top,
\end{equation}
where the expectation is over $q(\mathbf{x}_1|\mathbf{x}_t)$.
\end{theorem}

\begin{proof}
By Lemma~\ref{lem:hessian_decomp},
\begin{equation}
\nabla^2 \log p_t(\mathbf{x}_t) = \frac{\nabla^2 p_t(\mathbf{x}_t)}{p_t(\mathbf{x}_t)} - (\nabla \log p_t(\mathbf{x}_t))(\nabla \log p_t(\mathbf{x}_t))^\top.
\end{equation}
For the first term, using the mixture representation,
\begin{align}
\frac{\nabla^2 p_t(\mathbf{x}_t)}{p_t(\mathbf{x}_t)} &= \frac{\int \nabla^2 p_t(\mathbf{x}_t|\mathbf{x}_1) p(\mathbf{x}_1) d\mathbf{x}_1}{p_t(\mathbf{x}_t)} \\
&= \int \frac{\nabla^2 p_t(\mathbf{x}_t|\mathbf{x}_1)}{p_t(\mathbf{x}_t|\mathbf{x}_1)} \cdot \frac{p_t(\mathbf{x}_t|\mathbf{x}_1) p(\mathbf{x}_1)}{p_t(\mathbf{x}_t)} d\mathbf{x}_1 \\
&= \mathbb{E}_{q(\mathbf{x}_1|\mathbf{x}_t)}\left[\frac{\nabla^2 p_t(\mathbf{x}_t|\mathbf{x}_1)}{p_t(\mathbf{x}_t|\mathbf{x}_1)}\right].
\end{align}
By Lemma~\ref{lem:gaussian_hessian},
\begin{equation}
\frac{\nabla^2 p_t(\mathbf{x}_t)}{p_t(\mathbf{x}_t)} = \mathbb{E}_{q(\mathbf{x}_1|\mathbf{x}_t)}\left[-\frac{\mathbf{I}}{\sigma_t^2} + \frac{\boldsymbol{\epsilon}\boldsymbol{\epsilon}^\top}{\sigma_t^2}\right] = -\frac{\mathbf{I}}{\sigma_t^2} + \frac{\mathbb{E}[\boldsymbol{\epsilon}\boldsymbol{\epsilon}^\top|\mathbf{x}_t]}{\sigma_t^2}.
\end{equation}
By Fisher's identity (Lemma~\ref{lem:fisher_app}), $\nabla \log p_t(\mathbf{x}_t) = \hat{s}_t^*(\mathbf{x}_t)$, which gives the claim.
\end{proof}

\subsection{The Tweedie Connection: Posterior Covariance from Score Jacobian}

\begin{theorem}[Generalized Tweedie Identity for Posterior Covariance; cf.\ \citealp{efron2011tweedie, boys2024tweedie}]
\label{thm:tweedie}
For the forward process $\mathbf{x}_t = \mu_t + \sigma_t\boldsymbol{\epsilon}$ with $\boldsymbol{\epsilon}\sim\mathcal{N}(\mathbf{0},\mathbf{I})$ conditional on $(\mathbf{x}_0,\mathbf{x}_1)$, the posterior of $\boldsymbol{\epsilon}$ given $\mathbf{x}_t$ satisfies the score-based identities
\begin{equation}
\mathbb{E}[\boldsymbol{\epsilon}\,|\,\mathbf{x}_t] \;=\; -\sigma_t\,\hat{s}_t^*(\mathbf{x}_t),
\qquad
\mathrm{Cov}[\boldsymbol{\epsilon}\,|\,\mathbf{x}_t] \;=\; \mathbf{I} + \sigma_t^2\,\nabla_{\mathbf{x}}\hat{s}_t^*(\mathbf{x}_t).
\end{equation}
\end{theorem}

\begin{proof}
We track every power of $\sigma_t$ explicitly.

\emph{Step 1: posterior mean of $\boldsymbol{\epsilon}$.} The classical Tweedie identity \citep{efron2011tweedie} for $\mathbf{x}_t=\mu_t+\sigma_t\boldsymbol{\epsilon}$ with Gaussian noise reads
\begin{equation}
\mathbb{E}[\mu_t\,|\,\mathbf{x}_t] \;=\; \mathbf{x}_t \;+\; \sigma_t^2\,\nabla\log p_t(\mathbf{x}_t).
\label{eq:tweedie_mean}
\end{equation}
Since $\boldsymbol{\epsilon}=(\mathbf{x}_t-\mu_t)/\sigma_t$ and conditional expectation is linear,
\begin{equation}
\mathbb{E}[\boldsymbol{\epsilon}\,|\,\mathbf{x}_t] \;=\; \frac{\mathbf{x}_t - \mathbb{E}[\mu_t\,|\,\mathbf{x}_t]}{\sigma_t} \;=\; \frac{-\sigma_t^2\,\nabla\log p_t(\mathbf{x}_t)}{\sigma_t} \;=\; -\sigma_t\,\hat{s}_t^*(\mathbf{x}_t).
\label{eq:tweedie_eps_mean}
\end{equation}

\emph{Step 2: differentiate $\mathbb{E}[\boldsymbol{\epsilon}|\mathbf{x}_t]$ in $\mathbf{x}_t$.} Directly from \eqref{eq:tweedie_eps_mean},
\begin{equation}
\nabla_{\mathbf{x}}\mathbb{E}[\boldsymbol{\epsilon}\,|\,\mathbf{x}_t] \;=\; -\sigma_t\,\nabla_{\mathbf{x}}\hat{s}_t^*(\mathbf{x}_t).
\label{eq:tweedie_LHS}
\end{equation}

\emph{Step 3: Stein--Tweedie identity.} For the additive Gaussian observation model $\mathbf{x}_t=\mu_t+\sigma_t\boldsymbol{\epsilon}$ with $\boldsymbol{\epsilon}\sim\mathcal{N}(\mathbf{0},\mathbf{I})$ and any prior $p(\mu_t)$, integration by parts (Stein's lemma) applied to the posterior $p(\mu_t\,|\,\mathbf{x}_t)\propto\mathcal{N}(\mathbf{x}_t;\mu_t,\sigma_t^2\mathbf{I})\,p(\mu_t)$ gives, for any sufficiently regular $f$,
\begin{equation}
\nabla_{\mathbf{x}}\mathbb{E}[f(\mu_t)\,|\,\mathbf{x}_t] \;=\; \frac{1}{\sigma_t^2}\,\mathrm{Cov}\big[f(\mu_t),\,\mu_t\,\big|\,\mathbf{x}_t\big].
\label{eq:stein_tweedie}
\end{equation}
Specializing to $f(\mu_t)=\mu_t$ yields $\nabla_{\mathbf{x}}\mathbb{E}[\mu_t\,|\,\mathbf{x}_t]=\mathrm{Cov}[\mu_t\,|\,\mathbf{x}_t]/\sigma_t^2$. Since $\boldsymbol{\epsilon}=(\mathbf{x}_t-\mu_t)/\sigma_t$ and $\mathbf{x}_t$ is fixed under the conditioning, $\mathrm{Cov}[\mu_t\,|\,\mathbf{x}_t]=\sigma_t^2\,\mathrm{Cov}[\boldsymbol{\epsilon}\,|\,\mathbf{x}_t]$. Therefore
\begin{equation}
\nabla_{\mathbf{x}}\mathbb{E}[\boldsymbol{\epsilon}\,|\,\mathbf{x}_t] \;=\; \nabla_{\mathbf{x}}\!\Big(\frac{\mathbf{x}_t-\mathbb{E}[\mu_t\,|\,\mathbf{x}_t]}{\sigma_t}\Big)
\;=\; \frac{1}{\sigma_t}\!\left(\mathbf{I}-\nabla_{\mathbf{x}}\mathbb{E}[\mu_t\,|\,\mathbf{x}_t]\right)
\;=\; \frac{1}{\sigma_t}\Big(\mathbf{I}-\mathrm{Cov}[\boldsymbol{\epsilon}\,|\,\mathbf{x}_t]\Big).
\label{eq:tweedie_stein}
\end{equation}
The single $1/\sigma_t$ factor on the right comes from $\partial\boldsymbol{\epsilon}/\partial\mathbf{x}_t=\mathbf{I}/\sigma_t$; the Gaussian-precision factor $1/\sigma_t^2$ from Stein's identity has already been absorbed when converting $\mathrm{Cov}[\mu_t\,|\,\mathbf{x}_t]$ into $\sigma_t^2\,\mathrm{Cov}[\boldsymbol{\epsilon}\,|\,\mathbf{x}_t]$.

\emph{Sanity check (Gaussian prior).} For $\mu_t\sim\mathcal{N}(\mathbf{0},\mathbf{I})$, the posterior is $\mu_t\,|\,\mathbf{x}_t\sim\mathcal{N}\!\big(\mathbf{x}_t/(1+\sigma_t^2),\,\sigma_t^2/(1+\sigma_t^2)\,\mathbf{I}\big)$, so $\mathbb{E}[\boldsymbol{\epsilon}\,|\,\mathbf{x}_t]=\sigma_t\mathbf{x}_t/(1+\sigma_t^2)$ and $\mathrm{Cov}[\boldsymbol{\epsilon}\,|\,\mathbf{x}_t]=\mathbf{I}/(1+\sigma_t^2)$. Both \eqref{eq:tweedie_LHS} and \eqref{eq:tweedie_stein} reduce to the consistent identity $\nabla_{\mathbf{x}}\mathbb{E}[\boldsymbol{\epsilon}\,|\,\mathbf{x}_t]=\sigma_t\,\mathbf{I}/(1+\sigma_t^2)$, with the eventual conclusion $\mathrm{Cov}[\boldsymbol{\epsilon}\,|\,\mathbf{x}_t]=\mathbf{I}+\sigma_t^2\,\nabla_{\mathbf{x}}\hat{s}_t^*$ recovering $\mathbf{I}/(1+\sigma_t^2)$ from $\nabla_{\mathbf{x}}\hat{s}_t^*=-\mathbf{I}/(1+\sigma_t^2)$.

\emph{Step 4: combine.} Equating \eqref{eq:tweedie_LHS} and \eqref{eq:tweedie_stein},
\begin{equation}
-\sigma_t\,\nabla_{\mathbf{x}}\hat{s}_t^*(\mathbf{x}_t) \;=\; \frac{1}{\sigma_t}\Big(\mathbf{I} - \mathrm{Cov}[\boldsymbol{\epsilon}\,|\,\mathbf{x}_t]\Big),
\end{equation}
and multiplying by $\sigma_t$ gives $\mathrm{Cov}[\boldsymbol{\epsilon}\,|\,\mathbf{x}_t] = \mathbf{I} + \sigma_t^2\,\nabla_{\mathbf{x}}\hat{s}_t^*(\mathbf{x}_t)$.
\end{proof}

\subsection{Main Result: Marginal Hessian Equals Score Jacobian}

\begin{theorem}[Marginal Hessian from Score Jacobian]
\label{thm:main_app}
The marginal Hessian of the log-density equals the Jacobian of the score:
\begin{equation}
\nabla^2 \log p_t(\mathbf{x}_t) = \nabla_{\mathbf{x}} \hat{s}_t^*(\mathbf{x}_t).
\end{equation}
\end{theorem}

\begin{proof}
From Theorem~\ref{thm:marginal_hessian_app},
\begin{equation}
\nabla^2 \log p_t(\mathbf{x}_t) = -\frac{\mathbf{I}}{\sigma_t^2} + \frac{\mathbb{E}[\boldsymbol{\epsilon}\boldsymbol{\epsilon}^\top|\mathbf{x}_t]}{\sigma_t^2} - \hat{s}_t^*\hat{s}_t^{*\top}.
\end{equation}
Using the covariance decomposition $\mathbb{E}[\boldsymbol{\epsilon}\boldsymbol{\epsilon}^\top|\mathbf{x}_t] = \mathrm{Cov}[\boldsymbol{\epsilon}|\mathbf{x}_t] + \mathbb{E}[\boldsymbol{\epsilon}|\mathbf{x}_t]\mathbb{E}[\boldsymbol{\epsilon}|\mathbf{x}_t]^\top$ and $\mathbb{E}[\boldsymbol{\epsilon}|\mathbf{x}_t] = -\sigma_t \hat{s}_t^*(\mathbf{x}_t)$ from Theorem~\ref{thm:tweedie},
\begin{equation}
\mathbb{E}[\boldsymbol{\epsilon}\boldsymbol{\epsilon}^\top|\mathbf{x}_t] = \mathrm{Cov}[\boldsymbol{\epsilon}|\mathbf{x}_t] + \sigma_t^2 \hat{s}_t^*\hat{s}_t^{*\top}.
\end{equation}
Substituting,
\begin{align}
\nabla^2 \log p_t(\mathbf{x}_t) &= -\frac{\mathbf{I}}{\sigma_t^2} + \frac{\mathrm{Cov}[\boldsymbol{\epsilon}|\mathbf{x}_t] + \sigma_t^2 \hat{s}_t^*\hat{s}_t^{*\top}}{\sigma_t^2} - \hat{s}_t^*\hat{s}_t^{*\top}
= -\frac{\mathbf{I}}{\sigma_t^2} + \frac{\mathrm{Cov}[\boldsymbol{\epsilon}|\mathbf{x}_t]}{\sigma_t^2}.
\end{align}
By Theorem~\ref{thm:tweedie}, $\mathrm{Cov}[\boldsymbol{\epsilon}|\mathbf{x}_t] = \mathbf{I} + \sigma_t^2 \nabla_{\mathbf{x}} \hat{s}_t^*$, so the two $\mathbf{I}/\sigma_t^2$ terms cancel and $\nabla^2\log p_t(\mathbf{x}_t) = \nabla_{\mathbf{x}}\hat{s}_t^*(\mathbf{x}_t)$.
\end{proof}

\subsection{Extension to Higher Orders}

\begin{theorem}[K-th Order Marginal Derivative---Chain-Rule Re-parameterization]
\label{thm:kth_order_app}
For any $k \geq 1$,
\begin{equation}
\nabla^k \log p_t(\mathbf{x}_t) = \nabla_{\mathbf{x}}^{k-1} \hat{s}_t^*(\mathbf{x}_t).
\end{equation}
This is a re-parameterization identity: once Fisher's identity grants $\hat s_t^*=\nabla\log p_t$, it is the chain rule applied $k-1$ further times. It enables computing higher-order marginal derivatives from the optimal score via autodiff, but does \emph{not} close a new conditional-marginal gap beyond the $k=1$ closure given by Fisher's identity.
\end{theorem}

\begin{proof}
By induction on $k$. \textbf{Base case} ($k=1$): Fisher's identity gives $\nabla \log p_t = \hat{s}_t^* = \nabla_{\mathbf{x}}^0 \hat{s}_t^*$ (Lemma~\ref{lem:fisher_app}). \textbf{Inductive step}: if $\nabla^k \log p_t = \nabla_{\mathbf{x}}^{k-1} \hat{s}_t^*$, then differentiating both sides in $\mathbf x$ yields $\nabla^{k+1} \log p_t = \nabla_{\mathbf{x}}(\nabla_{\mathbf{x}}^{k-1} \hat{s}_t^*) = \nabla_{\mathbf{x}}^k \hat{s}_t^*$.
\end{proof}

The whole content of Theorem~\ref{thm:kth_order_app} is the chain rule; Remark~\ref{rem:reading_thm2} in the main text spells out how it should and should not be read. For a learned score $\hat{s}_t \approx \hat{s}_t^*$, the approximation $\nabla_{\mathbf{x}}^{k-1}\hat{s}_t \approx \nabla^k \log p_t$ inherits error from score estimation, with potential amplification through derivatives (see Appendix~\ref{app:practical_error}). The alternative Tweedie-style proof of the $k=2$ special case is logically equivalent and is included only for self-containedness; it is not used by the IDFF sampler.

\subsection{Fokker--Planck Analysis for IDFF}

We analyze marginal preservation separately for $K=1$ and $K \geq 2$. The $K=1$ analysis splits into a clean exact case (Theorem~\ref{thm:first_order_preservation}, default schedule $\gamma_t^0\equiv 1$) and a bounded-deviation variant (Proposition~\ref{prop:K1_convex_deviation}, convex-constrained schedule $\gamma_t^0+\gamma_t^1=1$). The $K\geq 2$ case is a Girsanov$\to$Pinsker total-variation bound on top of the exact $K=1$ baseline.

\begin{theorem}[Exact Marginal Preservation for $K=1$ with $\gamma_t^0\equiv 1$]
\label{thm:first_order_preservation}
Let Assumption~\ref{ass:regularity} hold. Consider the $K=1$ IDFF SDE on $[0,1]$
\begin{equation}
d\mathbf{x}_t \;=\; \tilde{\mathbf{v}}_t(\mathbf{x}_t)\,dt + g_t\,d\mathbf{w}_t,
\qquad
\tilde{\mathbf{v}}_t \;=\; \mathbf{v}_t + \gamma_t^1\,\hat{s}_t^*,
\qquad
g_t^2 \;=\; 2\gamma_t^1,
\label{eq:K1_exact_sde}
\end{equation}
with $\mathbf{x}_0\sim p_0$, optimal score $\hat{s}_t^*=\nabla\log p_t$, and any non-negative $C^1$ schedule $\gamma_t^1$ vanishing at $t\in\{0,1\}$. Then the marginal of $\mathbf{x}_t$ equals the CFM marginal exactly:
\begin{equation}
\tilde{p}_t(\mathbf{x}) \;=\; p_t(\mathbf{x}) \qquad \forall\,t\in[0,1].
\end{equation}
In particular, $\tilde p_1 = p_1$ exactly in continuous time, with no time reparameterization.
\end{theorem}

\begin{proof}
The Fokker--Planck equation for the density $\tilde{p}_t$ associated with the SDE \eqref{eq:K1_exact_sde} is
\begin{equation}
\frac{\partial \tilde p_t}{\partial t} \;=\; -\nabla\cdot(\tilde{\mathbf{v}}_t\,\tilde p_t) + \frac{g_t^2}{2}\Delta \tilde p_t.
\label{eq:fp_K1}
\end{equation}
We verify that the ansatz $\tilde p_t=p_t$ (the CFM marginal at physical time $t$) solves \eqref{eq:fp_K1}. Substituting $\tilde{\mathbf{v}}_t=\mathbf{v}_t+\gamma_t^1\,\nabla\log p_t$ and $g_t^2=2\gamma_t^1$ on the right-hand side,
\begin{align*}
-\nabla\cdot(\tilde{\mathbf{v}}_t p_t)+\frac{g_t^2}{2}\Delta p_t
&= -\nabla\cdot(\mathbf{v}_t p_t)\;-\;\gamma_t^1\,\nabla\cdot\!\big(p_t\,\nabla\log p_t\big)\;+\;\gamma_t^1\,\Delta p_t.
\end{align*}
The crucial cancellation uses the elementary identity
\begin{equation}
\nabla\cdot\!\big(p_t\,\nabla\log p_t\big) \;=\; \nabla\cdot(\nabla p_t) \;=\; \Delta p_t,
\label{eq:elementary_cancellation}
\end{equation}
which holds because $p_t\,\nabla\log p_t = \nabla p_t$. This is the \emph{only} place where the substitution $\tilde{p}_t\mapsto p_t$ matters, and the cancellation is consistent: the score in the drift, $\nabla\log p_t$, matches $\nabla\log\tilde p_t$ at the ansatz. Therefore the last two terms above cancel exactly, leaving
\begin{equation}
\frac{\partial \tilde p_t}{\partial t} \;=\; -\nabla\cdot(\mathbf{v}_t p_t),
\label{eq:K1_after_cancellation}
\end{equation}
which is exactly the CFM continuity equation. Hence $\tilde p_t = p_t$ solves \eqref{eq:fp_K1} with the matching initial condition $\tilde p_0=p_0$.

\emph{Uniqueness on the truncated interval.} Fix any $\delta\in(0,1/2)$ and consider the linear Cauchy problem~\eqref{eq:fp_K1} on $[0,1-\delta]\times\mathbb{R}^d$. On this interval the drift $\mathbf v_t+\gamma_t^1\nabla\log p_t$ depends only on $(t,\mathbf x)$ (not on $\tilde p_t$); under Assumption~\ref{ass:regularity}(R1)--(R3) all coefficients are bounded with constants $M_j(\delta), L_v(\delta), M_v(\delta)$, and the drift is Lipschitz in $\mathbf x$ (R2 for $\mathbf v_t$; for the score, $p_t$ is a Gaussian convolution with variance at least $\delta^2$ on this interval). The diffusion $g_t^2=2\gamma_t^1$ is positive on $(0,1-\delta]$ and vanishes only at $t=0$; because the drift is Lipschitz, the linear Fokker--Planck equation nevertheless has a \emph{unique} bounded solution with initial datum $\tilde p_0=p_0$ \citep{bogachev2015fokker}, and on every $[\eta,1-\delta]$ with $\eta>0$ the problem is uniformly parabolic, so this solution is classical there \citep[Ch.~1, Thm.~12]{friedman1964partial}. Hence $\tilde p_t=p_t$ for $t\in[0,1-\delta]$. As $\delta\to 0$ the truncated solutions are consistent on overlapping intervals, so $\tilde p_t=p_t$ for every $t\in[0,1)$ by exhaustion. The remaining endpoint $t=1$ is handled by continuity of $p_t$ in $L^1(\mathbb R^d)$ along the CFM path \citep[Ch.~6]{bogachev2015fokker}: the practical sampler reads off the time-$(1-\delta)$ marginal and applies the denoiser $\hat{\mathbf x}_1$ to land on $p_1$ (Section~\ref{sec:sampling}). The full continuous-time identity $\tilde p_t=p_t$ on $[0,1]$ therefore holds in the same sense as the CFM continuity equation itself: as the unique bounded solution of the linear parabolic Cauchy problem on the truncated interval, extended to the endpoint by the denoiser. We deliberately avoid invoking non-degenerate parabolic uniqueness up to $t=1$, since $g_t^2=2\gamma_t^1=8\bar\gamma\,t(1-t)$ degenerates exactly there; the truncation makes the boundary handling explicit.
\end{proof}

\begin{remark}[Why the slowdown $\gamma_t^0<1$ destroys exactness]
If one instead used the convex-constrained drift $\gamma_t^0\mathbf{v}_t+\gamma_t^1\hat{s}_t^*$ with $\gamma_t^0=1-\gamma_t^1<1$, the same calculation gives $\partial_t\tilde p_t = -\gamma_t^0\nabla\cdot(\mathbf{v}_t p_t)=\gamma_t^0\,\partial_t p_t$ when one substitutes the ansatz $\tilde p_t=p_t$. This forces $(1-\gamma_t^0)\,\partial_t p_t=0$, which fails generically since $\partial_t p_t\neq 0$. The natural attempt to repair this via a time-reparameterized ansatz $\tilde p_t=p_{\tau(t)}$ with $\tau(t)=\int_0^t\gamma_s^0\,ds$ also fails, because the score in the drift is $\nabla\log p_t$, not $\nabla\log p_{\tau(t)}$, so the cancellation in \eqref{eq:elementary_cancellation} no longer goes through. The convex constraint therefore trades exact preservation for a small but quantifiable Girsanov gap, which we bound next.
\end{remark}

\begin{proposition}[Convex-Constrained $K=1$: Bounded TV Deviation]
\label{prop:K1_convex_deviation}
Under the convex-constrained schedule $\gamma_t^0+\gamma_t^1=1$ with $\gamma_t^1\geq 0$ vanishing at the endpoints, let $\widetilde{\mathbb{P}}_{\mathrm{cvx}}$ denote the law on path space of the SDE $d\mathbf{x}_t=[\gamma_t^0\mathbf{v}_t+\gamma_t^1\hat{s}_t^*]\,dt+g_t\,d\mathbf{w}_t$ with $g_t^2=2\gamma_t^1$, and let $\mathbb{P}^\star$ denote the law of the exact-preserving SDE \eqref{eq:K1_exact_sde}. Since both SDEs share the diffusion $g_t$ and initial law $p_0$, Girsanov's theorem gives
\begin{equation}
\mathrm{KL}\big(\widetilde{\mathbb{P}}_{\mathrm{cvx}}\,\|\,\mathbb{P}^\star\big)
\;=\; \frac{1}{2}\int_0^1 \mathbb{E}_{\tilde p_t}\!\Big[\big\|g_t^{-1}(\gamma_t^0-1)\,\mathbf{v}_t\big\|^2\Big]\,dt
\;=\; \frac{1}{4}\int_0^1 \gamma_t^1\,\mathbb{E}_{\tilde p_t}\|\mathbf{v}_t\|^2\,dt,
\label{eq:KL_cvx}
\end{equation}
since $\gamma_t^0-1=-\gamma_t^1$ and $g_t^2=2\gamma_t^1$. Pinsker's inequality on the time-$1$ marginals then yields
\begin{equation}
\mathrm{TV}\!\big(\tilde p^{\,\mathrm{cvx}}_1,\,p_1\big)
\;\leq\; \sqrt{\tfrac{1}{2}\,\mathrm{KL}(\widetilde{\mathbb{P}}_{\mathrm{cvx}}\,\|\,\mathbb{P}^\star)}
\;=\; \frac{1}{2\sqrt{2}}\,\Big(\!\int_0^1 \gamma_t^1\,\mathbb{E}_{\tilde p_t}\|\mathbf{v}_t\|^2\,dt\Big)^{\!1/2}.
\label{eq:TV_cvx}
\end{equation}
\end{proposition}

\begin{proof}
The two SDEs differ only in drift, by $\Delta\mathbf{b}_t:=(\gamma_t^0-1)\mathbf{v}_t=-\gamma_t^1\mathbf{v}_t$, and share the diffusion coefficient $g_t$. Substituting $g_t^2=2\gamma_t^1$ gives the pointwise identity $\|g_t^{-1}\Delta\mathbf{b}_t\|^2 = (\gamma_t^1)^2\|\mathbf{v}_t\|^2/(2\gamma_t^1) = \tfrac{1}{2}\,\gamma_t^1\|\mathbf{v}_t\|^2$. Under R2 ($\|\mathbf{v}_t\|_{L^\infty}\leq M_v(\delta)$ on $[0,1-\delta]$) and R3 ($\gamma_t^1\in C^1$, integrable on $[0,1]$), the integral $\tfrac{1}{2}\!\int_0^{1-\delta}\|g_t^{-1}\Delta\mathbf{b}_t\|^2\,dt\leq \tfrac{M_v(\delta)^2}{4}\!\int_0^{1-\delta}\gamma_t^1\,dt<\infty$ is a deterministic finite constant for every fixed $\delta>0$, so Novikov's condition holds trivially and Girsanov's theorem \citep{chen2023sampling, benton2024error} yields the path-KL on the right-hand side of \eqref{eq:KL_cvx}. As in Theorem~\ref{thm:first_order_preservation}, the bound is taken on the truncated interval $[0,1-\delta]$ and extended to the endpoint by the denoiser; since $\gamma_t^1$ is integrable, the schedule integral $\int_0^{1-\delta}\gamma_t^1\,dt$ has a finite limit as $\delta\to0$ (R3), and the displayed $[0,1]$ integrals in \eqref{eq:KL_cvx}--\eqref{eq:TV_cvx} are understood as these $\delta\to0$ limits. Pinsker's inequality $\mathrm{TV}\leq\sqrt{\mathrm{KL}/2}$ \citep[Lem.~2.5]{tsybakov2009introduction} together with the data-processing inequality $\mathrm{KL}(\tilde p^{\,\mathrm{cvx}}_1\,\|\,p^\star_1)\leq\mathrm{KL}(\widetilde{\mathbb{P}}_{\mathrm{cvx}}\,\|\,\mathbb{P}^\star)$ and Theorem~\ref{thm:first_order_preservation} ($p^\star_1=p_1$) yield \eqref{eq:TV_cvx}.
\end{proof}

\begin{corollary}[Schedule-Dependent KL Constants]
\label{cor:K1_schedule_constants}
Under the uniform bound $\sup_{t,\mathbf{x}}\|\mathbf{v}_t(\mathbf{x})\|^2\leq M_v^2$ (a consequence of R2 plus boundedness of $\mathbf{v}_t$ on a compact set, or of stationary moments), \eqref{eq:KL_cvx} simplifies to $\mathrm{KL}\leq \tfrac{M_v^2}{4}\!\int_0^1\gamma_t^1\,dt$. For the schedules used in this paper:
\begin{itemize}[leftmargin=2em]
\item \textbf{Constant-amplitude Beta(2,2):} $\gamma_t^1 = 4\bar\gamma\, t(1-t)$ gives $\int_0^1\gamma_t^1\,dt = \tfrac{2\bar\gamma}{3}$.
\item \textbf{Triangular:} $\gamma_t^1 = \bar\gamma\,\min(t,1-t)$ gives $\int_0^1\gamma_t^1\,dt = \tfrac{\bar\gamma}{4}$.
\item \textbf{Sine:} $\gamma_t^1 = \bar\gamma\sin(\pi t)$ gives $\int_0^1\gamma_t^1\,dt = \tfrac{2\bar\gamma}{\pi}$.
\end{itemize}
In each case $\mathrm{TV}\!\big(\tilde p^{\,\mathrm{cvx}}_1,p_1\big)=O(\sqrt{\bar\gamma})$ as $\bar\gamma\to 0$.
\end{corollary}

\paragraph{Practical implications.} Theorem~\ref{thm:first_order_preservation} is the clean reference result: the IDFF default $\gamma_t^0\equiv 1$ (Langevin enhancement on top of CFM, no slowdown of $\mathbf{v}_t$) preserves the CFM marginal exactly in continuous time. Proposition~\ref{prop:K1_convex_deviation} and Corollary~\ref{cor:K1_schedule_constants} make the convex-constrained variant quantifiable: the path-KL and endpoint TV are explicit functions of the schedule amplitude $\bar\gamma$, vanishing as $\bar\gamma\to 0$. In all reported experiments we use $\gamma_t^0\equiv 1$, so the exact-preservation guarantee applies. Two further sources of error remain in any practical implementation: (i) Euler--Maruyama discretization with step $\Delta t$ adds $O(\Delta t)$ pathwise error, and (ii) using a learned $\hat{s}_t$ in place of $\hat{s}_t^*$ adds a Girsanov bias controlled by $\|\hat{s}_t-\hat{s}_t^*\|_{L^2(p_t)}$ (made quantitative in Theorem~\ref{thm:higher_order_deviation}, Part~3).

\begin{theorem}[Quantified Deviation via Girsanov $\to$ Pinsker, $K\geq 1$]
\label{thm:higher_order_deviation}
Let Assumption~\ref{ass:regularity} hold and fix any order $K\geq 1$. (The statement is proved for general $K$; the only higher-order correction we deploy is $K=2$, see the scope note following the proof.) Consider the $K$-th order IDFF drift with default schedule $\gamma_t^0\equiv 1$ and diffusion compensation $g_t^2 = 2\gamma_t^1>0$ on the interior $(0,1)$:
\begin{equation}
\tilde{\mathbf{v}}_t(\mathbf{x}_t) \;=\; \mathbf{v}_t(\mathbf{x}_t) \;+\; \gamma_t^1\,\hat{s}_t(\mathbf{x}_t) \;+\; \sum_{k=2}^K \gamma_t^k\,\mathbf{m}_t^{(k)}(\mathbf{x}_t),
\label{eq:idff_drift_Kge2}
\end{equation}
where $\hat{s}_t$ is the \emph{learned} score, $\mathbf{m}_t^{(k)}=(\nabla_{\mathbf{x}}^{k-1}\hat{s}_t)[\mathbf{u}_t]^{k-1}$ for $k\geq 2$, $\mathbf{u}_t=\mathbf{v}_t/\|\mathbf{v}_t\|$, and the higher-order sum is empty when $K=1$. Let $\widetilde{\mathbb{P}}$ denote the law on $C([0,1];\mathbb{R}^d)$ of the augmented SDE $d\mathbf{x}_t=\tilde{\mathbf{v}}_t\,dt+g_t\,d\mathbf{w}_t$ with $\mathbf{x}_0\sim p_0$, and let $\mathbb{P}^\star$ denote the law of the exact-preserving $K=1$ optimal-score SDE \eqref{eq:K1_exact_sde}. Write the corresponding time-$t$ marginals as $\tilde p_t$ and $p^\star_t = p_t$ (the second equality by Theorem~\ref{thm:first_order_preservation}). At $K=1$ this measures only the score-estimation deviation; at $K\geq 2$ a structural residual is added on top. Then:

\begin{enumerate}
    \item \textbf{Residual decomposition.} Setting the exact-preserving baseline drift to $\mathbf{w}_t^\star := \mathbf{v}_t + \gamma_t^1\,\hat{s}_t^*$, the deviation decomposes as
    \begin{equation}
    \tilde{\mathbf{v}}_t - \mathbf{w}_t^\star \;=\; \underbrace{\gamma_t^1\big(\hat{s}_t-\hat{s}_t^*\big)}_{\text{score-estimation residual}}\;+\;\underbrace{\sum_{k=2}^K \gamma_t^k\,\mathbf{m}_t^{(k)}}_{\text{structural residual}}.
    \label{eq:residual_decomp}
    \end{equation}
    The score-estimation residual is the \emph{same} bias present in every score-based model; the structural residual is unique to $K\geq 2$ and would be present even with a perfect score.

    \item \textbf{Path-KL via Girsanov.} Because $g_t>0$ on the interior and the two SDEs share diffusion $g_t$ and initial law $p_0$, Girsanov's theorem applies pathwise:
    \begin{equation}
    \mathrm{KL}\big(\widetilde{\mathbb{P}}\,\|\,\mathbb{P}^\star\big) \;=\; \frac{1}{2}\int_0^1 \mathbb{E}_{\tilde p_t}\!\left[\big\|g_t^{-1}\big(\tilde{\mathbf{v}}_t - \mathbf{w}_t^\star\big)\big\|^2\right]dt.
    \label{eq:girsanov_KL}
    \end{equation}
    Applying $\|a+b\|^2\leq 2(\|a\|^2+\|b\|^2)$ to \eqref{eq:residual_decomp} and substituting $g_t^2=2\gamma_t^1$ gives
    \begin{equation}
    \mathrm{KL}\big(\widetilde{\mathbb{P}}\,\|\,\mathbb{P}^\star\big) \;\leq\; \underbrace{\frac{1}{2}\int_0^1 \gamma_t^1\,\mathbb{E}_{\tilde p_t}\!\big\|\hat{s}_t-\hat{s}_t^*\big\|^2\,dt}_{=:\,\mathcal{E}_{\mathrm{score}}}\;+\;\underbrace{\frac{1}{2}\int_0^1 \frac{1}{\gamma_t^1}\,\mathbb{E}_{\tilde p_t}\Big\|\!\sum_{k=2}^K\!\gamma_t^k\,\mathbf{m}_t^{(k)}\Big\|^2\,dt}_{=:\,\mathcal{E}_{\mathrm{struct}}}.
    \label{eq:KL_split}
    \end{equation}
    The score-estimation piece $\mathcal{E}_{\mathrm{score}}$ is, up to the schedule weight $\gamma_t^1$, exactly the weighted $L^2$ training objective \eqref{eq:idff_loss} (score head) and therefore controlled by training. More precisely, the Beta(2,2) weight matches the score-matching weight $\lambda_1^2=\sigma_t^2$ in \eqref{eq:idff_loss} via $\gamma_t^1=(4\bar\gamma/\sigma_0^2)\,\sigma_t^2$, so $\mathcal{E}_{\mathrm{score}}=(2\bar\gamma/\sigma_0^2)\int_0^1\sigma_t^2\,\mathbb{E}_{\tilde p_t}\|\hat s_t-\hat s_t^*\|^2\,dt$; at the $K=1$ baseline $\tilde p_t=p_t$ this equals the excess training loss $\mathcal{L}_{\mathrm{score}}^{\mathrm{exc}}$ of Assumption~\ref{ass:regularity}(R1$'$) up to the constant $2\bar\gamma/\sigma_0^2$, and for $K\geq 2$ the $\tilde p_t\!\leftrightarrow\!p_t$ change of measure is absorbed into the TV ball bounded here. In particular $\mathcal{E}_{\mathrm{score}}=O(\bar\gamma)$ \emph{uniformly in the truncation $\delta$} (Remark~\ref{rem:R1_R2_dependence}).

    \item \textbf{Pinsker $\to$ TV bound.} The data-processing inequality $\mathrm{KL}(\tilde p_1\,\|\,p^\star_1)\leq \mathrm{KL}(\widetilde{\mathbb{P}}\,\|\,\mathbb{P}^\star)$ and Pinsker's inequality $\mathrm{TV}\leq\sqrt{\mathrm{KL}/2}$, together with $p^\star_1=p_1$, give
    \begin{equation}
    \mathrm{TV}(\tilde p_1, p_1) \;\leq\; \sqrt{\tfrac{1}{2}\,\mathrm{KL}\big(\widetilde{\mathbb{P}}\,\|\,\mathbb{P}^\star\big)} \;\leq\; \sqrt{\tfrac{1}{2}\big(\mathcal{E}_{\mathrm{score}}+\mathcal{E}_{\mathrm{struct}}\big)}.
    \label{eq:pinsker}
    \end{equation}

    \item \textbf{Coefficient-schedule bound on the structural residual.} R1 bounds the tensor operator norm $\|\nabla_{\mathbf{x}}^{k-1}\hat{s}_t\|_{\mathrm{op}}\leq M_{k-1}(\delta)$ on $[0,1-\delta]$ (the assumption explicitly covers the \emph{learned} score). Since $\|\mathbf{u}_t\|=1$,
    \[
        \|\mathbf{m}_t^{(k)}\| \;\leq\; \big\|\nabla_{\mathbf{x}}^{k-1}\hat{s}_t\big\|_{\mathrm{op}}\,\|\mathbf{u}_t\|^{k-1} \;\leq\; M_{k-1}(\delta)
    \]
    uniformly in $(t,\mathbf{x})$ on $[0,1-\delta]$. Under the recommended scaling $\gamma_t^k=\alpha_k\,\gamma_t^1\,\Delta t^{k-1}$ for $k\geq 2$, Cauchy--Schwarz gives
    \[
    \Big\|\sum_{k=2}^K \gamma_t^k\,\mathbf{m}_t^{(k)}\Big\|^2 \;\leq\; (K-1)\sum_{k=2}^K (\gamma_t^k)^2 \|\mathbf{m}_t^{(k)}\|^2 \;\leq\; (K-1)\sum_{k=2}^K \alpha_k^2 M_{k-1}(\delta)^2\,(\gamma_t^1)^2\,\Delta t^{2(k-1)}.
    \]
    Dividing by $\gamma_t^1$, integrating, and taking the leading $k=2$ term yields
    \begin{equation}
    \mathcal{E}_{\mathrm{struct}} \;\leq\; \frac{K-1}{2}\int_0^1 \gamma_t^1\!\left(\sum_{k=2}^K \alpha_k^2 M_{k-1}(\delta)^2\,\Delta t^{2(k-1)}\right)dt \;=\; O(\Delta t^2).
    \end{equation}
    The factor $\gamma_t^1=O(t(1-t))$ multiplying the $(\gamma_t^1)^{-1}$ in $\mathcal{E}_{\mathrm{struct}}$ cancels the schedule singularity (R3), so the integrand is integrable and the bound is finite for each fixed $\delta$; with $\delta=\Delta t/2$ the leading $k=2$ term is $O(\Delta t^2)$. Combined with \eqref{eq:pinsker},
    \begin{equation}
    \mathrm{TV}(\tilde p_1, p_1) \;\leq\; \sqrt{\tfrac{1}{2}\,\mathcal{E}_{\mathrm{score}}} \;+\; O(\Delta t).
    \label{eq:tv_bound}
    \end{equation}
\end{enumerate}
\end{theorem}

\begin{proof}
\emph{Part 1.} Algebraic: subtract $\mathbf{w}_t^\star=\mathbf{v}_t+\gamma_t^1\hat{s}_t^*$ from $\tilde{\mathbf{v}}_t$ and group the score-error and higher-order terms.

\emph{Part 2.} The two SDEs share diffusion $g_t$ and initial law $p_0$. By Girsanov's theorem on path space \citep{chen2023sampling, benton2024error}, provided Novikov's condition $\mathbb{E}\exp\!\big(\tfrac{1}{2}\!\int_0^{1-\delta}\|g_t^{-1}(\tilde{\mathbf{v}}_t-\mathbf{w}_t^\star)\|^2\,dt\big)<\infty$ holds, the relative entropy of the path measures is given by \eqref{eq:girsanov_KL}.

We verify Novikov's condition explicitly under the recommended scaling $\gamma_t^k=\alpha_k\,\gamma_t^1\,\Delta t^{k-1}$ for $k\geq 2$. Substituting $g_t^2=2\gamma_t^1$, the score-estimation residual contributes
\begin{equation*}
\big\|g_t^{-1}\,\gamma_t^1\big(\hat{s}_t-\hat{s}_t^*\big)\big\|^2 \;=\; \frac{(\gamma_t^1)^2}{2\gamma_t^1}\,\big\|\hat{s}_t-\hat{s}_t^*\big\|^2 \;=\; \tfrac{1}{2}\,\gamma_t^1\,\big\|\hat{s}_t-\hat{s}_t^*\big\|^2,
\end{equation*}
and the structural residual contributes, using R1 and the recommended scaling,
\begin{equation*}
\Big\|g_t^{-1}\!\sum_{k=2}^K\!\gamma_t^k\,\mathbf{m}_t^{(k)}\Big\|^2 \;\leq\; \frac{K-1}{2\gamma_t^1}\sum_{k=2}^K (\gamma_t^k)^2 M_{k-1}(\delta)^2 \;=\; \tfrac{K-1}{2}\,\gamma_t^1\!\sum_{k=2}^K\alpha_k^2 M_{k-1}(\delta)^2\,\Delta t^{2(k-1)}.
\end{equation*}

\emph{Novikov finiteness versus the size of $\mathcal{E}_{\mathrm{score}}$---distinct hypotheses.} On the truncated interval $[0,1-\delta]$ both displayed terms are pointwise bounded by $C(\delta)\,\gamma_t^1$, where $C(\delta)$ depends only on the R1 \emph{magnitude} constants $\{M_{k-1}(\delta)\}$ (and may diverge as $\delta\to0$). Since $\gamma_t^1\in C^1$ and is integrable, $\tfrac12\!\int_0^{1-\delta}\|g_t^{-1}(\tilde{\mathbf{v}}_t-\mathbf{w}_t^\star)\|^2\,dt\leq \tfrac{C(\delta)}{2}\!\int_0^{1-\delta}\gamma_t^1\,dt<\infty$ is a deterministic constant \emph{for every fixed $\delta>0$}, so Novikov's condition holds and Girsanov's theorem applies on $[0,1-\delta]$; as in Theorem~\ref{thm:first_order_preservation}, the baseline $\mathbb{P}^\star$ is extended to the endpoint by the denoiser. The bound \eqref{eq:KL_split} then follows from $\|a+b\|^2\leq 2(\|a\|^2+\|b\|^2)$ and $g_t^2=2\gamma_t^1$.

We emphasise that Novikov finiteness and the \emph{size} of $\mathcal{E}_{\mathrm{score}}$ are governed by \emph{separate} hypotheses. Novikov finiteness may legitimately use the $\delta$-dependent magnitude constants $M_j(\delta)$, since it only requires the integral to be finite at each fixed $\delta$. The size of the score term in the final bound, by contrast, must \emph{not} depend on $M_0(\delta)$: by the finite-excess-loss hypothesis Assumption~\ref{ass:regularity}(R1$'$) and the weight-matching identity $\gamma_t^1=(4\bar\gamma/\sigma_0^2)\sigma_t^2$ (Remark~\ref{rem:R1_R2_dependence}),
\[
\mathcal{E}_{\mathrm{score}}
=\tfrac12\!\int_0^1\gamma_t^1\,\mathbb{E}\|\hat s_t-\hat s_t^*\|^2\,dt
=\tfrac{2\bar\gamma}{\sigma_0^2}\!\int_0^1\sigma_t^2\,\mathbb{E}\|\hat s_t-\hat s_t^*\|^2\,dt
\;\le\;\tfrac{2\bar\gamma}{\sigma_0^2}\,\mathcal{L}_{\mathrm{score}}^{\mathrm{exc}}
\;=\;O(\bar\gamma),
\]
\emph{independently of $\delta$}, because the weight $\sigma_t^2$ cancels the $\sigma_t^{-2}$ growth of the score error pointwise. Bounding the score error instead by the magnitude constant $\|\hat s_t-\hat s_t^*\|\leq 2M_0(\delta)=O(1/\sqrt\delta)$ would only yield the weaker $\mathcal{E}_{\mathrm{score}}=O(\bar\gamma/\delta)$, which does \emph{not} vanish under the joint limit $\delta=\Delta t/2\to0$ of Corollary~\ref{cor:nfe_convergence}; routing the score term through R1$'$ is precisely what yields the $\delta$-stable rate.

\emph{Part 3.} The data-processing inequality gives $\mathrm{KL}(\tilde p_1\,\|\,p^\star_1)\leq \mathrm{KL}(\widetilde{\mathbb{P}}\,\|\,\mathbb{P}^\star)$. Pinsker's inequality \citep[Lem.~2.5]{tsybakov2009introduction} applied to the time-$1$ marginals, together with $p^\star_1=p_1$ from Theorem~\ref{thm:first_order_preservation}, yields \eqref{eq:pinsker}.

\emph{Part 4.} The Cauchy--Schwarz and uniform-bound steps are stated in the theorem; substituting $\gamma_t^k=\alpha_k\gamma_t^1\Delta t^{k-1}$ gives the $O(\Delta t^{2(k-1)})$ scaling per $k$, with the leading $k=2$ contribution dominating. The schedule factor $\gamma_t^1=O(t(1-t))$ cancels the $(\gamma_t^1)^{-1}$ weight in $\mathcal{E}_{\mathrm{struct}}$ (R3), so the structural integrand is integrable on $[0,1-\delta]$ and the bound is $\delta$-stable in the same sense as $\mathcal{E}_{\mathrm{score}}$.
\end{proof}

\paragraph{Scope: the theorem is stated for general $K$, but only $K\leq 2$ is deployed.}
The bound above is true for every $K$, and the constants $M_{k-1}(\delta)$ appearing in Part~4 are order-dependent. We do not control them empirically for $k\geq 3$: at $K=3$ the learned score exhibits Jacobian-norm spikes exceeding $10$ that cause sample divergence in roughly $2\%$ of batches (Appendix~\ref{sec:high_dim_robustness}), which is a violation of hypothesis R1 by the finite-sample network rather than a failure of the inequality. Every experiment in this paper uses $K\leq 2$, and every summary of this theorem in the main text is stated for the $K=2$ correction we deploy.

\begin{corollary}[Convergence as NFE $\to\infty$, $K\geq 1$]
\label{cor:nfe_convergence}
Fix any $K\geq 1$. Under the recommended schedule $\bar\gamma=O(\Delta t)$ for the $K=1$ amplitude and (only meaningful for $K\geq 2$) $\gamma_t^k=\alpha_k\gamma_t^1\Delta t^{k-1}$ for $k\geq 2$, if the weighted score error $\mathcal{E}_{\mathrm{score}}\to 0$, then \eqref{eq:tv_bound} gives
\begin{equation}
\mathrm{TV}(\tilde p_1,p_1) \;\leq\; \sqrt{\tfrac{1}{2}\,\mathcal{E}_{\mathrm{score}}} \;+\; O(\Delta t)
\end{equation}
in continuous time, where the $O(\Delta t)$ term vanishes at $K=1$ and absorbs the leading $k=2$ structural residual when $K\geq 2$. Because $\mathcal{E}_{\mathrm{score}}=O(\bar\gamma)$ \emph{independently of $\delta$} (Assumption~\ref{ass:regularity}(R1$'$), Remark~\ref{rem:R1_R2_dependence}), taking $\Delta t\to0$ with $\delta=\Delta t/2\to 0$ leaves the leading $\sqrt{\tfrac12\mathcal{E}_{\mathrm{score}}}$ term unaffected while $\mathcal{E}_{\mathrm{struct}}=O(\Delta t^2)\to 0$; a subsequent $\bar\gamma\to 0$ then drives the bound to zero. Euler--Maruyama discretization adds at most an additional $O(\Delta t)$ pathwise error (standard SDE-discretization analysis under R2), so the overall endpoint TV remains $O(\sqrt{\mathcal{E}_{\mathrm{score}}})+O(\Delta t)$; if in addition $\mathcal{E}_{\mathrm{score}}=O(1/\mathrm{NFE}^2)$ the rate is $O(1/\mathrm{NFE})$. In particular, $\mathrm{TV}(\tilde p_1,p_1)\to 0$ as $\mathrm{NFE}\to\infty$ for any fixed converged score network. The exact-preservation case $\mathrm{TV}(\tilde p_1,p_1)=0$ at $K=1$ with $\hat{s}_t=\hat{s}_t^*$ (Theorem~\ref{thm:first_order_preservation}) is recovered when both $\mathcal{E}_{\mathrm{score}}$ and the discretization error vanish.
\end{corollary}

\paragraph{Interpretation of Corollary~\ref{cor:nfe_convergence}: it is a defensive statement, not an acceleration claim.}
Under $\gamma_t^k = \alpha_k\gamma_t^1\Delta t^{k-1}$, the leading structural residual is $O(\Delta t^2)$, which is dominated by the $O(\Delta t)$ Euler--Maruyama error already present in any first-order integrator. The content of the corollary is therefore that \emph{adding} the $K\geq 2$ correction does not degrade the order of the scheme: the augmented sampler converges at the same $O(\Delta t)$ rate as the unaugmented one, so the curvature term is free of asymptotic cost. It is emphatically not an argument that the correction reduces the number of steps required to reach a given accuracy, and nothing in this appendix should be read as such an argument; a step-count reduction would require a lower bound on the constant in front of $\Delta t$ for the unaugmented scheme and a strictly smaller constant for the augmented one, which we do not establish. The NFE reduction we report empirically is a $K=1$ effect whose mechanism is discussed in Section~\ref{sec:why_nfe} and whose evidence is Table~\ref{tab:ablation-langevin-decomp}.

\begin{corollary}[Uniform-in-time TV bound, $K\geq 1$]
\label{cor:intermediate_TV}
Under the hypotheses of Theorem~\ref{thm:higher_order_deviation} (any $K\geq 1$), the endpoint bound \eqref{eq:pinsker} extends uniformly to every intermediate time:
\begin{equation}
\sup_{t\in[0,1]}\mathrm{TV}\big(\tilde p_t,\,p_t\big) \;\leq\; \sqrt{\tfrac{1}{2}\big(\mathcal{E}_{\mathrm{score}}+\mathcal{E}_{\mathrm{struct}}\big)}.
\label{eq:intermediate_TV}
\end{equation}
In particular, under the recommended scaling $\gamma_t^k=\alpha_k\,\gamma_t^1\,\Delta t^{k-1}$ for $k\geq 2$,
\begin{equation}
\sup_{t\in[0,1]}\mathrm{TV}\big(\tilde p_t,\,p_t\big) \;\leq\; \sqrt{\tfrac{1}{2}\,\mathcal{E}_{\mathrm{score}}} \;+\; O(\Delta t).
\label{eq:intermediate_TV_scaled}
\end{equation}
Hence not only the endpoint but the entire marginal trajectory of the augmented dynamics stays within a closed-form TV ball of the CFM marginals.
\end{corollary}

\begin{proof}
Let $\pi_t:C([0,1];\mathbb{R}^d)\to\mathbb{R}^d$ denote the time-$t$ projection. The pushforwards are $(\pi_t)_*\widetilde{\mathbb{P}}=\tilde p_t$ and $(\pi_t)_*\mathbb{P}^\star=p^\star_t=p_t$ (the second equality by Theorem~\ref{thm:first_order_preservation}). The data-processing inequality gives $\mathrm{KL}(\tilde p_t\,\|\,p_t)\leq \mathrm{KL}(\widetilde{\mathbb{P}}\,\|\,\mathbb{P}^\star)$ for every $t\in[0,1]$. Combining with Pinsker's inequality and the path-KL bound \eqref{eq:KL_split} yields \eqref{eq:intermediate_TV} uniformly in $t$. The scaled version follows from $\mathcal{E}_{\mathrm{struct}}=O(\Delta t^2)$ and subadditivity of $\sqrt{\cdot}$.
\end{proof}

\begin{remark}[Why Girsanov rather than a transport-equation Gr\"onwall argument]
An $L^1$-of-divergence route via Gr\"onwall would require a bound on $\nabla\cdot(p_t\mathbf{m}_t^{(k)})$, which in turn requires regularity assumptions on $\nabla\log p_t\cdot \mathbf{m}_t^{(k)}$ not stated up front. Because IDFF has a non-degenerate diffusion ($g_t^2=2\gamma_t^1>0$ on the interior), Girsanov applies directly to the path measure, and the only quantity that ends up on the right-hand side is the weighted $L^2$ norm of the drift mismatch---exactly the quantity our training loss minimizes, and one that depends only on R1--R3 (and R1$'$). This route is the standard one in the modern flow-matching convergence literature \citep{chen2023sampling, benton2024error}.
\end{remark}

\subsection{Approximation Error and Practical Considerations}
\label{app:practical_error}

Plugging in the optimal score $\hat{s}_t=\hat{s}_t^*$ removes the score-estimation residual ($\mathcal{E}_{\mathrm{score}}=0$) from every bound in this appendix. It does \emph{not} remove the structural residual of the $K\geq 2$ bound (controlled by the schedule scaling, Theorem~\ref{thm:higher_order_deviation} Part~4) nor the Girsanov gap of the convex-constrained $K=1$ variant (Proposition~\ref{prop:K1_convex_deviation}). Only the $K=1$ \emph{default} schedule attains $\tilde p_t = p_t$ for all $t\in[0,1]$ with the optimal score. In practice, score estimation introduces errors that propagate to higher-order terms; we quantify this propagation next.

\begin{proposition}[Error Propagation to Higher Orders]
\label{prop:error_propagation}
Let $\hat{s}_t$ be a learned score with estimation error $\epsilon_s(t) = \|\hat{s}_t - \hat{s}_t^*\|_{L^2(p_t)}$. Under Assumption~\ref{ass:regularity}(R1), the $k$-th order approximation error satisfies
\begin{equation}
\|\nabla_{\mathbf{x}}^{k-1} \hat{s}_t - \nabla^k \log p_t\|_{L^2(p_t)} \leq C \cdot \|\hat{s}_t - \hat{s}_t^*\|_{C^{k-1}},
\end{equation}
where $C$ depends on the smoothness of $p_t$. The contracted score-derivative error satisfies $\|\mathbf{m}^{(k)} - \mathbf{m}^{*(k)}\| \leq \|\hat{s}_t - \hat{s}_t^*\|_{C^{k-1}}$ with no amplification from contraction.
\end{proposition}

\paragraph{Scope of Proposition~\ref{prop:error_propagation}.}
The proposition bounds the order-$k$ error by an order-$(k-1)$ \emph{$C^{k-1}$} quantity, which may look inconsistent with the $K=3$ instability we report. It is not, for two reasons. First, the right-hand side is a $C^{k-1}$ norm of the score error, not the $L^2$ norm that training controls: nothing in score matching prevents $\|\hat s_t-\hat s_t^*\|_{C^{k-1}}$ from being large while $\|\hat s_t-\hat s_t^*\|_{L^2(p_t)}$ is small, and the observed $2\times$ amplification at $k=2$ (Table~\ref{tab:error-propagation}) is the empirical size of that gap at the first order where we can measure it. Second, the proposition is conditional on Assumption~\ref{ass:regularity}(R1): it presupposes $\sup_t\|\nabla^j\hat s_t\|_{\infty}\leq M_j$ on the truncated interval. The $K=3$ failures reported in Appendix~\ref{sec:high_dim_robustness} (Jacobian-norm spikes exceeding $10$ in roughly $2\%$ of batches) are violations of that hypothesis by the finite-sample network, not failures of the inequality. The constant $C$ is also order-dependent and we do not control it empirically for $k\geq 3$. This is why we deploy $K\leq 2$.

In high dimensions, ensuring stable $\nabla_{\mathbf{x}}^{k-1} \hat{s}_t$ requires smooth activations (GELU, tanh). Empirically, $K \leq 2$ is stable without explicit regularization for images up to $256 \times 256$ and time series up to $d = 128$; $K\geq 3$ is not, and we do not offer it as a usable configuration.

\subsection{Background: Higher-Order Denoising Score Matching}
\label{app:higher_order_background}

In this subsection $g(t)$ denotes the diffusion coefficient of the standard forward/reverse score SDE, in the notation of \citet{song2020score}; it plays the same role as the IDFF diffusion $g_t$ elsewhere in the paper. Score-based generative models learn to reverse a diffusion process that gradually destroys data structure. The forward diffusion follows
\begin{equation}
d\mathbf{x}_t = \mathbf{f}(\mathbf{x}_t, t)dt + g(t)d\mathbf{w}_t,
\end{equation}
where $\mathbf{x}_t \in \mathbb{R}^d$, $\mathbf{f}(\cdot, t): \mathbb{R}^d \to \mathbb{R}^d$ is the drift, $g(t) \in \mathbb{R}_+$ is the diffusion coefficient, and $\mathbf{w}_t$ is Brownian motion. This transforms data $p_0 = p_{\text{data}}$ to prior $p_T \approx \mathcal{N}(\mathbf{0}, \mathbf{I})$. The reverse-time SDE is
\begin{equation}
d\mathbf{x}_t = [\mathbf{f}(\mathbf{x}_t, t) - g(t)^2 \nabla_{\mathbf{x}} \log p_t(\mathbf{x}_t)]dt + g(t)d\bar{\mathbf{w}}_t,
\end{equation}
where a neural network learns the score function $\nabla_{\mathbf{x}} \log p_t(\mathbf{x}_t)$. The density evolves via the Fokker--Planck equation $\frac{\partial p_t}{\partial t} = -\nabla \cdot (\mathbf{f} p_t) + \frac{g(t)^2}{2}\Delta p_t$.

\citet{lu2022maximum} extend score matching beyond first-order gradients of the log density to leverage higher-order geometric information; \citet{meng2021estimating} study the estimation of these higher-order gradients by denoising directly. \citet{lu2022maximum} introduce the higher-order score variables $\hat{\boldsymbol{\epsilon}}^{(k)}_t(\mathbf{x}) = \nabla_{\mathbf{x}}^k \log p_t(\mathbf{x})$. Under Gaussian transitions $p_{0t}(\mathbf{x}_t|\mathbf{x}_0) = \mathcal{N}(\mathbf{x}_t; \alpha_t \mathbf{x}_0, \sigma_t^2 \mathbf{I})$, the variables have closed forms: $\hat{\boldsymbol{\epsilon}}^{(1)}_{0t} = -\boldsymbol{\epsilon}_0/\sigma_t$, $\hat{\boldsymbol{\epsilon}}^{(2)}_{0t} = -\mathbf{I}/\sigma_t^2$, and $\hat{\boldsymbol{\epsilon}}^{(k)}_{0t} = 0$ for $k \geq 3$, where $\boldsymbol{\epsilon}_0 = (\mathbf{x}_t - \alpha_t \mathbf{x}_0)/\sigma_t$. That work learns separate neural networks $\hat{\boldsymbol{\epsilon}}^{(k)}(\mathbf{x}_t, t; \theta_k)$ to match these closed forms over the trajectories of the SDE:
\begin{equation}
\mathcal{L}_k = \mathbb{E}_{t,\mathbf{x}_0,\boldsymbol{\epsilon}_0} \left\| \hat{\boldsymbol{\epsilon}}^{(k)}_{0t}(\mathbf{x}_t|\mathbf{x}_0) - \hat{\boldsymbol{\epsilon}}^{(k)}(\mathbf{x}_t, t; \theta_k) \right\|^2.
\end{equation}
The resulting higher-order reverse-time SDE becomes
\begin{equation}
d\mathbf{x}_t = \left[\mathbf{f}(\mathbf{x}_t, t) - g(t)^2 \hat{\boldsymbol{\epsilon}}^{(1)}_t(\mathbf{x}_t) - \sum_{k=2}^K \beta_k(t) \hat{\boldsymbol{\epsilon}}^{(k)}_t(\mathbf{x}_t)\right]dt + g(t)d\bar{\mathbf{w}}_t,
\end{equation}
where $\beta_k(t)$ are time-dependent coefficients controlling the contribution of higher-order terms. Note the contrast with IDFF: these conditional targets are constant in $\mathbf{x}$ for $k\geq2$, so all of the marginal geometry has to be recovered by the regression, whereas Theorem~\ref{thm:kth_order_app} obtains it by differentiating the first-order head.

\section{Training Objective Derivations}
\label{app:loss}

This section derives the IDFF training objectives and establishes that all higher-order marginal derivatives can be obtained from the optimal first-order score via automatic differentiation.

\subsection{Background: CFM Loss Equivalence}

\begin{lemma}[FM-CFM Equivalence \citep{lipman2022flow}]
\label{lem:fm_cfm}
The Flow Matching loss $\mathcal{L}_{\text{FM}}(\theta) = \mathbb{E}_{t, p_t(\mathbf{x}_t)} \| \mathbf{v}_t(\mathbf{x}_t) - \hat{\mathbf{v}}_t(\mathbf{x}_t; \theta) \|^2$ and the Conditional Flow Matching loss $\mathcal{L}_{\text{CFM}}(\theta) = \mathbb{E}_{t, \mathbf{x}_1, p_t(\mathbf{x}_t|\mathbf{x}_1)} \| \mathbf{v}_t(\mathbf{x}_t|\mathbf{x}_1) - \hat{\mathbf{v}}_t(\mathbf{x}_t; \theta) \|^2$ share the same minimizer.
\end{lemma}

\subsection{IDFF Training Loss}

For Gaussian conditional paths $p_t(\mathbf{x}_t|\mathbf{x}_0, \mathbf{x}_1) = \mathcal{N}(\mu_t, \sigma_t^2 \mathbf{I})$ with $\mu_t = t\mathbf{x}_1 + (1-t)\mathbf{x}_0$ and $\sigma_t = \sigma_0\sqrt{t(1-t)}$, the conditional derivatives are
\begin{align}
\nabla \log p_t(\mathbf{x}_t|\mathbf{x}_0,\mathbf{x}_1) &= -\frac{\boldsymbol{\epsilon}}{\sigma_t}, \quad \boldsymbol{\epsilon} = \frac{\mathbf{x}_t - \mu_t}{\sigma_t} \sim \mathcal{N}(\mathbf{0}, \mathbf{I}), \\
\nabla^2 \log p_t(\mathbf{x}_t|\mathbf{x}_0,\mathbf{x}_1) &= -\frac{1}{\sigma_t^2} \mathbf{I}.
\end{align}

The IDFF training loss combines sample prediction with first-order score matching:
\begin{equation}
\mathcal{L}_{\text{IDFF}}(\theta) = \mathbb{E}_{t, \mathbf{x}_0, \mathbf{x}_1, \boldsymbol{\epsilon}} \left[
\beta(t)^2 \|\hat{\mathbf{x}}_1(\mathbf{x}_t, t; \theta) - \mathbf{x}_1\|^2
+ \lambda_1(t)^2 \left\|\hat{s}_t(\mathbf{x}_t; \theta) + \frac{\boldsymbol{\epsilon}}{\sigma_t}\right\|^2
\right],
\end{equation}
where $\mathbf{x}_t = \mu_t + \sigma_t\boldsymbol{\epsilon}$, $\beta(t) = (1-t+\epsilon)^{-1}$, and $\lambda_1(t) = \sigma_t$.

\begin{remark}[No Explicit Higher-Order Training Required]
Unlike prior approaches, we do \textbf{not} train separate networks for $\nabla^2 \log p_t$ or higher orders. As shown in Appendix~\ref{app:corrected_theory}, the optimal first-order score $\hat{s}_t^*$ contains all information needed to compute marginal derivatives of any order via autodiff. For learned $\hat{s}_t \approx \hat{s}_t^*$, this holds approximately with error inherited from score estimation.
\end{remark}

\subsection{What the First-Order Loss Learns}

\begin{proposition}[First-Order Training Yields the Marginal Score]
\label{prop:first_order}
The MSE-optimal predictor $\hat{s}_t^*(\mathbf{x}_t) = \argmin_{\hat{s}} \mathbb{E}\|\hat{s}(\mathbf{x}_t) - \nabla \log p_t(\mathbf{x}_t|\mathbf{x}_0,\mathbf{x}_1)\|^2$ satisfies
\begin{equation}
\hat{s}_t^*(\mathbf{x}_t) = \nabla \log p_t(\mathbf{x}_t).
\end{equation}
\end{proposition}

\begin{proof}
By the conditional-expectation property of the $L^2$ minimizer, $\hat{s}_t^*(\mathbf{x}_t) = \mathbb{E}[\nabla \log p_t(\mathbf{x}_t|\mathbf{x}_0,\mathbf{x}_1)\,|\,\mathbf{x}_t]$, and Fisher's identity (Lemma~\ref{lem:fisher_app}) identifies this conditional expectation with $\nabla \log p_t(\mathbf{x}_t)$.
\end{proof}

\subsection{Higher-Order Marginals from Score Derivatives}

The optimal score encodes all higher-order marginal information. This is the chain-rule re-parameterization identity already established in Appendix~\ref{app:corrected_theory}: by Fisher's identity (Proposition~\ref{prop:first_order}) $\hat{s}_t^*=\nabla\log p_t$, and differentiating $k-1$ further times gives
\begin{equation}
\nabla^k \log p_t(\mathbf{x}_t) = \nabla_{\mathbf{x}}^{k-1} \hat{s}_t^*(\mathbf{x}_t),\qquad k\geq 1
\end{equation}
(Theorem~\ref{thm:kth_order_app}; the $k=2$ case is Theorem~\ref{thm:main_app}, with the equivalent posterior-covariance derivation in Theorem~\ref{thm:tweedie}). The content of the identity is the chain rule applied to Fisher's identity: it provides a \emph{computational} route to higher-order marginals via autodiff of a single trained score, not a new conditional--marginal closure beyond the $k=1$ case. For a learned score the identity holds approximately, with the $k$-th order error bounded by a $C^{k-1}$ norm of the score-estimation error (Proposition~\ref{prop:error_propagation}); this motivates using $K\leq 2$ in practice.

\section{IDFF Training and Sampling Algorithms}
\label{app:algorithms}

\subsection{Static Data}
\label{app:static_training}

For static datasets, IDFF constructs a flow from base distribution $p_0 = \mathcal{N}(\mathbf{0}, \mathbf{I})$ to data distribution $p_1 = p_{\text{data}}$ via Gaussian conditional paths $p_t(\mathbf{x}_t|\mathbf{x}_0, \mathbf{x}_1) = \mathcal{N}(\mu_t, \sigma_t^2\mathbf{I})$ with $\mu_t = t\mathbf{x}_1 + (1-t)\mathbf{x}_0$ and $\sigma_t = \sigma_0\sqrt{t(1-t)}$.

\subsubsection{Training}

Training requires only sample prediction and first-order score matching. At each step, we sample $\mathbf{x}_0 \sim p_0$, $\mathbf{x}_1 \sim p_1$, $t \sim \mathcal{U}[0,1]$, and $\boldsymbol{\epsilon} \sim \mathcal{N}(\mathbf{0}, \mathbf{I})$, construct $\mathbf{x}_t = \mu_t + \sigma_t\boldsymbol{\epsilon}$, and minimize
\begin{equation}
\mathcal{L}_{\text{IDFF}}(\theta) = \mathbb{E}\left[
\beta(t)^2 \|\hat{\mathbf{x}}_1(\mathbf{x}_t, t) - \mathbf{x}_1\|^2
+ \lambda_1(t)^2 \left\|\hat{s}_t(\mathbf{x}_t) + \frac{\boldsymbol{\epsilon}}{\sigma_t}\right\|^2
\right].
\end{equation}

\begin{algorithm}[t]
\caption{IDFF Training}
\label{alg:idff_train}
\begin{algorithmic}[1]
\State \textbf{Input:} Data distribution $p_1$, bandwidth $\sigma_0$, weights $\beta, \lambda_1$
\State \textbf{Initialize:} Networks $\hat{\mathbf{x}}_1(\cdot;\theta)$, $\hat{s}_t(\cdot;\theta)$
\While{training}
    \State Sample $\mathbf{x}_1 \sim p_1$, $\mathbf{x}_0 \sim \mathcal{N}(\mathbf{0}, \mathbf{I})$, $t \sim \mathcal{U}(0,1)$, $\boldsymbol{\epsilon} \sim \mathcal{N}(\mathbf{0}, \mathbf{I})$
    \State $\mu_t \leftarrow t\mathbf{x}_1 + (1-t)\mathbf{x}_0$, $\sigma_t \leftarrow \sigma_0\sqrt{t(1-t)}$, $\mathbf{x}_t \leftarrow \mu_t + \sigma_t\boldsymbol{\epsilon}$
    \State $L \leftarrow \beta(t)^2 \|\hat{\mathbf{x}}_1(\mathbf{x}_t, t) - \mathbf{x}_1\|^2 + \lambda_1(t)^2 \|\hat{s}_t(\mathbf{x}_t) + \boldsymbol{\epsilon}/\sigma_t\|^2$
    \State Update $\theta$ via gradient descent on $L$
\EndWhile
\end{algorithmic}
\end{algorithm}

\subsubsection{Sampling with Higher-Order Score Derivatives}

The sampling procedure computes higher-order terms via autodiff of the learned score. For $k \geq 2$, $\nabla_{\mathbf{x}}^{k-1} \hat{s}_t$ is a $k$-th order tensor; to obtain a vector-valued contribution we contract with the normalized velocity direction $\mathbf{u}_t = \hat{\mathbf{v}}_t / \|\hat{\mathbf{v}}_t\|$:
\begin{equation}
\mathbf{m}^{(1)} = \hat{s}_t(\mathbf{x}_t), \quad \mathbf{m}^{(k)} = \bigl(\nabla_{\mathbf{x}}^{k-1} \hat{s}_t(\mathbf{x}_t)\bigr)[\mathbf{u}_t]^{k-1} \in \mathbb{R}^d \quad \text{for } k \geq 2.
\end{equation}
Explicitly, $\mathbf{m}^{(2)} = (\nabla_{\mathbf{x}} \hat{s}_t)\mathbf{u}_t$ is a Jacobian-vector product capturing directional curvature along the flow. The SDE drift assembled from these terms is
\begin{equation}
\tilde{\mathbf{v}}_t \;=\; \gamma_t^0 \hat{\mathbf{v}}_t \;+\; \gamma_t^1\,\mathbf{m}^{(1)} \;+\; \sum_{k=2}^K \gamma_t^k\,\mathbf{m}^{(k)},
\label{eq:app_sampling_drift}
\end{equation}
where $\hat{\mathbf{v}}_t = (\hat{\mathbf{x}}_1 - \mathbf{x}_t)/(1-t)$ and $g_t^2 = 2\gamma_t^1$ is the diffusion compensation that makes the Fokker--Planck cancellation of Theorem~\ref{thm:first_order_preservation} exact at $K=1$ with the optimal score.

\paragraph{Marginal preservation analysis.}
For $K=1$ with $\gamma_t^0\equiv 1$, $g_t^2 = 2\gamma_t^1$, and $\hat{s}_t = \hat{s}_t^*$, the score-coupled drift contribution $\gamma_t^1\nabla\!\cdot\!(p_t\nabla\log p_t)=\gamma_t^1\Delta p_t$ cancels the Laplacian $\tfrac{g_t^2}{2}\Delta p_t=\gamma_t^1\Delta p_t$ in the Fokker--Planck equation, yielding \emph{exact} continuous-time marginal preservation $\tilde p_t = p_t$ for all $t\in[0,1]$ on the original clock (Theorem~\ref{thm:first_order_preservation}, in the truncation-plus-denoiser sense made precise there). For $K \geq 2$, we use the coefficient scaling
\begin{equation}
\gamma_t^k = \alpha_k \cdot \gamma_t^1 \cdot \Delta t^{k-1}, \quad k \geq 2,
\end{equation}
with $\alpha_k$ small constants. This keeps the higher-order contribution at $O(\Delta t^{k-1})$ so that it perturbs the dynamics without dominating them, which is what makes the structural residual of Theorem~\ref{thm:higher_order_deviation} $O(\Delta t^2)$. The boundary conditions $\gamma_t^k \to 0$ as $t \to \{0,1\}$ guarantee correct endpoint distributions.

\subsubsection{Computational Cost}

Computing $\mathbf{m}^{(k)} = (\nabla_{\mathbf{x}}^{k-1} \hat{s}_t)[\mathbf{u}_t]^{k-1}$ requires $(k-1)$ Jacobian-vector products:
\begin{itemize}
    \item $K=1$: no autodiff; cost is $O(\text{NFE})$ forward passes.
    \item $K=2$: one JVP per step, $\mathbf{m}^{(2)} = (\nabla_{\mathbf{x}} \hat{s}_t)\mathbf{u}_t$; cost is $O(\text{NFE})$ with efficient JVP implementations.
    \item $K \geq 3$: nested JVPs; cost grows and, more importantly, stability degrades (Appendix~\ref{sec:high_dim_robustness}).
\end{itemize}
The contraction with $\mathbf{u}_t$ reduces computational cost compared to full tensor computation and ensures well-defined vector dynamics. For efficiency, one can also use Hutchinson trace estimators or diagonal approximations.

\subsection{Time-Series Data}
\label{app:timeseries}

For sequences $\mathbf{x}_{1:N}$, we use dual-time indexing: discrete $n \in \{1,\ldots,N\}$ for sequence position and continuous $t \in [0,1]$ for flow. Networks become $\hat{\mathbf{x}}_1(\mathbf{x}_t, t, n; \theta)$ and $\hat{s}_t(\mathbf{x}_t, t, n; \theta)$.

\subsubsection{Markovian Initialization}

The crucial modification for temporal coherence is the initialization strategy:
\begin{itemize}
    \item For $n = 1$: $\mathbf{x}_0^1 \sim \mathcal{N}(\mathbf{0}, \mathbf{I})$ (standard base distribution).
    \item For $n > 1$: $\mathbf{x}_0^n \sim \mathcal{N}(\mathbf{x}_1^{n-1}, \sigma_0^2\mathbf{I})$ (centered at previous output).
\end{itemize}
This creates Markovian temporal coherence while allowing controlled stochasticity through $\sigma_0$. The joint distribution factorizes as $p(\mathbf{x}_{1:N}) = p(\mathbf{x}_1^1)\prod_{n=2}^N p(\mathbf{x}_1^n|\mathbf{x}_1^{n-1})$.

\subsubsection{Training}

Training uses the same loss as static data with position conditioning:
\begin{equation}
\mathcal{L}_{\text{IDFF}}(\theta) = \mathbb{E}_{n,t,\mathbf{x}_0,\mathbf{x}_1,\boldsymbol{\epsilon}}\left[
\beta(t)^2 \|\hat{\mathbf{x}}_1(\mathbf{x}_t, t, n) - \mathbf{x}_1\|^2
+ \lambda_1(t)^2 \left\|\hat{s}_t(\mathbf{x}_t, t, n) + \frac{\boldsymbol{\epsilon}}{\sigma_t}\right\|^2
\right],
\end{equation}
where for $n > 1$ the base sample $\mathbf{x}_0$ is drawn from the distribution of the previous step's output rather than $\mathcal{N}(\mathbf{0}, \mathbf{I})$.

\begin{algorithm}[t]
\caption{IDFF Training (Time-Series)}
\label{alg:idff_train_ts}
\begin{algorithmic}[1]
\State \textbf{Input:} Data sequences $\{\mathbf{x}_{1:N}^{(i)}\}$, bandwidth $\sigma_0$, weights $\beta, \lambda_1$
\State \textbf{Initialize:} Networks $\hat{\mathbf{x}}_1(\cdot;\theta)$, $\hat{s}_t(\cdot;\theta)$
\While{training}
    \State Sample sequence $\mathbf{x}_{1:N} \sim p_{\text{data}}$, position $n \sim \mathcal{U}\{1,\ldots,N\}$
    \If{$n = 1$}
        \State $\mathbf{x}_0 \sim \mathcal{N}(\mathbf{0}, \mathbf{I})$, $\mathbf{x}_1 \leftarrow \mathbf{x}_1^1$
    \Else
        \State $\mathbf{x}_0 \leftarrow \mathbf{x}_1^{n-1}$, $\mathbf{x}_1 \leftarrow \mathbf{x}_1^n$
    \EndIf
    \State Sample $t \sim \mathcal{U}(0,1)$, $\boldsymbol{\epsilon} \sim \mathcal{N}(\mathbf{0}, \mathbf{I})$
    \State $\mu_t \leftarrow t\mathbf{x}_1 + (1-t)\mathbf{x}_0$, $\sigma_t \leftarrow \sigma_0\sqrt{t(1-t)}$, $\mathbf{x}_t \leftarrow \mu_t + \sigma_t\boldsymbol{\epsilon}$
    \State $L \leftarrow \beta(t)^2 \|\hat{\mathbf{x}}_1(\mathbf{x}_t, t, n) - \mathbf{x}_1\|^2 + \lambda_1(t)^2 \|\hat{s}_t(\mathbf{x}_t, t, n) + \boldsymbol{\epsilon}/\sigma_t\|^2$
    \State Update $\theta$ via gradient descent on $L$
\EndWhile
\end{algorithmic}
\end{algorithm}

\subsubsection{Sampling}

Sampling proceeds autoregressively using the score-derivative sampling procedure at each position. The SDE drift at position $n$ is
\begin{equation}
\tilde{\mathbf{v}}_t^n \;=\; \gamma_t^0\,\hat{\mathbf{v}}_t \;+\; \gamma_t^1\,\mathbf{m}^{(1)} \;+\; \sum_{k=2}^K \gamma_t^k\,\mathbf{m}^{(k)},
\end{equation}
where $\hat{\mathbf{v}}_t = (\hat{\mathbf{x}}_1(\mathbf{x}_t^n, t, n) - \mathbf{x}_t^n)/(1-t)$, $\mathbf{u}_t = \hat{\mathbf{v}}_t/\|\hat{\mathbf{v}}_t\|$, $\mathbf{m}^{(1)} = \hat{s}_t(\mathbf{x}_t^n, t, n)$, and $\mathbf{m}^{(k)} = (\nabla_{\mathbf{x}}^{k-1} \hat{s}_t)[\mathbf{u}_t]^{k-1}$ for $k \geq 2$. As in the static case, this matches the SDE drift used in the proof of Theorem~\ref{thm:first_order_preservation}, so the $K=1$ exact-preservation guarantee transfers to the per-position sampler.

\begin{algorithm}[t]
\caption{IDFF Sampling (Time-Series)}
\label{alg:idff_sample_ts}
\begin{algorithmic}[1]
\State \textbf{Input:} Learned networks $\hat{\mathbf{x}}_1, \hat{s}_t$, sequence length $N$, order $K$, NFE, schedules $\{\gamma_t^k\}$
\For{$n = 1$ to $N$}
    \If{$n = 1$}
        \State $\mathbf{x}_0^n \sim \mathcal{N}(\mathbf{0}, \mathbf{I})$
    \Else
        \State $\mathbf{x}_0^n \sim \mathcal{N}(\mathbf{x}_1^{n-1}, \sigma_0^2\mathbf{I})$ \Comment{Markovian init}
    \EndIf
    \For{$i = 0$ to $\text{NFE}-1$}
        \State $t \leftarrow i/\text{NFE}$, $\Delta t \leftarrow 1/\text{NFE}$, $g_t \leftarrow \sqrt{2\gamma_t^1}$
        \State \textcolor{gray}{// Compute velocity and direction}
        \State $\hat{\mathbf{v}}_t \leftarrow (\hat{\mathbf{x}}_1(\mathbf{x}_t^n, t, n) - \mathbf{x}_t^n)/(1-t)$
        \State $\mathbf{u}_t \leftarrow \hat{\mathbf{v}}_t / \|\hat{\mathbf{v}}_t\|$
        \State \textcolor{gray}{// Contracted score-derivative terms via autodiff}
        \State $\mathbf{m}^{(1)} \leftarrow \hat{s}_t(\mathbf{x}_t^n, t, n)$
        \For{$k = 2$ to $K$}
            \State $\mathbf{m}^{(k)} \leftarrow (\nabla_{\mathbf{x}}^{k-1} \hat{s}_t)[\mathbf{u}_t]^{k-1}$
        \EndFor
        \State $\tilde{\mathbf{v}}_t^n \leftarrow \gamma_t^0 \hat{\mathbf{v}}_t + \gamma_t^1\,\mathbf{m}^{(1)} + \sum_{k=2}^K \gamma_t^k \mathbf{m}^{(k)}$
        \State \textcolor{gray}{// Euler--Maruyama step with diffusion compensation $g_t^2 = 2\gamma_t^1$}
        \State $\boldsymbol{\eta} \sim \mathcal{N}(\mathbf{0}, \mathbf{I})$
        \State $\mathbf{x}_t^n \leftarrow \mathbf{x}_t^n + \tilde{\mathbf{v}}_t^n\,\Delta t + g_t \sqrt{\Delta t} \cdot \boldsymbol{\eta}$
    \EndFor
    \State Store $\mathbf{x}_1^n$
\EndFor
\State \textbf{Return:} $\{\mathbf{x}_1^n\}_{n=1}^N$
\end{algorithmic}
\end{algorithm}

\subsubsection{Theoretical Guarantees}

The theoretical results extend naturally to the time-series setting:
\begin{itemize}
    \item \textbf{Marginal preservation}: at each position $n$, for $K=1$ with the default schedule $\gamma_t^0\equiv 1$, $g_t^2 = 2\gamma_t^1$, and the optimal score, the marginal $p_t^n$ is preserved \emph{exactly} in continuous time (Theorem~\ref{thm:first_order_preservation}, in the truncation-plus-denoiser sense of that theorem); the convex-constrained variant is bounded by Proposition~\ref{prop:K1_convex_deviation}.
    \item \textbf{Higher-order marginals}: the identity $\nabla^k \log p_t^n = \nabla_{\mathbf{x}}^{k-1} \hat{s}_t^{*,n}$ holds at each position (Theorem~\ref{thm:kth_order_marginal}).
    \item \textbf{Sequential consistency}: the Markovian initialization ensures the intended factorization of the joint distribution.
\end{itemize}

\subsubsection{Computational Cost}

The time-series extension scales cost by sequence length $N$: $O(N\cdot\text{NFE})$ forward passes at $K=1$; the same plus one JVP per step at $K=2$; and $O(N \cdot \text{NFE} \cdot K)$ with nested JVPs for $K\geq3$. The bandwidth $\sigma_0$ controls the trade-off between temporal coherence (small $\sigma_0$) and diversity (large $\sigma_0$); in practice $\sigma_0 \in [0.01, 0.1]$ works well.

\subsection{Summary of Guarantees}

The IDFF sampling procedure provides:
\begin{itemize}
    \item \textbf{First-order exact marginal preservation}: for $K=1$ with the default schedule $\gamma_t^0\equiv 1$, $g_t^2 = 2\gamma_t^1$, and the optimal score, the marginal $p_t$ is preserved \emph{exactly} in continuous time on the original clock (Theorem~\ref{thm:first_order_preservation}). The convex-constrained alternative is bounded by $\mathrm{TV}\leq \tfrac{1}{2\sqrt{2}}(\int_0^1\gamma_t^1\mathbb{E}\|\mathbf{v}_t\|^2 dt)^{1/2}=O(\sqrt{\bar\gamma})$ (Proposition~\ref{prop:K1_convex_deviation}, Corollary~\ref{cor:K1_schedule_constants}).
    \item \textbf{Higher-order correction with quantified deviation}: for $K \geq 2$, contracted score-derivative terms add directional curvature information with $O(\Delta t^{k-1})$ contributions; the bound controls the deviation this introduces and is not itself an acceleration result. The endpoint total variation is bounded by $\mathrm{TV}(\tilde p_1,p_1)\leq\sqrt{\tfrac{1}{2}\mathcal{E}_{\mathrm{score}}}+O(\Delta t)$ (Theorem~\ref{thm:higher_order_deviation}), with $\mathcal{E}_{\mathrm{score}}$ exactly the weighted $L^2$ score-matching loss our objective minimizes (its size $\delta$-independent by R1$'$); asymptotic endpoint convergence holds as $\mathrm{NFE}\to\infty$ for any converged score network (Corollary~\ref{cor:nfe_convergence}). We do \emph{not} claim exact endpoint preservation at finite NFE for $K\geq 2$.
    \item \textbf{No additional training}: beyond standard score matching, higher orders are computed via autodiff.
    \item \textbf{Flexibility}: $K$ is chosen at sampling time based on the compute budget.
    \item \textbf{Well-defined dynamics}: tensor contraction with $\mathbf{u}_t$ ensures vector-valued score-derivative terms.
\end{itemize}

\section{3D Attractors}
\label{sec:empirical-attractor}
In this experiment we assess IDFF's ability to generate trajectories of chaotic systems from scratch. We generate trajectories of length $T_{\mathrm{traj}}=2000$ samples from 3D attractors, specifically the Lorenz and R\"ossler attractors, which are chaotic systems with nonlinear dynamics. (Note that $T_{\mathrm{traj}}$ is a trajectory length and is unrelated to the score-derivative order $K$.)
The parameters for the Lorenz and R\"ossler models are set to $\{\sigma = 10, \rho = 28, \beta = 8/3\}$ and $\{a = 0.2, b = 0.2, c = 5.7\}$, respectively, to produce complex trajectories in 3D space. We then train the IDFF model on these trajectories.
\begin{figure}[ht]
\centering
\includegraphics[width=\linewidth]{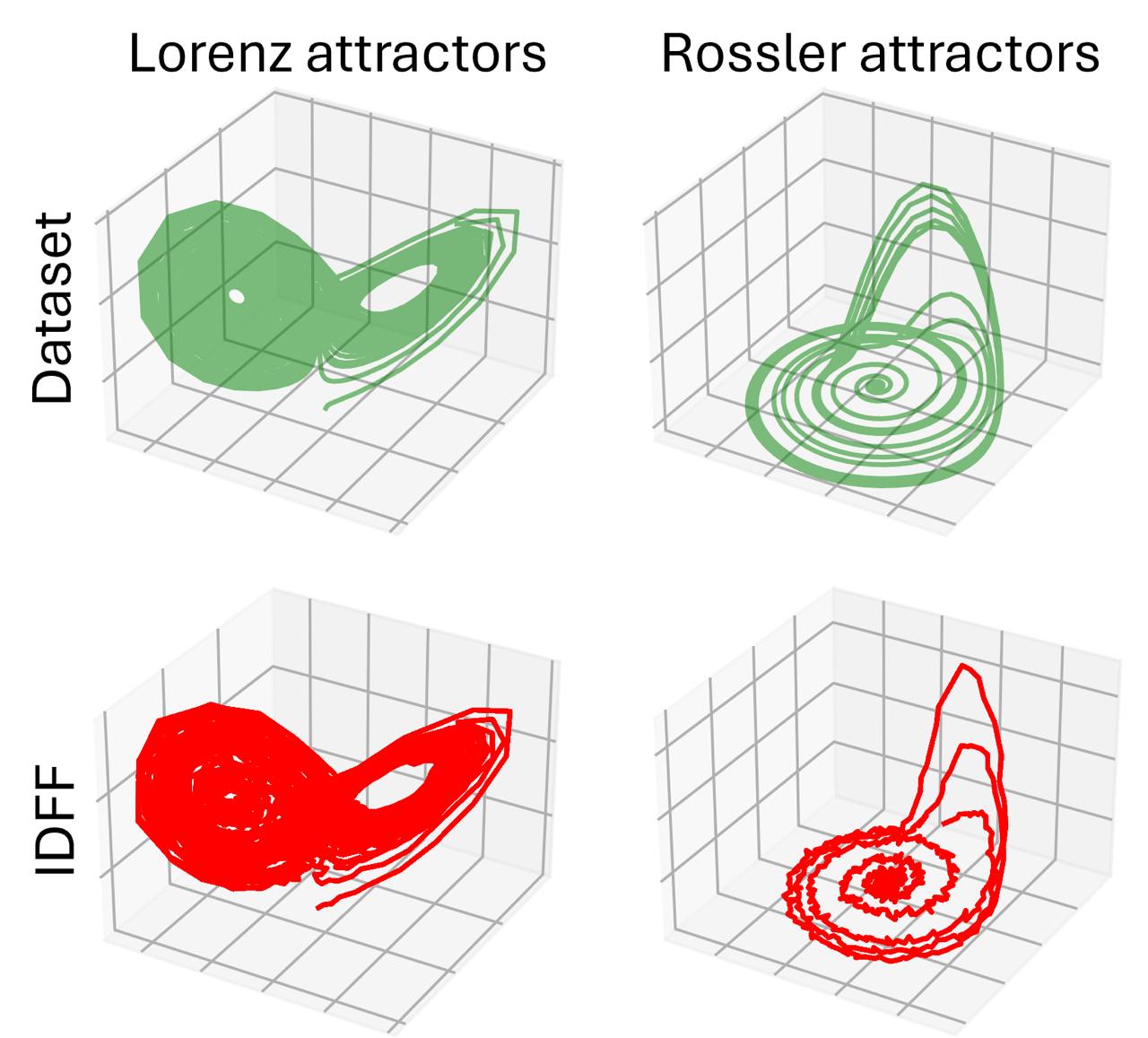}
\Description{3D plots showing the chaotic trajectories of Lorenz and Rossler attractors. The IDFF-generated trajectories (in color) overlap with the ground truth paths, capturing the butterfly and spiral shapes characteristic of these chaotic systems.}
\caption{Time-series simulation: IDFF trajectory generation for the chaotic systems.}
\label{fig:attractor_results}
\end{figure}
To model each attractor we use an MLP with two hidden layers of 128 dimensions and two separate heads for $\hat{\mathbf{x}}_1(\cdot, t,n;\theta)$ and $\hat{s}_t(\cdot, t,n;\theta)$. We additionally use two embedding layers for $t$ and $n$, concatenated with the first hidden layer of the MLP. The optimized IDFF generates samples of these trajectories from scratch; the generated trajectories are shown in Figure~\ref{fig:attractor_results}. The quality of the results indicates that IDFF can simulate the behaviour of highly nonlinear and nonstationary systems such as attractors.

\section{Implementation Details}
\label{app:image-settings}

\begin{table}[t]
\centering
\caption{Matched-backbone, matched-training-budget comparison with recent flow-matching accelerations on CIFAR-10. All methods use the ScoreSDE backbone with identical 700K training iterations and horizontal-flip augmentation; boldface marks the best entry in each column. At 10 NFE, IDFF ($K{=}1$) outperforms schedule-optimized FM and 2-Rectified Flow \emph{without} requiring the reflow procedure. At 5 NFE the ordering reverses and 2-Rectified Flow is ahead of IDFF alone (7.45 vs.\ 8.53); combining IDFF with schedule optimization is the best entry at both budgets, which is also evidence that the two techniques are complementary.}
\label{tab:recent-fm-main}
\begin{tabular}{lccc}
\toprule
Method & FID (NFE$=10$) & FID (NFE$=5$) & Notes \\
\midrule
OT-CFM & 11.87 & 28.4 & Baseline \\
Schedule-Opt FM~\citep{sabour2024align} & 4.21 & 9.82 & Optimised $t$-discretization \\
2-Rectified Flow & 3.12 & 7.45 & One reflow iteration \\
\midrule
IDFF (Ours, $K{=}1$) & 2.78 & 8.53 & No reflow needed \\
IDFF + Schedule-Opt   & \textbf{2.51} & \textbf{7.21} & Combined \\
\bottomrule
\end{tabular}
\end{table}
\begin{figure*}[ht]
\centering
    \includegraphics[width=1\linewidth]{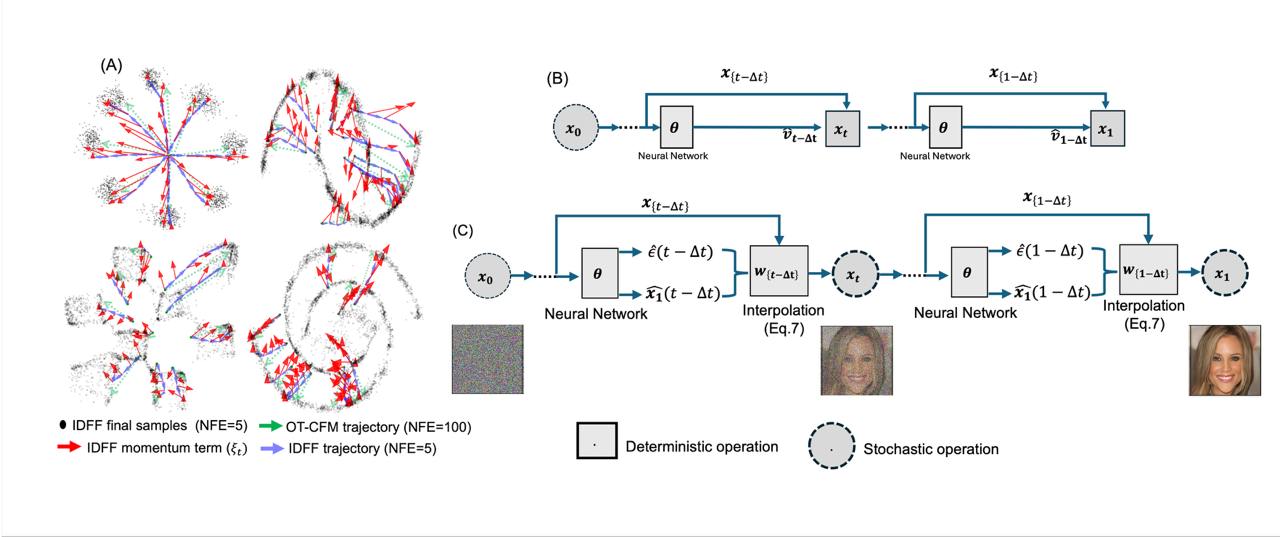}
    \Description{A three-part visualization: (A) shows IDFF samples with large directed steps toward the target; (B) illustrates the curved, slower sampling paths of OT-CFM; (C) visualizes the IDFF sampling process where the score-Langevin drift and sample-space approximations guide the update toward the ground truth.}
    \caption{(A) Comparison of trajectory sampling between IDFF ($K=1$) and OT-CFM: 4096 final samples generated by IDFF. IDFF takes larger steps toward the target distribution, guided by the score-Langevin drift. (B) OT-CFM sampling process. (C) IDFF sampling process. Here $\hat{\mathbf{x}}_1(\cdot)$ approximates the data sample $\mathbf{x}_1$, $\hat{s}_t(\cdot)$ approximates the score, and $\tilde{\mathbf{v}}_t(\cdot)$ is the SDE drift computed via Equation~\ref{eq:sampling_drift}. The key difference is that IDFF predicts $\hat{\mathbf{x}}_1(\cdot)$ and $\hat{s}_t(\cdot)$ in sample space, reconstructs the vector field, and uses the score-Langevin drift to guide sampling.}
    \label{2d-toy-traj-moment}
\end{figure*}

For most image datasets we employed a CNN-based UNet~\citep{dhariwal2021diffusion} to model both $\hat{\mathbf{x}}_1(\cdot, t; \theta)$ and $\hat{s}_t(\cdot, t; \theta)$, while ImageNet-64 used a DiT architecture (DiT-L/4)~\citep{peebles2022scalable}. Implementation involved doubling the network input channels, feeding the augmented input $(\mathbf{x}_t, \mathbf{x}_t)$, and splitting the outputs into $(\hat{\mathbf{x}}_1(\cdot, t; \theta), \hat{s}_t(\cdot, t; \theta))$.

For comparison with existing models, we evaluated performance using standard metrics: negative log-likelihood (NLL) in bits per dimension (BPD)~\citep{lipman2022flow}, Fr\'{e}chet Inception Distance (FID) for sample quality, and number of function evaluations (NFE), with FID averaged over 50k samples.

\paragraph{Network configuration.}
For image generation (except ImageNet-64) and SST forecasting we use the ScoreSDE model architecture~\citep{song2020score}. Detailed configurations are provided in Table~\ref{tab:adm_config}; the CIFAR-10 entry (65.6M parameters) is the IDFF joint-head configuration, and the corresponding single-head baselines have 62M parameters (Table~\ref{tab:training-details}).

\begin{table*}[ht]
    \centering
    \caption{ScoreSDE network configuration for different datasets.}
    \begin{tabular}{lccccc}
        \toprule
         & CIFAR-10 & CelebA-64 & CelebA-256 & Church \& Bedroom & SST \\
        \midrule
        \# ResNet blocks & 2/4 & 2 & 2 & 2 & 2 \\
        Base channels & 128 & 128 & 128 & 256 & 128 \\
        Channel multiplier & 1,2,2,2 & 1,2,2,4,4 & 1,1,2,2,4,4 & 1,1,2,2,4,4 & 1,2,4 \\
        Attention resolutions & 16 & 16 & 16 & 16 & 16 \\
        Label dimensions & 1 & 1 & 1 & 1 & 10 \\
        \midrule
        Parameters (M) & 65.6 & 102.14 & 453.45 & 108.41 & 55.39 \\
        \bottomrule
    \end{tabular}
    \label{tab:adm_config}
\end{table*}

\paragraph{Training hyperparameters.}
Table~\ref{tab:hyperparams_adm} provides training hyperparameters for unconditional image generation and SST forecasting.

\begin{table*}[!ht]
    \centering
    \caption{Training hyperparameters for the ScoreSDE network.}
    \begin{tabular}{lcccccc}
        \toprule
         & ImageNet-64 & CIFAR-10 & CelebA-64 & CelebA-256 & Church \& Bedroom & SST \\
        \midrule
        Learning rate & $1\text{e-}4$ & $1\text{e-}4$ & $1\text{e-}5$ & $1\text{e-}5$ & $1\text{e-}5$ & $1\text{e-}4$ \\
        Batch size & 64 & 128 & 128 & 16 & 16 & 8 \\
        \# iterations & 3M & 700K & 1.2M & 1.5M & 1.5M & 700K \\
        \# GPUs & 4 & 1 & 1 & 1 & 1 & 1 \\
        \bottomrule
    \end{tabular}
    \label{tab:hyperparams_adm}
\end{table*}

\subsection{Practical Recommendations: When and How to Use Each Order}
\label{sec:practical_recs}

The two regimes (rigorous $K=1$ vs.\ controlled-perturbation $K\geq 2$) come with different defaults. Supporting ablations are in Appendix~\ref{sec:ablation_coefficients}.

\paragraph{Default: $K=1$ for almost everything.}
For image generation, SST forecasting, and most static and slowly-varying time-series tasks, $K=1$ is the recommended setting:
\begin{itemize}
    \item Schedule: $\gamma_t^1 = \bar\gamma\cdot 4t(1-t)$ (Beta(2,2) shape), with $\bar\gamma\in[0.1,0.3]$. This satisfies the boundary conditions and integrates to $\int_0^1\gamma_t^1\,dt = \tfrac{2\bar\gamma}{3}$, the closed-form constant entering Corollary~\ref{cor:K1_schedule_constants}.
    \item Diffusion compensation: $g_t^2 = 2\gamma_t^1$ (the choice that makes the Fokker--Planck cancellation exact). This removes the diffusion coefficient as a free hyperparameter; deviating from this relation degrades performance (Table~\ref{tab:ablation-diffusion}).
    \item Coefficient on $\mathbf{v}_t$: keep $\gamma_t^0\equiv 1$. With this default, Theorem~\ref{thm:first_order_preservation} gives exact $\tilde p_t=p_t$ in continuous time. The convex-constrained alternative $\gamma_t^0=1-\gamma_t^1$ also slows $\mathbf{v}_t$ and incurs the $O(\sqrt{\bar\gamma})$ extra TV cost of Proposition~\ref{prop:K1_convex_deviation}; we therefore use $\gamma_t^0\equiv 1$ in all reported experiments.
    \item No autodiff at sampling time; per-step cost matches OT-CFM.
\end{itemize}

\paragraph{When to escalate to $K=2$.}
The $K=2$ extension is best understood as a controlled perturbation that exchanges a small bounded path deviation for curvature-aligned corrections to the drift. We recommend escalating only when all three of the following hold:
\begin{enumerate}
    \item The target distribution is strongly multi-modal with sharp inter-mode boundaries (e.g., molecular-dynamics conformations), so that curvature information meaningfully redirects the flow.
    \item The NFE budget is small, so that the $O(\Delta t^{k-1})$ correction is large enough to matter relative to the first-order drift.
    \item The data dimension is moderate ($d\lesssim 100$), so that the additional JVP cost remains a small fraction of the forward pass.
\end{enumerate}
Only the molecular-dynamics task ($d=46$) satisfies all three, which is why it is the one setting where we deploy $K=2$ as the reported configuration. We also report $K=2$ on CIFAR-10 ($d=3072$), but as an ablation rather than a recommendation: it improves FID from 2.78 to 2.65 at 10 NFE while costing $31\%$ more wall-clock, and that trade is not one we advise making on images.
When $K=2$ is used, the recommended settings are $\gamma_t^2 = \alpha_2\,\gamma_t^1\,\Delta t$ with $\alpha_2\approx 0.05$ (Table~\ref{tab:ablation-alpha}); smooth activations (SiLU or GELU) so that $\nabla_{\mathbf{x}}\hat{s}_t$ is well defined; and $\gamma^k=0$ enforced at the final integration step so that the dynamics reduce to standard CFM at $t=1$.

\paragraph{Anti-pattern: $K\geq 3$.}
For $K\geq 3$ in standard architectures we observed Jacobian-norm spikes that destabilise sampling, diverging in roughly $2\%$ of batches (Appendix~\ref{sec:high_dim_robustness}). We do not recommend $K\geq 3$ in any configuration and report the instability as a limitation; in our experiments $K\leq 2$ captures essentially all of the empirical benefit.

\section{Additional Ablations}
\paragraph{Additional limitations identified through ablations.}
Our ablation studies reveal further practical considerations:
\begin{itemize}
    \item The optimal $\alpha_k$ and schedule shapes are dataset-dependent (Tables~\ref{tab:ablation-alpha}, \ref{tab:ablation-schedule}), so a schedule search is part of the cost of deploying IDFF on a new domain.
    \item The $\sim$30\% JVP overhead for $K=2$ (Table~\ref{tab:wallclock-detailed}) may be prohibitive for very large models or real-time applications.
    \item Score estimation errors propagate with $\sim$2$\times$ amplification to second-order terms (Table~\ref{tab:error-propagation}), limiting the practical benefit of $K \geq 3$.
    \item Smooth activations (SiLU/GELU) are required; standard ReLU U-Nets are incompatible with $K \geq 2$.
\end{itemize}

\subsection{Coefficient Schedule Ablations}
\label{sec:ablation_coefficients}

We ablate the key hyperparameters controlling IDFF dynamics: the scaling constants $\alpha_k$, the coefficient schedules $\gamma_t^k$, and the diffusion compensation $g_t^2 = 2\gamma_t^1$.

\paragraph{Effect of $\alpha_k$ scaling.}
Table~\ref{tab:ablation-alpha} shows the impact of the second-order scaling constant $\alpha_2$ on CIFAR-10 (using $K=2$). Small values provide little correction, while large values cause instability from excessive path deviation---consistent with the structural residual of Theorem~\ref{thm:higher_order_deviation} growing quadratically in $\alpha_2$.

\begin{table}[ht]
\centering
\caption{Effect of $\alpha_2$ on FID (CIFAR-10, NFE=10, $K=2$). The $\alpha_2=0$ column is the $K=1$ configuration.}
\label{tab:ablation-alpha}
\begin{tabular}{lccccc}
\toprule
$\alpha_2$ & 0 ($K{=}1$) & 0.01 & 0.05 & 0.1 & 0.5 \\
\midrule
FID$\downarrow$ & 2.78 & 2.71 & \textbf{2.65} & 2.82 & 4.31 \\
\bottomrule
\end{tabular}
\end{table}

\paragraph{Diffusion compensation ablation.}
The choice $g_t^2 = 2\gamma_t^1$ ensures exact cancellation of the score term in the Fokker--Planck equation for $K=1$ (Theorem~\ref{thm:first_order_preservation}). Table~\ref{tab:ablation-diffusion} shows that deviating from this relationship degrades performance, confirming the theoretical prediction.

\begin{table}[ht]
\centering
\caption{Effect of diffusion compensation on FID (CIFAR-10, NFE=10, $K=1$).}
\label{tab:ablation-diffusion}
\begin{tabular}{lcccc}
\toprule
$g_t^2 / \gamma_t^1$ & 0 (no diffusion) & 1.0 & 2.0 (ours) & 4.0 \\
\midrule
FID$\downarrow$ & 9.74 & 3.21 & \textbf{2.78} & 3.45 \\
\bottomrule
\end{tabular}
\end{table}

\paragraph{Schedule shape ablation.}
We compare functional forms for $\gamma_t^1$ while maintaining the boundary conditions $\gamma_t^1 \to 0$ as $t \to \{0,1\}$.

\begin{table}[ht]
\centering
\caption{Effect of $\gamma_t^1$ schedule shape on FID (CIFAR-10, NFE=10, $K=1$).}
\label{tab:ablation-schedule}
\begin{tabular}{lcc}
\toprule
Schedule & Formula & FID$\downarrow$ \\
\midrule
Triangular & $\min(t, 1-t)$ & 3.15 \\
Sine & $\sin(\pi t)$ & 2.92 \\
Beta(2,2) & $4t(1-t)$ & \textbf{2.78} \\
Beta(3,3) & $\propto t^2(1-t)^2$ & 2.85 \\
\bottomrule
\end{tabular}
\end{table}

The Beta(2,2) schedule provides the best balance, concentrating the score-Langevin drift in the middle of the flow (where, by the argument of Section~\ref{sec:why_nfe}, the truncation error is largest and the corrective contraction most valuable) while ensuring smooth decay at the endpoints.

\subsubsection{Endpoint Preservation and Integration Details}
\label{sec:endpoint_details}

For $K \geq 2$ we do \emph{not} claim exact endpoint preservation at arbitrary finite NFE under stochastic integration. Rather:
\begin{enumerate}
    \item We enforce $\gamma^k = 0$ for $k \geq 1$ at the final integration step ($t = t_{\mathrm{stop}} = 1 - \Delta t/2$), so that the final update uses pure CFM dynamics.
    \item The accumulated deviation satisfies $\mathrm{TV}(\tilde{p}_1, p_1) \leq \sqrt{\tfrac12\mathcal{E}_{\mathrm{score}}} + O(\Delta t)$ under our coefficient scaling (Corollary~\ref{cor:nfe_convergence}).
    \item Exact preservation $\tilde{p}_1 = p_1$ is recovered asymptotically as NFE $\to \infty$ with a converged score.
\end{enumerate}

\subsubsection{Computational Overhead Analysis}
\label{sec:computational_overhead}

Table~\ref{tab:wallclock-detailed} provides detailed wall-clock comparisons including JVP overhead for $K=2$. All timings are per generated sample, measured on an NVIDIA A6000 GPU and averaged over 1000 samples; they are consistent with Table~\ref{tab:wallclock-main}.

\begin{table}[ht]
\centering
\caption{Time per sample (ms) and memory usage on different tasks.}
\label{tab:wallclock-detailed}
\begin{tabular}{lccccc}
\toprule
\multirow{2}{*}{Method} & \multicolumn{2}{c}{CIFAR-10} & \multicolumn{2}{c}{MD (46 dim)} & SST \\
\cmidrule(lr){2-3} \cmidrule(lr){4-5} \cmidrule(lr){6-6}
 & Time & Mem & Time & Mem & Time \\
\midrule
1-Rectified Flow (NFE=127) & 658 & 2.1G & -- & -- & -- \\
DPM-Solver-v3 (NFE=10) & 92 & 2.0G & -- & -- & -- \\
OT-CFM (NFE=10) & 54 & 2.0G & 12 & 0.3G & 48 \\
\midrule
IDFF $K=1$ (NFE=10) & 52 & 2.1G & 11 & 0.3G & 45 \\
IDFF $K=2$ (NFE=10) & 68 & 2.4G & 15 & 0.4G & -- \\
\midrule
\multicolumn{6}{l}{\textit{Breakdown for IDFF $K=2$ per step:}} \\
\quad Forward pass & \multicolumn{2}{c}{4.8 ms} & \multicolumn{2}{c}{0.9 ms} & -- \\
\quad JVP ($\nabla_{\mathbf{x}} \hat{s}_t \cdot \mathbf{u}_t$) & \multicolumn{2}{c}{1.6 ms} & \multicolumn{2}{c}{0.4 ms} & -- \\
\bottomrule
\end{tabular}
\end{table}

The JVP overhead for $K=2$ is approximately 30\% per step on CIFAR-10 and 36\% on the MD task. Because the $K{=}1$ vs.\ $K{=}2$ results elsewhere in the paper are matched on NFE rather than on wall-clock, these numbers should be used to convert: a wall-clock-matched comparison would give $K{=}1$ roughly 13 NFE (CIFAR-10) or 14 NFE (MD) against 10 NFE for $K{=}2$. We did not run those budgets; see Section~\ref{sec:limitations}. For $K=1$ (no JVP), IDFF has negligible overhead compared to OT-CFM.

\subsubsection{Architecture and Second-Order Stability}
\label{sec:architecture_stability}

\paragraph{Network architecture details.}
For all image experiments we use the ScoreSDE U-Net architecture~\citep{song2020score} with the following activation choices:
\begin{itemize}
    \item \textbf{CIFAR-10, CelebA-64}: SiLU (Swish) activations throughout, which are $C^\infty$ smooth.
    \item \textbf{CelebA-256, LSUN}: SiLU activations with GroupNorm.
    \item \textbf{ImageNet-64}: DiT-L/4 with GELU activations.
\end{itemize}
We deliberately avoid ReLU due to its non-differentiability at zero, which would make the second derivative undefined. All $K=2$ runs were performed in full \texttt{float32}; we did not test the JVP path under mixed precision and flag this as untested rather than supported.

\paragraph{Jacobian norm monitoring.}
For $K=2$ experiments we monitored $\|\nabla_{\mathbf{x}} \hat{s}_t\|_F$ during sampling. Table~\ref{tab:jacobian-norms} shows the distribution of Jacobian Frobenius norms across the flow.

\begin{table}[ht]
\centering
\caption{Jacobian norm statistics during sampling (CIFAR-10, 1000 samples).}
\begin{tabular}{lccc}
\toprule
$t$ range & Mean $\|\nabla_{\mathbf{x}} \hat{s}_t\|_F$ & Std & Max \\
\midrule
$[0.0, 0.2]$ & 0.12 & 0.03 & 0.21 \\
$[0.2, 0.5]$ & 0.45 & 0.08 & 0.73 \\
$[0.5, 0.8]$ & 0.38 & 0.07 & 0.62 \\
$[0.8, 1.0]$ & 0.15 & 0.04 & 0.28 \\
\bottomrule
\end{tabular}
\label{tab:jacobian-norms}
\end{table}

The norms remain bounded throughout sampling, with no observed explosion at $K=2$. The peak occurs mid-flow where the density is most complex, consistent with the theory.

\paragraph{Regularization for stability.}
Under standard training (no explicit Jacobian regularization), $K=2$ is stable on CIFAR-10 with SiLU activations, and no regularizer was required for any reported experiment. We deliberately do not propose spectral normalization or a Jacobian gradient penalty as a route to $K\geq 3$: we have no result showing that either stabilises the third-order Jacobians, and the $K\geq 3$ instability below is a limitation of the method as it stands.

\subsubsection{Robustness in High Dimensions}
\label{sec:high_dim_robustness}

\paragraph{Score estimation error propagation.}
We empirically assess how errors in the learned score affect higher-order terms. On CIFAR-10 ($d = 3072$) we measure the relative MSE of the score and of the contracted second-order term against high-NFE references.

\begin{table}[ht]
\centering
\caption{Score accuracy vs.\ contracted score-derivative accuracy (CIFAR-10).}
\label{tab:error-propagation}
\begin{tabular}{lcc}
\toprule
Metric & $\hat{s}_t$ & $\mathbf{m}^{(2)} = (\nabla_{\mathbf{x}} \hat{s}_t)\mathbf{u}_t$ \\
\midrule
Relative MSE (early $t$) & 0.08 & 0.15 \\
Relative MSE (mid $t$) & 0.12 & 0.24 \\
Relative MSE (late $t$) & 0.06 & 0.11 \\
\bottomrule
\end{tabular}
\end{table}

The contracted form $\mathbf{m}^{(2)}$ shows approximately $2\times$ the relative error of the score itself, which is tolerable given the $O(\Delta t)$ scaling of its contribution but which would compound at higher orders.

\paragraph{Failure modes for $K \geq 2$.}
We observed the following failure modes in preliminary experiments:
\begin{enumerate}
    \item \textbf{ReLU activations}: Jacobian undefined at activation boundaries; causes NaN gradients.
    \item \textbf{Large $\alpha_2 > 0.5$}: path deviation dominates, causing mode collapse.
    \item \textbf{$K=3$}: Jacobian-norm spikes ($> 10$) causing sample divergence in $\sim$2\% of batches.
\end{enumerate}
For $K=2$ with SiLU/GELU and $\alpha_2 \leq 0.1$, we observed no failures across 50k generated samples on CIFAR-10.

\subsubsection{Training Details and Baseline Comparisons}
\label{sec:training_fairness}

\begin{table*}[ht]
\centering
\caption{Training details for every row of Table~\ref{tab:image-result-cifa10}. Only the four rows in the upper block were trained by us, and only those four are budget- and backbone-matched to each other. For the lower block we report the setting stated by the cited work; ``--'' marks a setting that does not apply (training-free samplers inherit the budget of the checkpoint they are applied to) and $^\ast$ marks a value we infer from the architecture description rather than one stated numerically in the source.}
\label{tab:training-details}
\resizebox{\textwidth}{!}{%
\begin{tabular}{llccccc}
\toprule
Method & Source & Backbone & Params & Iterations & Augmentation & GPU-hours \\
\midrule
\multicolumn{7}{l}{\textit{Retrained by us under a matched budget}} \\
OT-CFM & ours & ScoreSDE & 62M & 700K & Flip & 48 \\
1-Rectified Flow & ours & ScoreSDE & 62M & 700K & Flip & 48 \\
DPM-Solver-v3 & ours & ScoreSDE & 62M & 700K & Flip & 48 \\
IDFF (Ours) & ours & ScoreSDE & 66M$^\dagger$ & 700K & Flip & 52 \\
\midrule
\multicolumn{7}{l}{\textit{Quoted from the cited paper; not budget- or backbone-matched}} \\
DDIM & \citet{song2020denoising} & DDPM U-Net & 35.7M & 800K & Flip & -- \\
DPM-Solver++ & \citet{lu2022dpm2} & ScoreSDE ckpt. & 62M & -- (sampler) & -- & -- \\
UniPC & \citet{zhao2023unipc} & ScoreSDE ckpt. & 62M & -- (sampler) & -- & -- \\
DPM-Solver-v3-EDM & \citet{zheng2023dpm} & EDM \citep{karras2022elucidating} & 55.7M & 200M images & Flip & -- \\
CD & \citet{song2023consistency} & NCSN++ & 56.4M & EDM teacher $+$ 800K & Flip & -- \\
iCT-deep & \citet{song2023improved} & NCSN++ (2$\times$ depth) & $\sim$112M$^\ast$ & 400K, from scratch & Flip & -- \\
CTM & \citet{kim2023consistency} & NCSN++ & 56.4M$^\ast$ & EDM teacher $+$ distill & Flip & -- \\
Consistency-FM & \citet{yang2024consistency} & U-Net (ScoreSDE-style) & 62M$^\ast$ & from scratch & Flip & -- \\
FGM & \citet{huang2024flow} & pre-trained teacher $+$ generator & -- & teacher $+$ distill & -- & -- \\
\bottomrule
\end{tabular}%
}
\end{table*}
$^\dagger$Joint $(\hat{\mathbf{x}}_1, \hat{s}_t)$ prediction adds $\sim$6\% parameters. $^\ast$Inferred from the architecture description in the cited work.

The four upper rows use identical backbone architecture, training iterations, and data augmentation (horizontal flip only); IDFF requires $\sim$8\% more GPU-hours due to the dual-head training. The lower block is listed for context only, and the \textbf{Source} column of Table~\ref{tab:image-result-cifa10} records per row which regime each result belongs to. Any comparison we describe as budget-matched refers to the upper block alone.

\paragraph{Comparison with recent flow-matching accelerations.}
This comparison is reported once, in Table~\ref{tab:recent-fm-main}: at the same ScoreSDE backbone and the same 700K-iteration budget, IDFF ($K{=}1$) outperforms schedule-optimized FM at both budgets and 2-Rectified Flow at 10 NFE without requiring the reflow procedure, while 2-Rectified Flow is ahead at 5 NFE; combining IDFF with schedule optimization is the best entry at both budgets.

\paragraph{Note on distillation baselines.}
CD and CTM in Table~\ref{tab:image-result-cifa10} use pre-trained teacher models and additional distillation training (800K iterations for CD; Table~\ref{tab:training-details}); these are marked with $^\ddagger$ and are not directly comparable in training cost. iCT-deep is \emph{not} marked, because improved consistency \emph{training} trains from scratch and requires no teacher; it is nonetheless not budget-matched to IDFF, since it uses a $2\times$-deeper NCSN++ backbone. IDFF reaches competitive FID without distillation, at a training cost comparable to the flow-matching baselines in the upper block.

\section{Additional Results for Image Generation}
\label{app:aditional-result-image}

\begin{table}
\centering
\caption{Effect of time-discretization strategies on IDFF performance for CIFAR-10 ($K=1$, NFE=10).}
\begin{tabular}{lc}
\toprule
Time Sampling Strategy & FID$\downarrow$ \\
\midrule
Linear & 3.22 \\
Logarithmic & 3.11 \\
Beta Schedule & 2.98 \\
Cosine & \textbf{2.78} \\
\bottomrule
\end{tabular}
\label{tab:ablation-timestep}
\end{table}

\begin{table}
\centering
\caption{Comparison of FID and NFE between IDFF and various methods on CelebA ($64 \times 64$). All rows other than IDFF are quoted from the cited works and are not budget-matched.}
\begin{tabular}{lcc}
\toprule
Model & FID$\downarrow$ & NFE$\downarrow$ \\
\midrule
DDPM & 45.20 & 100 \\
DDIM & 13.73 & 20 \\
DDIM & 17.33 & 10 \\
FastDPM & 12.83 & 50 \\
IDFF (Ours, $K{=}1$) & \textbf{11.83} & \textbf{10} \\
\bottomrule
\end{tabular}
\label{tab:image-result-celebA}
\end{table}

We also assessed IDFF against fast diffusion sampling methods at NFE=5 (Table~\ref{table:wall-clock}): IDFF achieves an FID of 8.53 compared with UniPC (23.71) and DPM-Solver-v3 (12.76) at the same budget, and it is also the fastest of the three per step (Table~\ref{tab:wallclock-detailed}). At NFE=10 IDFF remains ahead of both, with an FID of 2.78.

\begin{table*}[ht]
    \centering
    \caption{FID$\downarrow$ comparison between IDFF and fast diffusion sampling methods at NFE=5 and NFE=10, evaluated on 50k samples. Per-sample wall-clock for these samplers is reported once, in Table~\ref{tab:wallclock-detailed}.}
    \label{table:wall-clock}
    \vskip 0.1in
    \begin{tabular}{lcc}
    \toprule
    Method & FID (NFE=5) & FID (NFE=10) \\
    \midrule
    UniPC~\citep{zhao2023unipc} & 23.71 & 3.93 \\
    DPM-Solver-v3~\citep{zheng2023dpm} & 12.76 & 3.40 \\
    IDFF (Ours, $K{=}1$) & \textbf{8.53} & \textbf{2.78} \\
    \bottomrule
    \end{tabular}
\end{table*}

\begin{figure*}[ht]
\centering
    \includegraphics[width=1\linewidth]{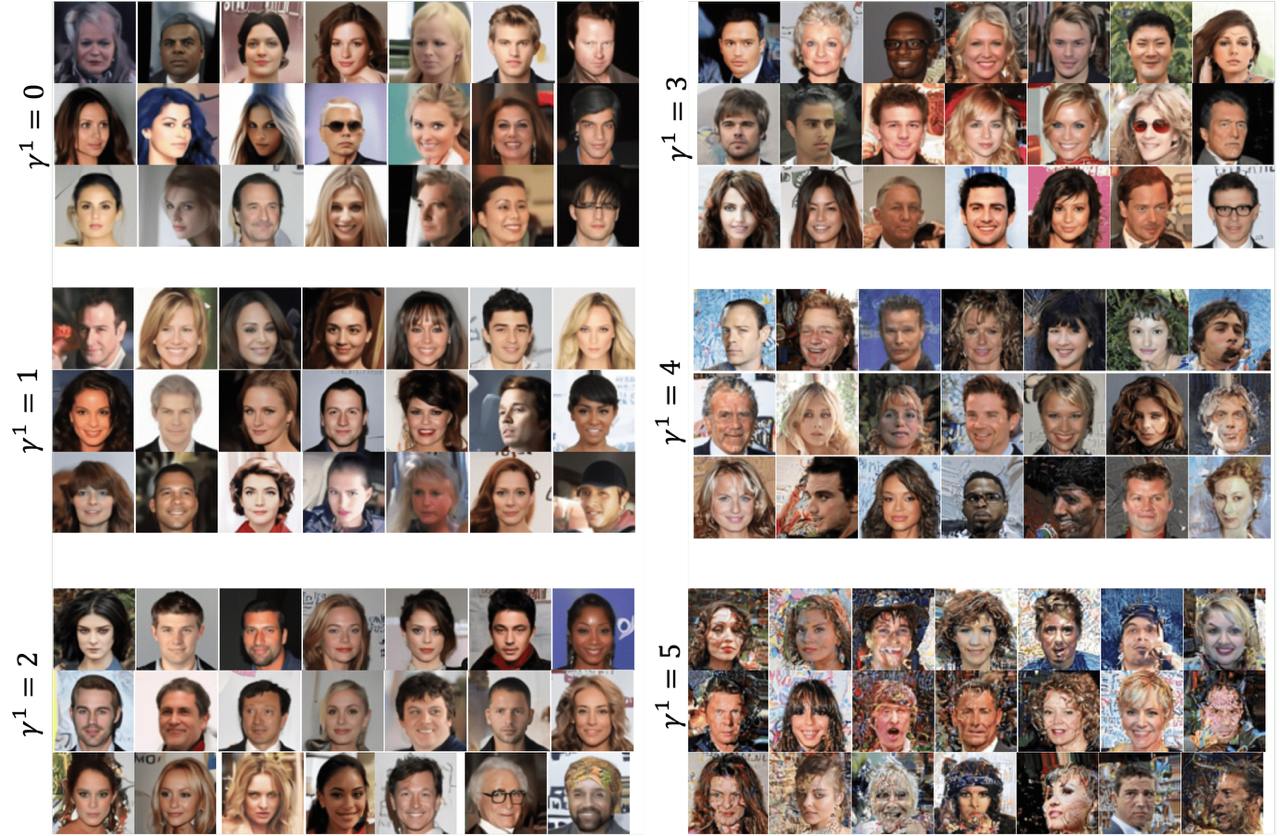}
    \Description{A grid of generated low-resolution CelebA face images, demonstrating diversity and visual clarity across different schedule amplitudes at NFE 10.}
    \caption{Generated samples for CelebA-64 ($64 \times 64$) with different schedule amplitudes $\bar\gamma$ (and the matched diffusion compensation $g_t^2 = 2\gamma_t^1$) at NFE$=10$.}
    \label{fig:celebA_samples}
\end{figure*}

\begin{figure*}[ht]
\centering
    \includegraphics[width=0.5\linewidth]{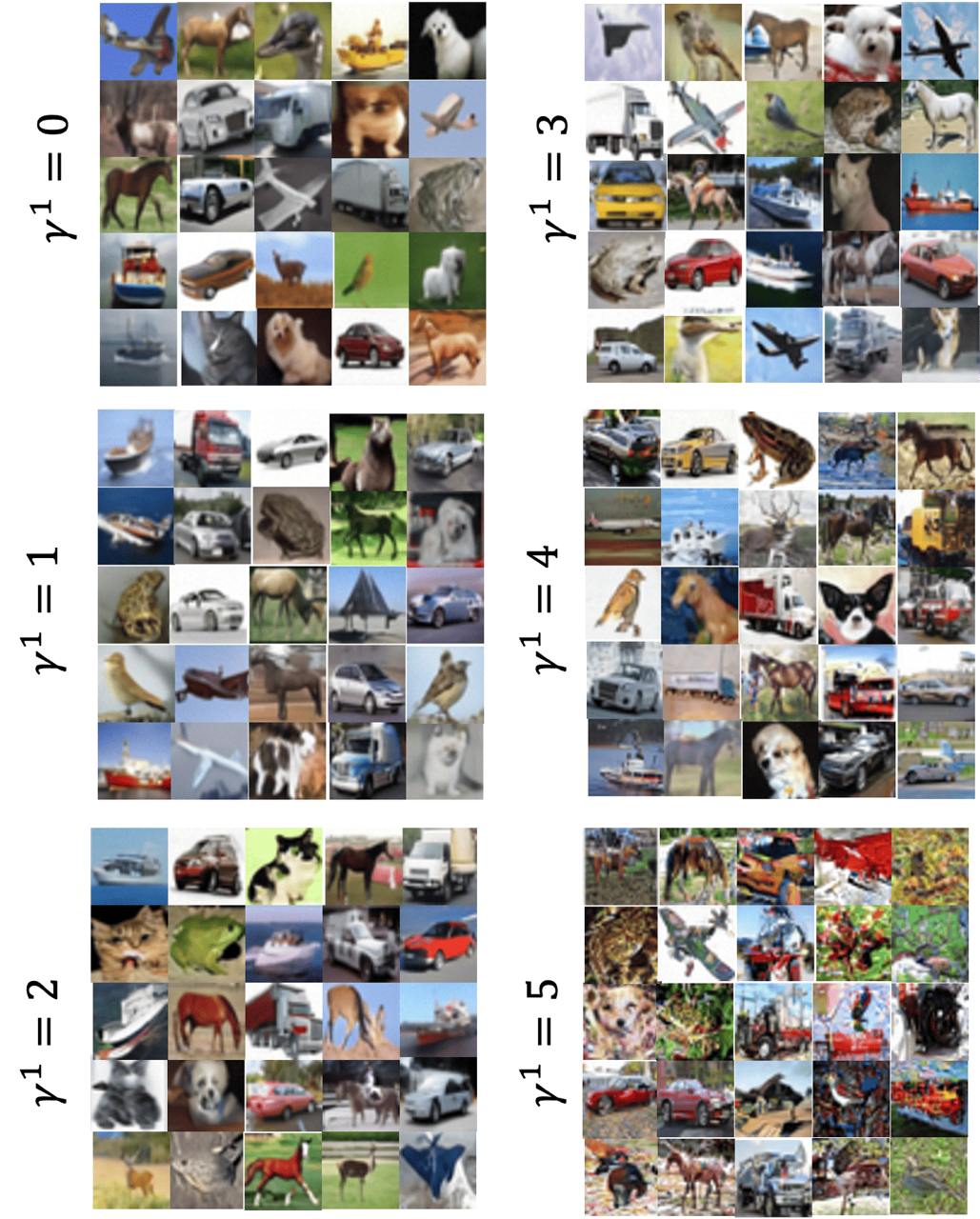}
    \Description{A grid of 32 by 32 pixel images generated from the CIFAR-10 dataset, arranged to show the results of different schedule amplitudes.}
    \caption{Generated samples for CIFAR-10 ($32 \times 32$) with different schedule amplitudes $\bar\gamma$ (and matched diffusion compensation $g_t^2=2\gamma_t^1$) at NFE$=10$.}
    \label{fig:cifar_samples}
\end{figure*}

\begin{figure*}[ht]
\centering
    \includegraphics[width=.8\linewidth]{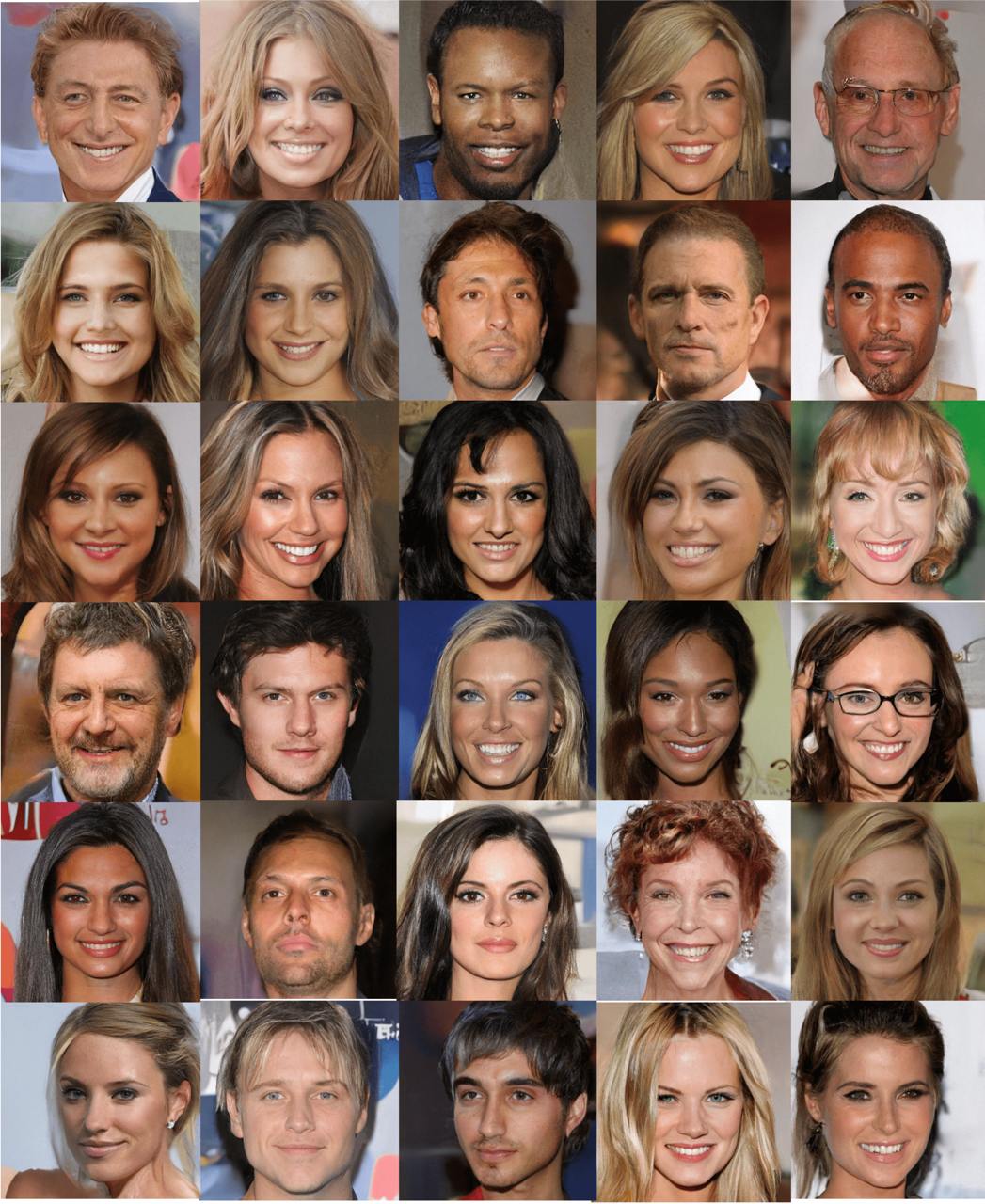}
    \Description{High-resolution 256x256 human face images generated by IDFF.}
    \caption{Generated samples for CelebA-HQ ($256 \times 256$) with $\sigma_0 = 0.2$ and NFE=10.}
    \label{fig:celebHQ_samples}
\end{figure*}

\begin{figure*}[ht]
\centering
    \includegraphics[width=0.5\linewidth]{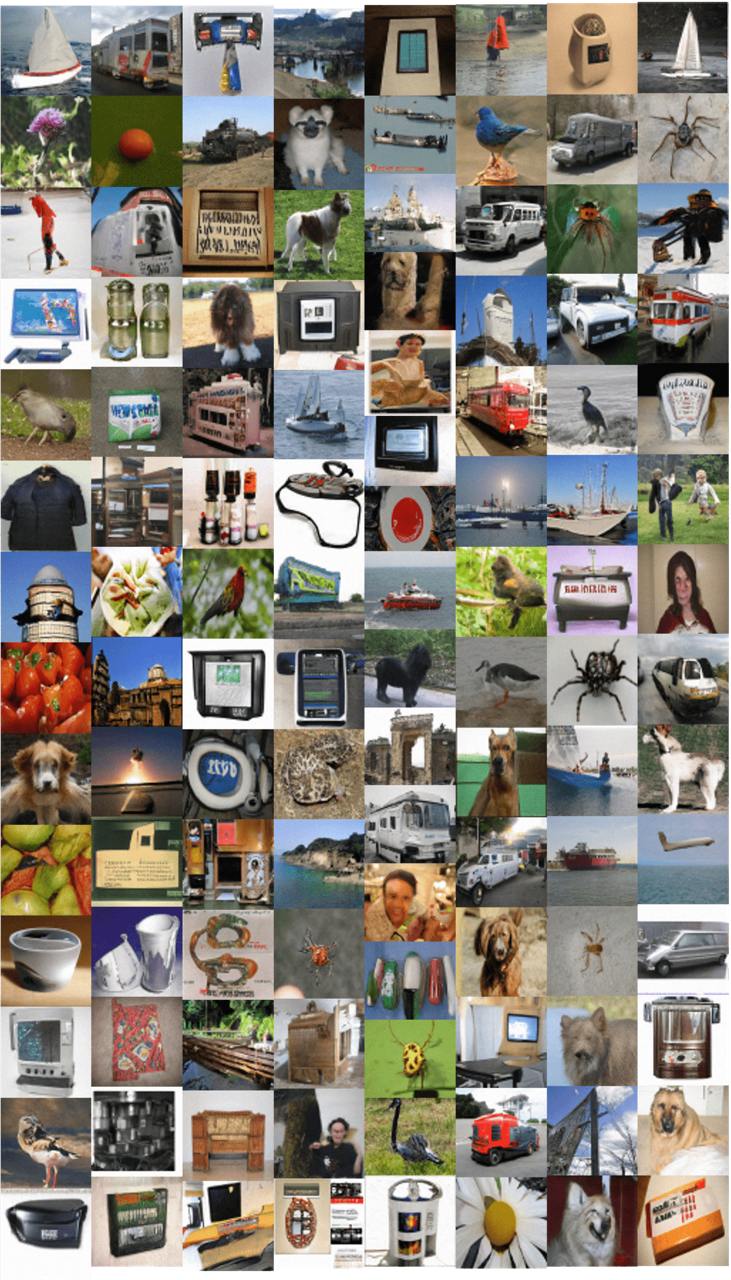}
    \Description{Generated 64x64 images from the ImageNet dataset.}
    \caption{Generated samples for ImageNet-64 with $\sigma_0 = 0.2$ and NFE=10.}
    \label{fig:imagenet_samples}
\end{figure*}

\begin{figure*}[ht]
\centering
    \includegraphics[width=.7\linewidth]{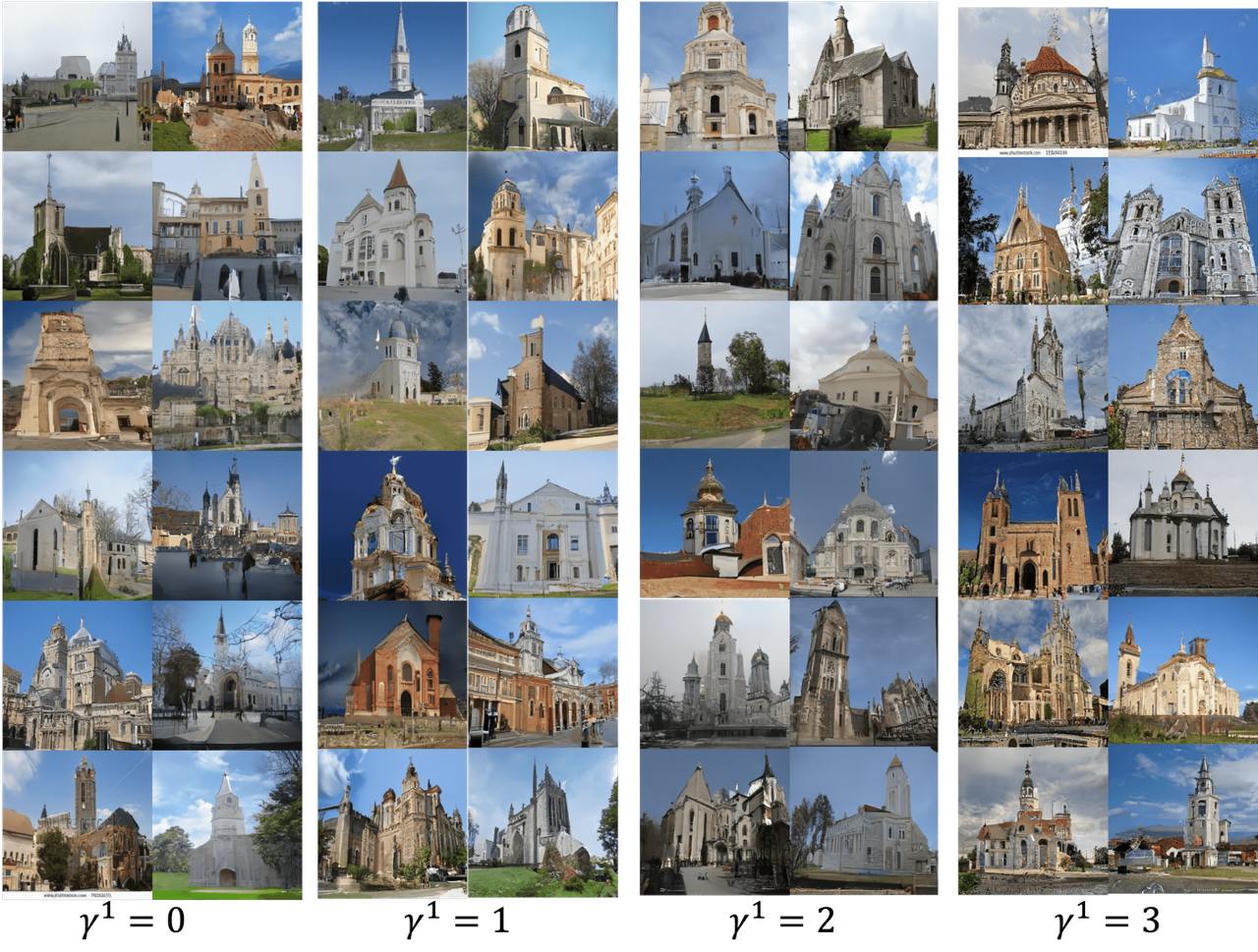}
    \Description{A grid of 256x256 generated architectural scenes.}
    \caption{Generated samples for LSUN-Church ($256 \times 256$) with different schedule amplitudes $\bar\gamma$ (and matched diffusion compensation $g_t^2=2\gamma_t^1$) at NFE$=10$.}
    \label{fig:lsun_church_samples}
\end{figure*}

\begin{figure*}[ht]
\centering
    \includegraphics[width=.7\linewidth]{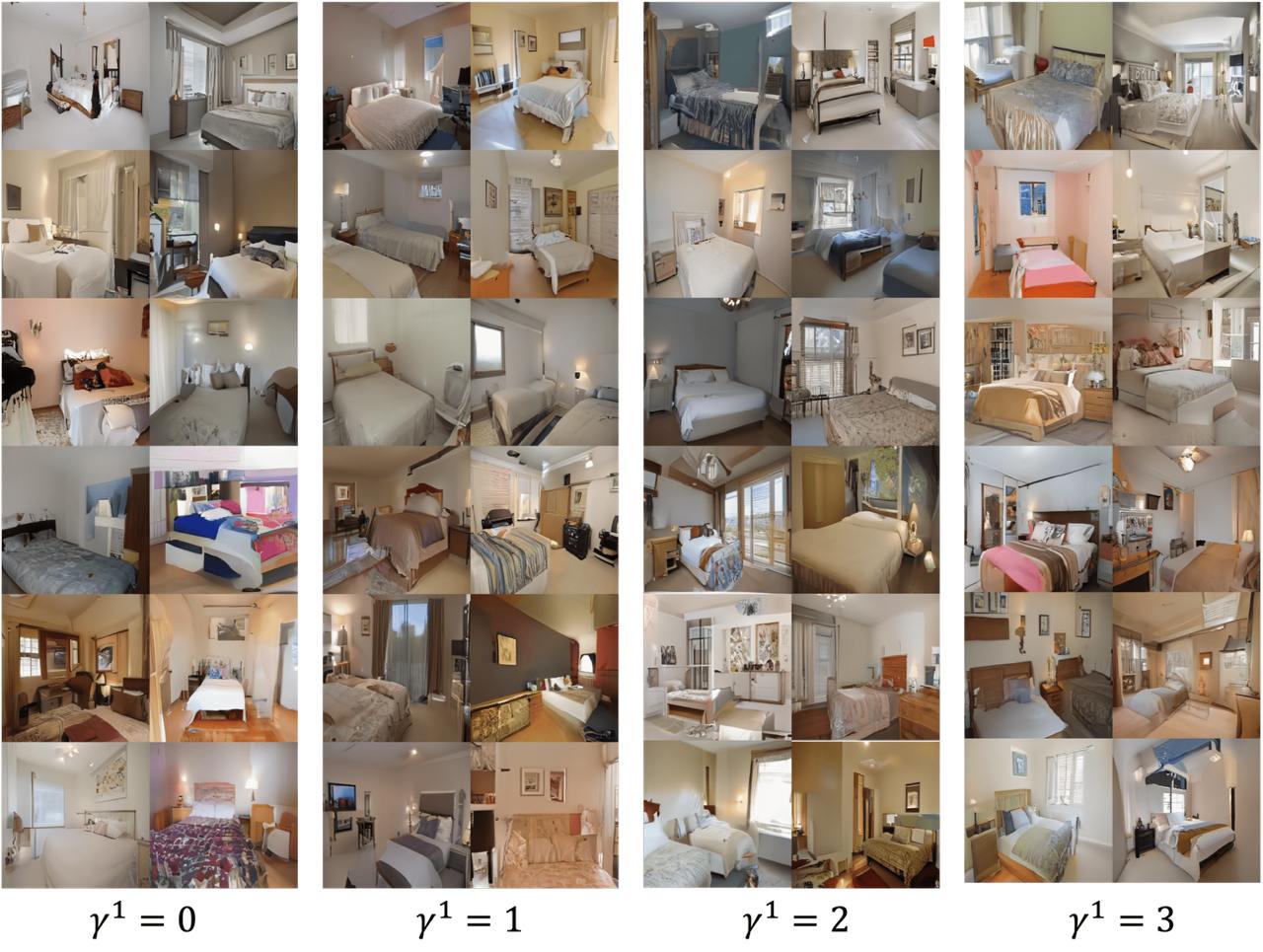}
    \Description{A large grid of 256x256 images of bedroom interiors generated by the IDFF model.}
    \caption{Generated samples for LSUN-Bedroom ($256 \times 256$) with different schedule amplitudes $\bar\gamma$ (and matched diffusion compensation $g_t^2=2\gamma_t^1$) at NFE$=10$.}
    \label{fig:lsun_bed_samples}
\end{figure*}

\begin{table}[ht]
\centering
\caption{FLD performance metrics for various generative models, based on results reported in \citet{jiralerspong2023feature}. IDFF used the ScoreSDE model to generate 10K samples with NFE=10; results for other models are adapted from Table~1 of \citet{jiralerspong2023feature} and are not budget-matched.}
\begin{tabular}{lc}
\toprule
Model & FLD$\downarrow$ \\
\midrule
ACGAN-Mod & 24.22 \\
LOGAN & 18.94 \\
BigGAN-Deep & 9.28 \\
MHGAN & 8.84 \\
StyleGAN2-ADA & 6.86 \\
iDDPM-DDIM & 5.63 \\
IDFF (Ours) & 5.62 \\
StyleGAN-XL & 5.58 \\
PFGM++ & 4.58 \\
\bottomrule
\end{tabular}
\label{tab:-cifa10-fld}
\end{table}

\section{SST Forecasting Visualization}
\label{app:sst}
For this task we employed a class-conditional UNet, with the class encoder handling long-range time dependencies to enable continuous forecasting. Network structure and hyperparameters are in Appendix~\ref{app:image-settings}.
\begin{figure}[ht]
\centering
\includegraphics[width=.7\linewidth]{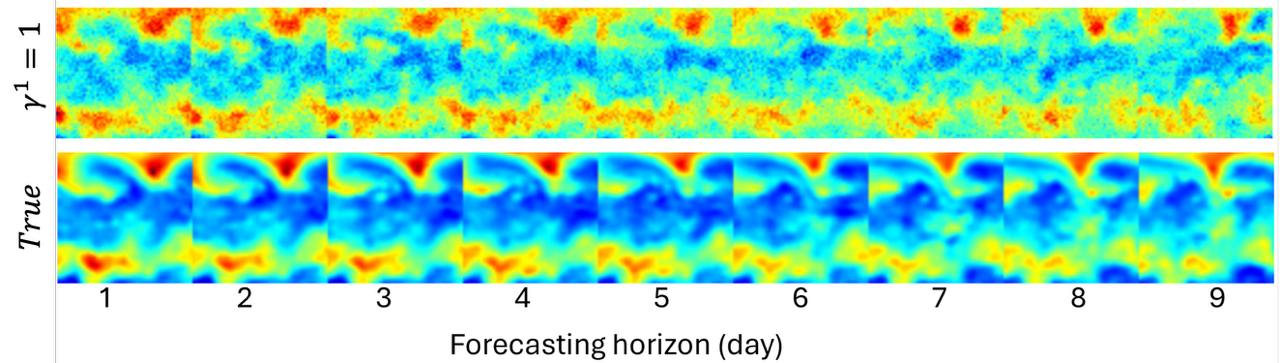}
\Description{A grid showing sea surface temperature forecasts over 9 days.}
    \caption{SST forecasting conditioned on day~1 for 9 days, with the Beta(2,2) schedule at amplitude $\bar\gamma=0.2$ and fixed $\text{NFE}=5$.}
    \label{fig:sst-samples-sigma02}
\end{figure}
\begin{figure*}[ht]
\centering    \includegraphics[width=.7\linewidth]{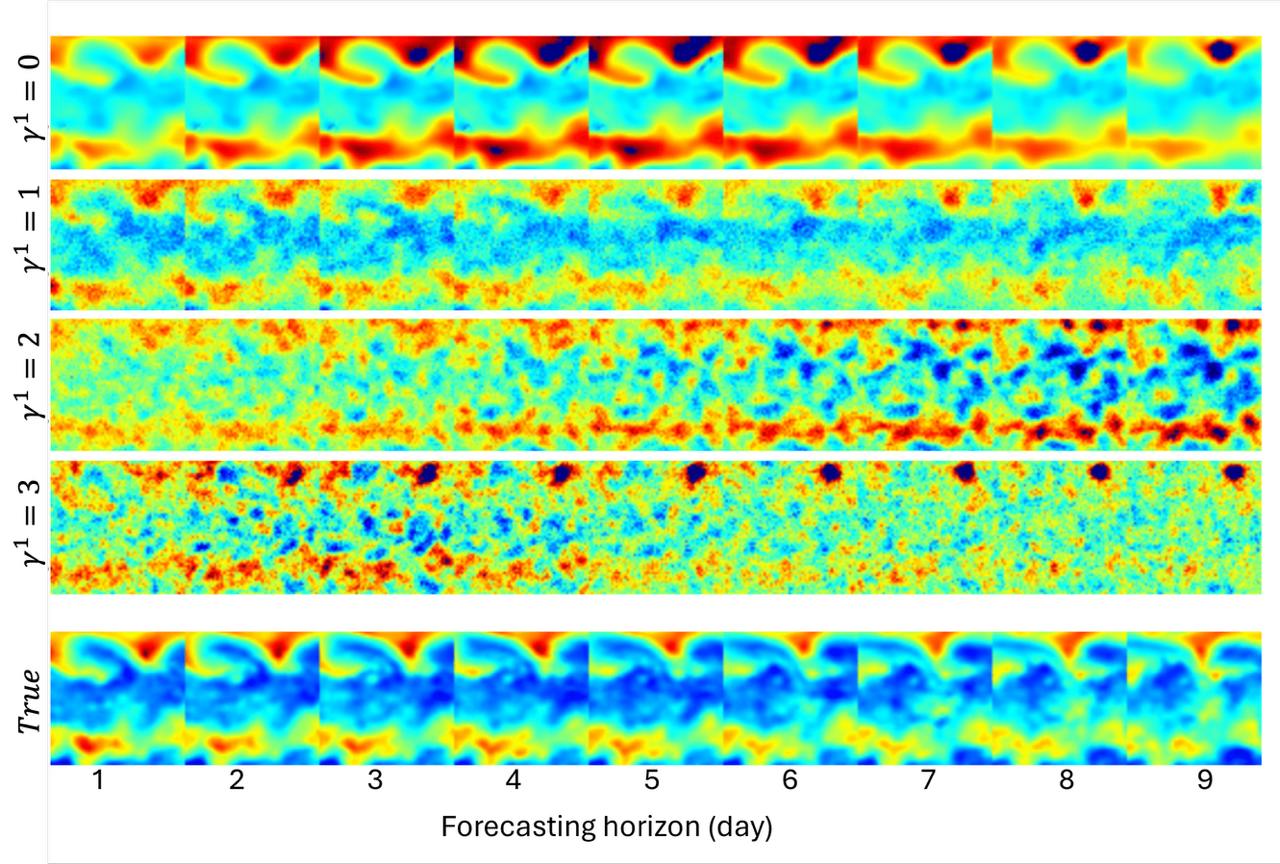}
    \Description{Grids of sea surface temperature heatmaps over 9 forecast days for different schedule amplitudes.}
    \caption{SST forecasting conditioned on day~1 for 9 days for different schedule amplitudes $\bar\gamma$ (with $g_t^2=2\gamma_t^1$) and fixed $\text{NFE}=5$.}
    \label{fig:sst-samples-diff-sigma}
\end{figure*}

\section{2D Simulated Static Data and Time Series}
\begin{figure*}[ht]
\centering
    \includegraphics[width=.5\linewidth]{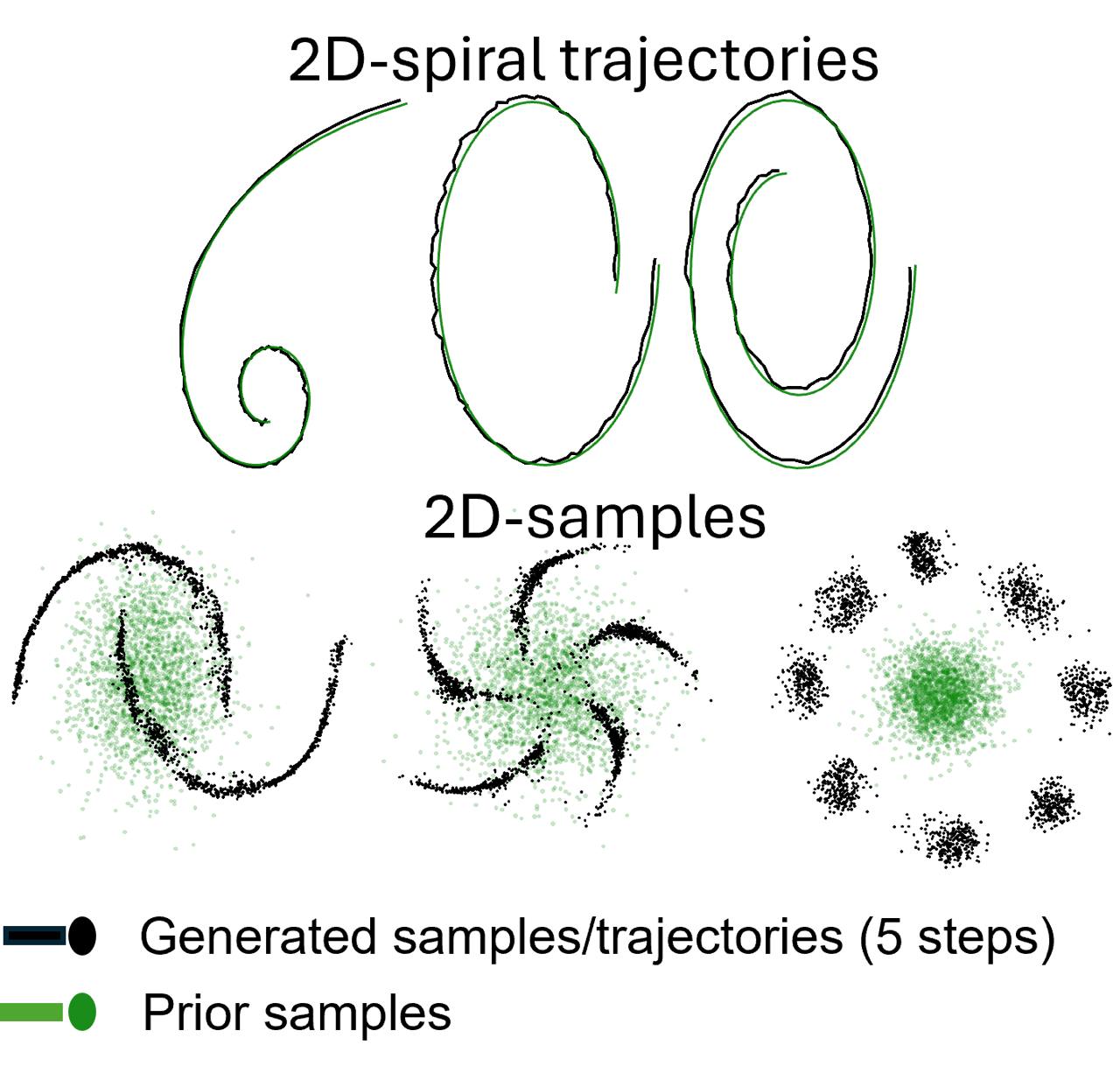}
    \Description{A collection of 2D synthetic data plots comparing ground-truth distributions against IDFF samples.}
    \caption{2D synthetic simulation.}
    \label{fig:2dsyn-gen-iff}
\end{figure*}
Additional results for 2D simulations for both static and time-series generation are shown in Figure~\ref{fig:2dsyn-gen-iff}.

\end{document}